\documentclass[preprint,1p,authoryear,times]{elsarticle} 

\journal{Journal of Systems and Software}

\usepackage{lineno,hyperref}
\usepackage{libertine}
\usepackage{bm}
\usepackage{latexsym}
\usepackage{ifthen}
\usepackage{graphicx}
\usepackage{amssymb}
\usepackage{mathrsfs}
\usepackage[all]{xy}
\usepackage{ifthen}
\usepackage{diagbox}
\usepackage{algorithm}
\usepackage{algpseudocode}
\usepackage{listings}
\usepackage{color}
\usepackage{multirow}
\usepackage{url}
\usepackage{pifont}

\definecolor{codegreen}{rgb}{0,0.6,0}
\definecolor{codegray}{rgb}{0.95,0.95,0.95}
\definecolor{codepurple}{rgb}{0.68,0,0.82}
\definecolor{codeblue}{rgb}{0,0,0.98}
\definecolor{backcolour}{rgb}{0.95,0.98,0.98}

\lstdefinestyle{JavaCode}{
    backgroundcolor=\color{white},   
    commentstyle=\color{codeblue},
    keywordstyle=\color{magenta}\textbf,
    identifierstyle=\color{black},
    numberstyle=\tiny\color{black},
    stringstyle=\color{codepurple},
    basicstyle=\color{red}\ttfamily\scriptsize,
    breakatwhitespace=false,         
    breaklines=true,                 
    captionpos=b,                    
    keepspaces=true,                 
    numbers=none,                    
    numbersep=5pt,                  
    showspaces=false,                
    showstringspaces=false,
    showtabs=false,                  
    tabsize=2,
	frame=single,
}

\floatname{algorithm}{Algorithm}

\newtheorem{Definition}{Definition}
\newtheorem{Theorem} {Theorem}

\newtheorem{Example}{Example}
\newtheorem{Lemma}{Lemma}
\newcommand{\xmark}{\ding{55}}%
\newcommand{\cmark}{\ding{51}}%

\begin{document} 

\begin{frontmatter}

\title{Discovering Boundary Values of Feature-based Machine Learning Classifiers through Exploratory Datamorphic Testing\tnoteref{mytitlenote}}
\tnotetext[mytitlenote]{This paper is an extended and revised version of the conference paper by \citet{AST2020}.}

\author{Hong Zhu}
\ead[1]{hzhu@brookes.ac.uk}
\author{Ian Bayley}
\ead[2]{ibayley@brookes.ac.uk}

\address{School of Engineering, Computing and Mathematics, Oxford Brookes University, Oxford OX33 1HX, United Kingdom}

%\cortext[1]{Corresponding author}

\begin{abstract}
Testing has been widely recognised as difficult for AI applications. This paper proposes a set of testing strategies for testing machine learning applications in the framework of the datamorphism testing methodology. In these strategies, testing aims at exploring the data space of a classification or clustering application to discover the boundaries between classes that the machine learning application defines. This enables the tester to understand precisely the behaviour and function of the software under test. In the paper, three variants of exploratory strategies are presented with the algorithms implemented in the automated datamorphic testing tool Morphy. The correctness of these algorithms are formally proved. Their capability and cost of discovering borders between classes are evaluated via a set of controlled experiments with manually designed subjects and a set of case studies with real machine learning models. 
\end{abstract}

\begin{keyword}
Artificial intelligence, Software testing, Automation of software test, Datamorphic testing, Exploratory testing, Test strategies
\end{keyword}

\end{frontmatter}

%\linenumbers

\section{Introduction}

It is widely recognised that the generation of test data for AI applications is prohibitively expensive \citep{DeepTest2018}. Checking the correctness of a test result is also notoriously difficult, if not completely impossible \citep{Zhou2018Untestable, Zhou2019DriverlessCar}. Moreover, existing testing techniques for measuring test coverage and the automation of testing activities and processes are not directly applicable \citep{DatamorphicTR2019}. Testing AI applications is therefore a grave challenge for software engineering \citep{AST2018Proc}. Developing novel approaches to test AI applications is highly desirable \citep{AITest2019Proc}. 

In \citep{DatamorphicTR2019, Datamorphic2019}, we proposed a method called \emph{datamorphic testing} for testing AI applications and reported a case study with AI applications. In \citep{AITest2020TR, AITest2020} we developed this method further, defined the notion of test morphisms and reported an automated testing tool called \emph{Morphy}. In \citep{AITest2020}, we defined formally a set of test strategies that combine datamorphisms to cover various scenarios in AI applications; \citep{AITest2020TR} reports case studies that show the strategies significantly improve automated in testing AI applications. 

In \citep{AST2020}, we proposed another set of strategies to test the classification and clustering variety of AI applications, as they are very common and arise from machine learning and data analytics techniques; see, for example, \citep{DataMiningTextBook,FoundationsOfMLBook,UnderstandingMLBook}. These strategies are based on the idea of exploratory testing, in which outputs from the previous tests is used to change the focus of testing so that as much as possible of the application's functionality is explored \citep{whittaker2009exploratory}. Whereas confirmatory testing verifies and validates the correctness of the software under test with respect to a given specification, exploratory testing treats it as an object unknown and conducts experiments to discover its functions and features. The two approaches also differ in their treatment of test cases. Confirmatory testing treats test cases as being mutually independent whereas exploratory testing uses the results of earlier test cases to guide the selection of subsequent test cases. In particular, the strategies in \citep{AST2020} aim at discovering the borders between classes of a classifier. The main contributions of \citep{AST2020} are:

\begin{itemize}
\item The notion of \emph{Pareto front} was introduced and formally defined to represent borders betweel classes.
\item Strategies to produce Pareto fronts from machine learning models were formally defined as datamorphic testing algorithms. 
\item The algorithms were formally proved correct and implemented in the Morphy tool.
\item Their cost efficiency was demonstrated by conducting controlled experiments with 10 manually coded classifiers as subjects.
\end{itemize}

This paper extends that work and has the following main contributions: 
\begin{itemize}
\item The notion of completeness is formally defined for a datamorphic test system to be used for exploratory testing. 
\item A systematic method is proposed for constructing exploratory test systems for any feature-based classifier, which are among the most common types of machine learning applications; their completeness was also proven. 
\item We extend the evaluation in \citep{AST2020} by building 48 real machine learning models constructed from 3 real datasets using 8 different machine learning algorithms, in addition to 10 manually coded classifiers already used in \citep{AST2020}. For each strategy, we measure both its cost and its capability of discovering classifier borders. The evaluation found that cost-effectiveness is high for both. 
\end{itemize}

The paper is organised as follows. Section \ref{sec:Preliminaries} defines the basic concepts underlying the work: the basic notions and notations of machine learning classifiers, the exploratory testing approach, the datamorphic testing method and the automated testing tool Morphy. Section \ref{sec:TestSystem} is a theoretical study of the exploration test systems for various types of feature based classifiers, which proves that such test systems exist for all such types of feature based classifiers. Section \ref{sect:StrategyDefinitions} defines the exploration strategies and illustrates their uses with an example. Section \ref{sec:Evaluation} reports the controlled experiments with the 10 manually coded classifiers and 48 machine learning models. Section \ref{sec:RelatedWork} compares the proposal testing method with related work.  
Section \ref{sec:Conclusion} concludes the paper with a discussion of future work. 

\section{Preliminaries}\label{sec:Preliminaries}

In this section, we briefly review the notions and notations underlying our proposed approach. 

\subsection{Classification Applications} 

Clustering as a data mining and machine learning problem is the partitioning of a given set of data points into groups containing similar data points. The grouping is based on a notion of similarity between data points, defined formally with a distance function on the data space. Two pieces of data that are similar to each other should be put into the same group, whilst data that are dissimilar should be placed in different groups. Whereas clustering is unsupervised learning, classification is supervised learning. Given a number of examples of data points and their classifications, the algorithm learns how to assign data to groups \citep{DataMiningTextBook, FoundationsOfMLBook, UnderstandingMLBook}. 

In both clustering and classification, the result is a program  $P$ that maps from the data space $D$ into a number of non-empty groups $G$ such that $D=\bigcup_{g \in G}(g)$ and $\forall g,q \in G.(g \neq q \Rightarrow g \cap q = \emptyset)$. We say that $P$ is a \emph{classification application}. We will write $P(x)$ to denote the output of $P$ on an input $x \in D$, and call $P(x)$ the classification of $x$ by $P$. We also assume that there is a function $\|\cdot,\cdot\|: D \times D \rightarrow \mathbb{R}^+$ ($\mathbb{R}^+ = \{ x \in \mathbb{R} ~|~ x \geq 0 \}$) measuring the distances between any two points $x$ and $y$ in the data space $D$, with shorter distance denoting greater similarity, such that:

\begin{itemize}
\item $\forall x \in D. (\|x,x\|=0)$; 
\item $\forall x,y \in D. (\|x,y\|\geq 0)$;
\item $\forall x, y \in D. (\|x,y\| = \|y,x\|)$; 
\item $\forall x,y,z \in D. (\|x,y\|+ \|y,z\| \geq \|x,z\|)$.
\end{itemize}

For a classification program, it is crucial that data is assigned to the correct classes. However, the borders between classes are often unknown if the classification program is obtained through machine learning and data mining. The goal of the exploratory testing proposed in this paper is to find a set of data pairs that represents the borders between classes. Thus, we introduce the notion of a \emph{Pareto front} for the classification as defined by the program $P$ under test. 

\begin{Definition} \label{def:Pareto} (Pareto Front of Classification) 

Let $P: D \rightarrow G$ be a classification program, $\|\cdot,\cdot\|: D \times  D \rightarrow  \mathbb{R}^+$ be a distance metric defined on the input space $D$, and  $\delta > 0$ be any given real number. A set $\{\left<a_i,b_i\right> |~ a_i, b_i \in D, i=1,\cdots,n \}$  of data pairs is a \emph{Pareto front} of the classes of $D$ according to $P$ with respect to $\|\cdot,\cdot\|$ and  $\delta$, if for all $i=1,\cdots, n, P(a_i ) \neq  P(b_i)$ and $\|a_i,b_i\| \leq \delta$. \qed
\end{Definition}

A Pareto front can show accurately the borders between classes within a tolerable error margin $\delta$. In this way, it helps testers to determine whether the classification is correct or not.

The structure of the data space $D$ determines the type of the classification system. We now define a few standard types that are often seen in the literature.

\begin{Definition} \label{def:FeatureBasedClassifier} (Feature Based Classifier)

Let $P : D \rightarrow G$ be a classification program. We say that $P$ is a \emph{feature based classifier} if there is a natural number $K \geq 1$ such that $D = D_1 \times \cdots D_K$, where for every $i=1, \cdots, K$, $D_i$ is the set of values of a feature $f_i$.  Moreover, a feature $f_i$ is \emph{discrete non-numerical} if $D_i$ is a finite non-empty set. A feature $f_i$ is discrete numerical, if $D_i$ is the set of integer values or natural numbers. A feature $f_i$ is continuous numerical, if $D_i$ is the set of real numbers, or a non-empty interval of real numbers. \qed
\end{Definition}

As these are disjoint alternatives, a feature based classifier can further be classified disjointly according to the types of its features.

\begin{Definition} \label{def:TypesOfFeatureClassifer}(Types of Feature Based Classifiers)

Given a feature based classifier $P: D_1 \times \cdots D_K \rightarrow G$, where $D_i$ is the domain of feature $f_i$, we say that 
\begin{itemize}
\item $P$ is a \emph{discrete non-numerical feature based classifier} or simply a \emph{discrete non-numerical classifier}, if all features $f_i$ are discrete non-numeric. 
\item $P$ is a \emph{discrete numerical feature based classifier} or simply a \emph{discrete numerical classifier}, if all features $f_i$ are discrete numeric. 
\item $P$ is a \emph{continuous numerical feature based classifier}, or simply a \emph{continuous numerical classifier} if all features $f_i$ are continuous numeric. 
\item $P$ is a \emph{hybrid feature based classifier} or simply \emph{hybrid classifier}, if its data space contains more than one type of features. \qed
\end{itemize} 
\end{Definition}

Feature based classifiers are the most common kind of data analytic and machine learning applications. There are other more complicated classifiers, such as time series classifiers, but in this paper we will only study feature based classifiers.
 
%\begin{Definition} \label{def:TimeSeriesClassifier} (Time Series Classifier)
%
%We say that a classification program $P: D \rightarrow G$ is a \emph{time series classifier}, if $D = \bigcup _{n=k}^{u}{(V^n)}$, where $0 < k \leq u$, $V$ is a non-empty set. \qed
%\end{Definition}
%
%In a time series classifier's data space, each data point is a sequence of values $\left<a_1, \cdots, a_n \right>$ in the set $V$. It typically represents a trajectory of a system's behaviour in the form of a sequence of states, where $n>0$ is the \emph{length} of the trajectory. For all $i=1, \cdots, n$, $a_i \in V$ represents the state of the system at time moment $i$. $k$ and $u$ are the minimal and maximal lengths of the trajectories in the data space. When the upper limit of the length is unbounded, $u=\infty$. 
%

\begin{Example}\label{runningExample}
Consider a classifier that classifies the points in a two-dimensional continuous space $[0, 2\pi] \times [-1, 1]$ into three classes: \emph{red}, \emph{black} and \emph{blue} as illustrated in Figure \ref{fig:InputOutputSpace}. This example is a continuous numerical classifier. In this example, data points $x$ and $y$ are a Pareto front pair between \emph{black} and \emph{red} classes, if $x$ is \emph{red} and $y$ is \emph{black} and they are very close to each other. Such pairs can show accurately the borders between classes, and thus help testers to determine whether the classification is correct or not. \qed

\begin{figure}[htbp]
	\centering
	\includegraphics[width=7.5cm]{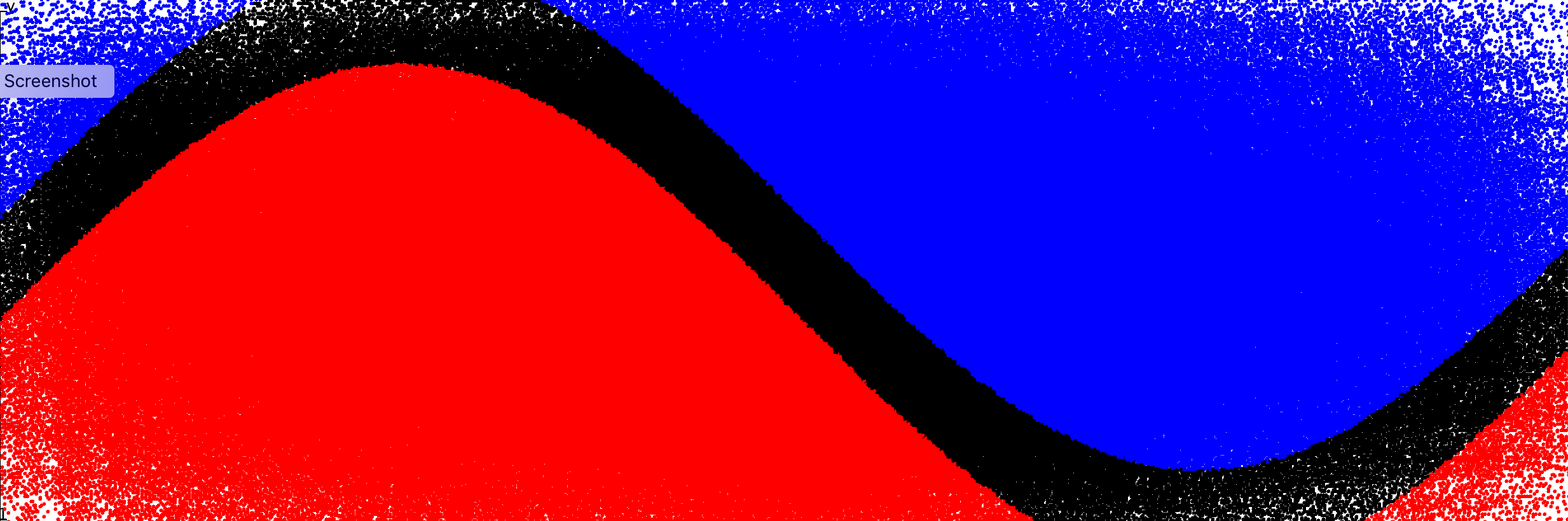}
	\caption{Data Space of the Running Example}
	\label{fig:InputOutputSpace}
\end{figure}
\end{Example}

In the rest of this paper, we will use the above classifier as a running example to explain the definitions of notions and to illustrate the exploration strategies. 

\subsection{Exploratory Testing}

Although exploratory testing (ET) has been widely practised in the industry for a long time, the first use of the term ``exploratory testing" was in a book by Kaner \citeyear{Kaner1988}. It takes a pragmatic approach to software testing under normal business conditions and is based on his experiences as a software testing engineer and manager in the IT industry. Kaner wrote the book initially as a training and survival guide for his staff, but it soon developed into a best seller textbook on software testing used by other practitioners throughout the IT industry \citep{Kaner1988, Kaner1999}. 

Exploration plays an important role in Kaner's approach to software testing. It was soon recognised as an alternative and complementary approach to existing techniques in the literature that emphasize the systematic design and scripting of test cases prior to testing. The notion of ET was further developed by Kaner and other researchers with industry background such as Bach (\citeyear{Bach2002, Bach2003}), Copeland (\citeyear{Copeland2004}), Whittaker (\citeyear{whittaker2009exploratory}), and Hendrickson (\citeyear{Hendrickson2013}). etc. Today, ET is not only widely recognised and practised in the industry, but also has become an active research topic on the software testing.

\cite{Bach2003} defines ET as ``simultaneous learning, test design, and test execution"; according to \cite{Hendrickson2013} this is widely quoted. Other advocates of ET give similar definitions. \citet[Page 113]{Graham_et_al2007} defines it as ``a test design technique where the tester actively controls the design of the tests as those tests are performed and uses information gained while testing to design new and better tests". \citet[Page 202]{Copeland2004} states that ``to the extent that the next test we do is influenced by the result of the last test we did, we are doing exploratory testing. We become more exploratory when we can't tell what tests should be run, in advance of the test cycle."  \citet[Page 339]{Loveland_et_al2005} call ET ``artistic testing", defined as ``testing that takes early experiences gained with the software and uses them to device new tests not imagined during initial planning. It is often guided by the intuition and investigative instincts of the tester". \citet[Page 16]{whittaker2009exploratory} also characterised ET as a process in which ``testers may interact with the application in whatever way they want and use the information the application provides to react, change course, and generally explore the application's functionality without restraint". He argued that ET is not ad hoc but a powerful testing technique.  The power comes from using the information provided by the software under test to alter the course of testing. This process is what \citet[Page 7]{Hendrickson2013} called ``steering". Given its importance in ET, \citet{Hendrickson2013} revised Bach's definition by including steering explicitly. She wrote that ET is ``simultaneously designing and executing tests to learn about the system, using your insights from the last experiment to inform the next". She further identified four essential elements of ET and explains these distinctive key attributes by regarding ET as experiments as follows.
 
\begin{itemize}
\item \emph{Designing}: identifying interesting things to vary and interesting ways in which to vary them so that the experiment can be better performed. 
\item \emph{Executing}: all dynamic testing involves executions of the software on test cases, but in ET a test case is executed immediately when it is designed. 
\item \emph{Learning}: the testers ``discover how the software operates". 
\item \emph{Steering}: using the insights gained from the previous test execution(s) to inform the next. 
\end{itemize}

It is worth noting that ``learning", or more precisely, ``discovery", is perhaps the most fundamental feature that distinguishes ET from traditional approaches to software testing, which is regarded as a validation and verification technique and/or method; see, for example, \citep{KungZhu2009}. \citet{Itkonen_et_al2016} regard such traditional approaches to software testing as confirmatory testing. In other words, it aims to confirm existing theories about the software under test, typically to prove (or disprove) the correctness of the software with regards to the expected output and behaviour. They pointed out that ET aims to discover behaviours that are new in contrast to mechanical executions of pre-scripted test cases. Therefore, as \citet{whittaker2009exploratory} pointed out, ET is most suitable for testing software where a precise specification of the system is not available, such as GUI-based systems. Machine learning applications are also lack precise specifications so ET is applicable for them as well. 

ET is often considered to be a manual testing approach but it need not be. \citet{whittaker2009exploratory} explicitly states that it ``doesn't mean we cannot employ automation tools as aids to the process". \citet{Itkonen_et_al2016} also point out that the goal of test automation in ET is ``to free human resources for other types of testing activities". The goal of this paper is to automate the application of ET in this way when testing machine learning applications. 

ET is usually unscripted, whereas traditional testing is scripted as it pre-specifies test cases, mechanically executes them and compares output values to expected values, also pre-specified. However, ET need not be unscripted. \citet{whittaker2009exploratory} pointed out that ``It isn't necessary to view exploratory testing as a strict alternative to script-based manual testing. In fact, the two can co-exist quite nicely". He distinguishes four types of ET: freestyle, scenarios-based, strategy-based, and feedback-based \cite[Page 184]{whittaker2009exploratory}. He proposed a set of test strategies as guides to exploratory testers and studied a set of scenarios in exploratory testing. From freestyle to feedback-based ET, the patterns and guides for the testers become more and more specific and prescriptive. However, none of these exploratory strategies have been automated. Our approach to automating ET is to formally define the strategies of exploration as algorithms and then to implement them in the framework of datamorphic testing. 

%There are other interesting features that often associated to ET that are different from traditional software testing techniques and methods. \citet{Itkonen_et_al2016} contrasted exploratory testing against confirmatory testing on the underlying philosophy of testing, the process of test design, the role of documentation, the knowledge needs to perform testing tasks and the way they are used, the repeatability of the testing activities, and the role automation in testing. 

\subsection{Datamorphic Testing}\label{sect:OverviewDatamorphicTest}

In the datamorphic software testing method \citep{AITest2020TR}, software artefacts involved in testing are classified into two types: \emph{entities} and \emph{morphisms}. 

\emph{Test entities} are objects and data that are used and/or generated in testing. These include test cases, test suites/sets, the programs under test, and test reports, etc. 

\emph{Test morphisms} are mappings between entities. They generate and transform test entities to achieve testing objectives. They can be implemented as test code and invoked to perform test activities and composed to form test processes. The following are the test morphisms recognised by the datamorphic test tool \emph{Morphy} \citep{AITest2020}. 

\begin{itemize}
 \item \emph{Test set creators} create sets of test cases. They are called \emph{seed test case makers} in \citep{Zhu2015JFuzz,Datamorphic2019}. A typical example is random test case generators like fuzzers \citep{FuzzTestBook2007}. 
 \item \emph{Datamorphisms} are mappings from existing test cases to new test cases. They are called data mutation operators in the data mutation testing method \citep{DataMutationJournal2009}.
 \item \emph{Metamorphisms} are mappings from test cases to Boolean values that assert a program's correctness on test cases. They are test oracles. Formal specifications and metamorphic relations in metamorphic testing \citep{MetamorphicTestingSurveyChen2018, Zhou2018Untestable} can also be used as metamorphisms. \emph{Mutational metamorphic relations} introduced in \citep{Zhu2015JFuzz} are metamorphisms. 
 \item \emph{Test case metrics} are mappings from test cases to real numbers. They measure test cases giving, for example, the similarity of a test case to the others in the test set. 
 \item \emph{Test case filters} are mappings from test cases to truth values. They can be used, for example, to decide whether a test case should be included in a test set. 
 \item \emph{Test set metrics} are mappings from test sets to real numbers. They measure the test set quality, such as its code coverage \citep{ZhuHallMay1997}.  
 \item \emph{Test set filters} are mappings from test sets to test sets. For example, they may remove redundant test cases from a test set for regression testing. 
 \item \emph{Test executers} execute the program under test on test cases and receive the outputs from the program. They are mappings from a piece of program to a mapping from input data to output. That is, they are functors in category theory \citep{CategoryTheoryBook}. 
 \item \emph{Test analysers} analyse test sets and generate test reports. Thus, they are mappings from test sets to test reports. 
 \end{itemize}
 
A \emph{test system} $\mathscr{T}=\left< \mathscr{E}, \mathscr{M}\right>$ in datamorphic testing consists of a set $\mathscr{E}$ of test entities and a set $\mathscr{M}$ of test morphisms. In Morphy \citep{AITest2020TR}, a test system is specified as a Java class that declares a set of attributes as test entities and a set of methods as test morphisms. 

Given a test system, Morphy provides testing facilities to automate testing at three different levels. At the lowest level, various test activities can be performed by invoking test morphisms via a click of buttons on Morphy's GUI. At the medium level, Morphy implements various test strategies to perform complex testing activities through combinations and compositions of test morphisms. At the highest level, test processes are automated by recording, editing and replaying test scripts that consist of a sequence of invocations of test morphisms and strategies. 

Test strategies are complex combinations of test morphisms designed to achieve test automation. Three sets of test strategies have been implemented in Morphy: 

\begin{itemize}
 \item \emph{Mutant combination}: combining datamorphisms to generate mutant test cases; see \citep{AITest2020TR}. 
 \item \emph{Domain exploration}: searching for the borders between clusters/subdomains of the input space; 
 \item \emph{Test set optimisation}: optimising test sets by employing genetic algorithms. 
 \end{itemize}
 
This paper focuses on domain exploration strategies, which will be defined in Section \ref{sect:StrategyDefinitions}. 

\subsection{Overview of The Proposed Approach}

The approach of this paper and its previous work \citep{AST2020} is to apply the four ET principles identified previously to test feature-based classifiers built using machine learning and data analytics techniques: 

Firstly, on test design, the variations in test cases are formally defined by a set of datamorphisms that can be applied to the features of the classifier under test. These datamorphisms are employed to explore the data space of the ML application. A major contribution of this paper is to formally define the notion of \emph{completeness} for ET test systems, and we prove that complete test systems exist for feature-based classifiers; see Section \ref{sec:TestSystem}. This enables a complete exploration of the input space. 

Secondly, on execution, in our approach, each time a new test case is generated, the ML model is invoked, and the output of the invocation is used to generate the next test case. In fact, the test executor is an important component of our definition of ET test systems; see Section \ref{sec:TestSystem}.  

Thirdly, on learning, our goal in testing is to discover the borders between classes as defined by the ML model under test. Such information is unknown before testing, but the results in the form of Pareto front can improve significantly the tester's knowledge about the behaviour of the model.

%For example, Figure 1 shows the Pareto fronts of various ML models of the manually coded classifier \emph{Box 2} generated from the same training dataset using different ML techniques. They provide a clear view of models' behaviours. The existence of Pareto front brings a number of possible new ways to analyse ML models. We are further investigating how to use the information contained in Pareto front in the testing and development of ML applications. 
 
%Figure 1. Visualisation of the Pareto Fronts of Various ML Models of Box 2

Finally, on steering, we study three strategies in which the outputs of previously executed test cases are used in three different ways to decide the next test case. These strategies are defined as algorithms and implemented in the automated datamorphic testing environment Morphy. We will also formally prove that these strategies correctly achieve the goal of exploration, i.e. they detect the borders between classes as defined by the ML model under test; see Section \ref{sect:StrategyDefinitions}. 

We will also automate the testing process by implementing the technique in the datamorphic testing framework. 

\section{Exploratory Test Systems for Feature Based Classifiers}\label{sec:TestSystem}

Exploratory test systems are test systems for ET. In this section, we will introduce the notion of exploratory test systems and the notion of completeness for such test systems. We will then constructively prove the existence of complete test systems for each type of feature based classifier. 

\subsection{Structure of Exploratory Test System}

To apply an exploratory test strategy to a classification program $P: D \rightarrow G$ with a distance function $\|\cdot,\cdot\|$, we require that the test system $\mathscr{T} = \left< \mathscr{E}, \mathscr{M}\right>$ has the following properties. 
\begin{enumerate}
\item The set $\mathscr{M}$ of morphisms contains a test executer $Exe_P(x)$ that executes the program $P$ under test on a test case $x$ and receives the output of $P$; that is $Exe_P(x)= P(x)$. In the sequel, we will write $P(x)$ for $Exe_P(x)$ for the sake of simplicity. 
\item There is a set $W \subseteq \mathscr{M}$ of unary datamorphisms defined on $D$. Informally, for each $w \in W$ and $x \in D$, $w(x),w^2 (x)$, $\cdots$, $w^n(x)$ can generate a sequence of data points in $D$, where $w^1(x)=w(x)$, $w^{n+1}(x)=w(w^n(x))$. These datamorphisms are called \emph{traversal methods}. 
\item There is also a binary datamorphism $m \in \mathscr{M}$ such that 
\begin{equation}
\forall x,y \in D. \left(\|x,y\| > \delta_m \Rightarrow \|x,m(x,y)\| < \|x,y\| \wedge \|y,m(x,y)\| < \|x,y\|\right), \label{eqn:E2_0}
\end{equation}
\noindent{where $\delta_m = Min_{ x\neq y\in D}\{\|x,y\|\}$.}

Informally, the datamorphism $m$ calculates a point between $x$ and $y$, if the distance between them is greater than the minimal distance $\delta_m$ among points in the data space. We will call $m$ the \emph{midpoint method}.
\end{enumerate}

Note that for all $x,y \in D$, because the program $P$ under test classifies $x$ and $y$ into different classes, the midpoint $m(x,y)$ between $x$ and $y$ must be either not in the same class as $x$ or not in the same class as $y$. Formally, we have: 
\begin{eqnarray}
(P(x) \neq P(y)) \Rightarrow (P(x) \neq P(m(x,y)) \vee (P(y) \neq P(m(x,y)). \label{eqn:Eqn0}
\end{eqnarray}

Also, note that it is unnecessary to include the distance metric $\|\cdot,\cdot\|$ in the test system as a test morphism. As we will see in Section \ref{sect:StrategyDefinitions}, the algorithms of exploratory test strategies do not need it.

\subsection{Completeness of Exploratory Test Systems}

For a test system to be able to explore the whole data space of a classifier, we require the set of datamorphisms is able to reach every data point in the space by applying the datamorphisms on any arbitrary starting point. We say such a set of datamorphisms is \emph{complete}. Completeness may not be possible for a classifier on continuous data space. In such cases, we would like to reach the target point as close as is desired. This property of test system is called \emph{approximate completeness}. 

Before we formally define these notions of completeness, we first define the notion of compositions of datamorphisms. Let $\mathscr{M} \neq \emptyset$ be a set of datamorphisms. 

\begin{Definition}\label{def:composition} (Composition of Datamorphisms)
Let $X$ be a set of variables ranging over test cases. The set of compositions of datamorphisms in $\mathscr{M}$ is recursively defined as follows. 
\begin{enumerate}
\item For all $x \in X$, $x$ is a composition of datamorphisms in $\mathscr{M}$ of order 0.
\item $m(e_1, \cdots, e_k)$ is a composition of datamorphisms in $\mathscr{M}$ of order $n+1$, if $m \in \mathscr{M}$ is $k$-ary, and $e_1, \cdots, e_k$ are compositions of datamorphisms in $\mathscr{M}$, and $n$ is the maximum of the orders of $e_1, \cdots, e_k$. \qed
\end{enumerate}
\end{Definition}

Informally, a composition of datamorphisms is an expression with datamorphisms as the operators and variables as the parameters. For example, $m_1(m_2(m_3(x_1, x_2)))$ is a composition of two unary datamorphisms $m_1, m_2$ and one binary datamorphism $m_3$, where $x_1$ and $x_2$ are variables. Given a composition of datamorphisms, a test case can be obtained by substituting existing test cases for the variables of the composition, and we say that the result is a mutant test case obtained by applying the composition to the existing test cases. 

%A composition can be equally represented in the form of a parse tree of the expression, in which the leaf nodes are variables and non-leaf nodes are datamorphisms. The sub-trees of a non-leaf node are its sub-expressions. The height of the tree equals the \emph{order} of the composition, which also equals the order of the mutant test cases obtained by applying the composition on seed test cases. For example, the tree in Figure \ref{fig:Trees}(1) is a mutant test case obtained by applying the composition $m_1(m_2(m_3(x_1, x_2)))$ to seed test cases $s_1$ and $s_2$. Its order is 3.
%
%
%\begin{figure}[h]
%\centering
%\scalebox{0.6}{\includegraphics{figures/MutantTree.pdf}}
%\caption{Examples of Mutant Trees}
%\label{fig:Trees}
%\end{figure}

\begin{Definition}\label{def:Completeness}(Completeness)

An exploratory test system $\mathscr{T} = \left< \mathscr{E}, \mathscr{M}\right>$ on data space $D$ is \emph{complete}, if for all $a, b \in D$, there is a composition $\phi(x)$ of datamorphisms in $\mathscr{M}$ such that $b = \phi(a)$. 

An exploratory test system $\mathscr{T}$ is \emph{approximately complete}, if for all $a, b \in D$ and every $\delta > \delta_m$, there is a composition $\phi$ of datamorphisms in $\mathscr{M}$ such that $\|b, \phi(a)\| \leq \delta$. \qed
\end{Definition}

Note that, in a real-world application, in a multi-dimensional data space some combinations of feature values may be invalid or meaningless. For example, a human who is 2 meters tall but only weights 20kg is physically impossible. Our completeness requirements on an exploratory test system still require the test system to cover such data. This will enable testing on invalid inputs, which are useful, for example, to understand how the software will react to input errors. 

In the remainder of this section, we construct a complete or approximately complete exploratory test system for each type of feature based classifier.
% to prove that exploratory testing can be applied to feature based classifiers. 

%Let $P$ be a feature based classifier on data space $D = D_1 \times \cdots \times D_K$.

\subsection{Continuous Numerical Classifiers}\label{sec:ConituousNumericalClassifier}

Given a continuous numerical classifier, we construct two unary datamprophisms $up_i(x)$ and $down_i(x)$ for each feature $f_i$ as the traversal methods and a binary datamorphism $mid_E(x,y)$ as the midpoint method. The set of datamorphisms will form an approximately complete test system. Let $c_i>0$ be a given constant real value. We define:
\begin{eqnarray}
& up_i(\left<x_1,\cdots, x_K\right>)=\left<x_1, \cdots, x_i+c_i, \cdots x_K\right> \label{eqn:e2_6}\\
& down_i(\left<x_1,\cdots, x_K\right>)=\left<x_1, \cdots, x_i-c_i, \cdots x_K\right> \label{eqn:e2_7}\\
& mid_E(\left<x_1,\cdots, x_K\right>, \left<y_1, \cdots, y_k\right>)=\left<\frac{x_1+y_1}{2}, \cdots \frac{x_K + y_K}{2}\right> \label{eqn:e2_8}
\end{eqnarray}

There are many different ways that we can define distance metrics on real numbers. The following is the Euclidean distance on multi-dimensional real numbers. 
\begin{equation}
\|\left<x_1,\cdots, x_K\right>, \left<y_1, \cdots, y_k\right>\|_E=\sqrt{\sum_{i=1}^{K}{(x_i -y_i)^2}} \label{eq18}
\end{equation}

The following are a few well-known properties of Euclidean distance, which are useful for proving the approximate completeness of the test system. 
\begin{Lemma}\label{lmm:Lmm2_5}
The distance metrics $\|\cdot,\cdot\|_E$ has the following properties. 
\begin{enumerate}
\item $\forall x \in D.~\|x,x\|_E=0$;
\item $\forall x, y \in D.~\|x,y\|_E \geq 0$; 
\item $\forall x,y \in D.~\|x, z\|_E < \|x,y\|_E \wedge \|z,y\|_E < \|x, y\|_E$, where $z=mid_E(x,y)$. 
\item $\forall x,y \in D.~\|x,z\|_E = \frac{\|x,y\|_E}{2}$, where $z=mid_E(x,y)$. \qed
\end{enumerate}
\end{Lemma}

Let $W_E = \{ up_i(x) ~|~ i=1, \cdots, K\} \cup \{down_i(x) ~|~ i=1, \cdots, K\} \cup \{mid(x,y)\}$.
Applying these properties of the midpoint datamorphism $mid_E(x,y)$ and Euclidean distance metrics $\|x,y\|_E$, we can prove that the set of datamorphisms $W_E$ defined above satisfies the requirements of exploratory test systems on datamorphisms.

\begin{Theorem}\label{thm:Thm2_5}
The set $W_E$ of datamorphisms together with the distance metrics $\|x,y\|_E$ satisfy the conditions of exploratory test systems on datamorphisms. 
\end{Theorem}
\noindent{\emph{Proof}.}
By (\ref{eq18}), $\delta_m = Min_{x \neq y \in D}\left(\|x,y\|_E\right) = 0$. Therefore, by Lemma \ref{lmm:Lmm2_5}(4), the condition given in Equation (\ref{eqn:E2_0}) is true. The theorem is true. \qed

\begin{figure}[htbp]
	\centering
	\includegraphics[width=10cm]{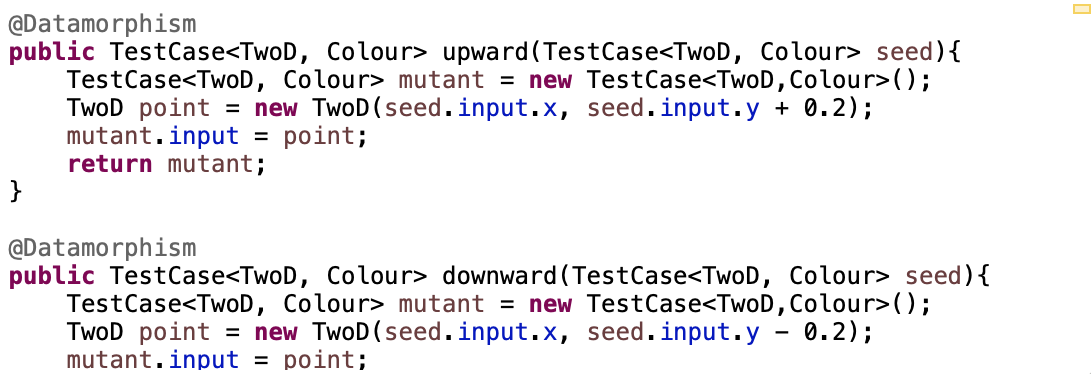}
	\includegraphics[width=10cm]{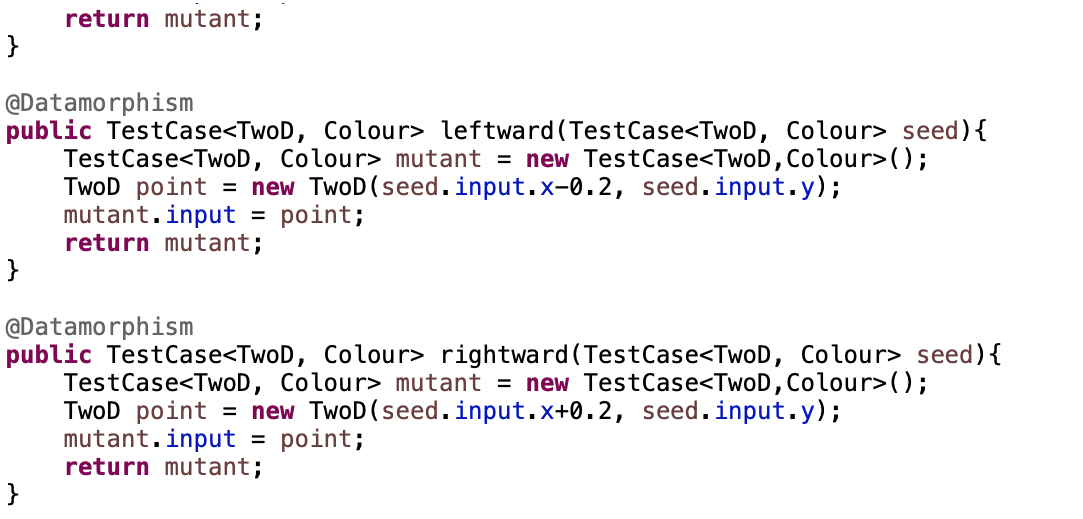}
	\includegraphics[width=10cm]{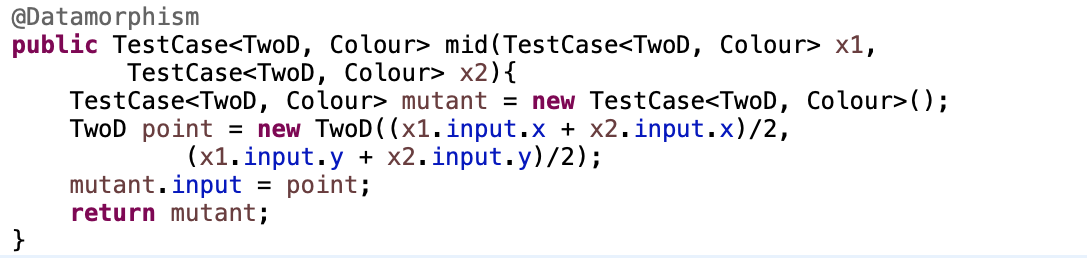}
%	\lstinputlisting[language=Java]{figures/SinClassify.java}
	\caption{Datamorphisms of the Running Example}
	\label{fig:WalkingMethods}
\end{figure}

\begin{Example} \label{RunningExampleTestSystem}
Figure \ref{fig:WalkingMethods} gives the traversal and midpoint methods in the Morphy test specification for the classifier of the running example. The \emph{leftward} and \emph{rightward} methods implement the traversal methods $down_x$ and $up_x$, respectively. The \emph{upwards} and \emph{downward} methods implement the traversal methods $up_y$ and $down_y$, respectively, where $c_x = c_y = 0.2$. The method \emph{mid} implements the $mid_E$ datamorphism, which calculates the geometric midpoint between $x$ and $y$ as defined in equation (\ref{eqn:e2_8}). Therefore, by Theorem \ref{thm:Thm2_5}, they form an exploratory test system with the following distance function.
\begin{eqnarray*}
\|\left<x_1, x_2\right>, \left<y_1, y_2\right>\| = \sqrt{(x_1 - y_1)^2 + (x_2 - y_2)^2}
\end{eqnarray*} \qed
\end{Example}

The following theorem states that $W_E$ is approximately complete. 

\begin{Theorem}\label{thm:Thm2_6}
The set $W_E$ of datamorphisms is approximately complete for a continuous numerical feature based classifier $P$ defined on the data space $D = D_1 \times \cdots \times D_K$, $K>0$. 

\noindent{\emph{Proof}.} 
We prove that for any given points $a=\left<a_1, \cdots, a_K\right>, b=\left<b_1, \cdots, b_K\right> \in D$ and $\delta > 0$, we can construct a composition $\phi$ of datamorphisms such that $\|b, \phi(a)\|_E <\delta$. The composition $\phi(x)$ is defined as follows. 
\begin{equation}
\phi(x) = m^{n_\delta} \circ ud^{n_1}_1 \circ \cdots \circ ud^{n_K}_K(x)),
\end{equation}
where 
\begin{eqnarray*}
&&m(x) = mid_E(b, x), 
~~~~ud^{n_i}_i(x) = \left\{ \begin{array}{ll}
up^{n_i}_i(x) & \textrm{if $a_i \geq b_i$}\\
down^{n_i}_i(x) & \textrm{if $a_i < b_i$}
\end{array} \right., \\
&&n_i = \lfloor \frac{|a_i - b_i|} {c_i} \rfloor, 
~~~~ n_\delta = \lceil ln(\frac{c}{\delta})\rceil, 
~~~~ c=\sqrt{\sum_{n=1}^K{c_i^2}}.
\end{eqnarray*}

Note that $ud_i(x)$ is either $up_i(x)$ or $down_i(x)$ depending on whether the $i'$th element of $a$ is greater than the $i'$th element of $b$. 

Let $a'=ud^{n_1}_1 \circ \cdots \circ ud^{n_K}_K(x)=\left<a'_1, \cdots, a'_K\right>$. We have that $a'$ is obtained by applying $ud_i(x)$ for $n_i$ times on $a$, for $i=1, \cdots, K$. The $i'$th element of $a'$ will be $a'_i = a_i \pm n_i \cdot c_i$ by the definition of datamorphisms $up_i(x)$ and $down_i(x)$. By the definition of $n_i$, we have that $|b_i-a'_i| < c_i$, for all $i=1,\cdots,K$. Therefore, \[\|b,a'\| = \sqrt{\sum_{n=1}^K{(b_i-a'_i)^2}} \leq \sqrt{\sum_{n=1}^K{c_i^2}}=c.\] 
Applying $m(x)$ on $a'$ for $n$ times, we get $a'' = m^n(a')$. By Lemma \ref{lmm:Lmm2_5}(4), we have that $\|b, a''\|=\|b, a'\|/2^n \leq c/2^n$. Therefore, when $n \geq ln(\frac{c}{\delta})$, we have that $\|b, a''\| \leq \delta$. The theorem follows immediately that $n_\delta = \lceil ln(\frac{c}{\delta})\rceil \geq ln(\frac{c}{\delta})$. \qed
\end{Theorem}

\begin{Example}
The exploratory test system given in Example \ref{RunningExampleTestSystem} is approximately complete, because for all points $a, b$ in the data space and $\delta >0$, we have a composition $\phi(x)$ of datamorphisms such that $\|b, \phi(a)\| \leq \delta$; see Figure \ref{fig:WalkingPath} for an illustration of how to construct the composition of datamorphisms. \qed

\begin{figure}[htbp]
	\centering
	\includegraphics[width=10cm]{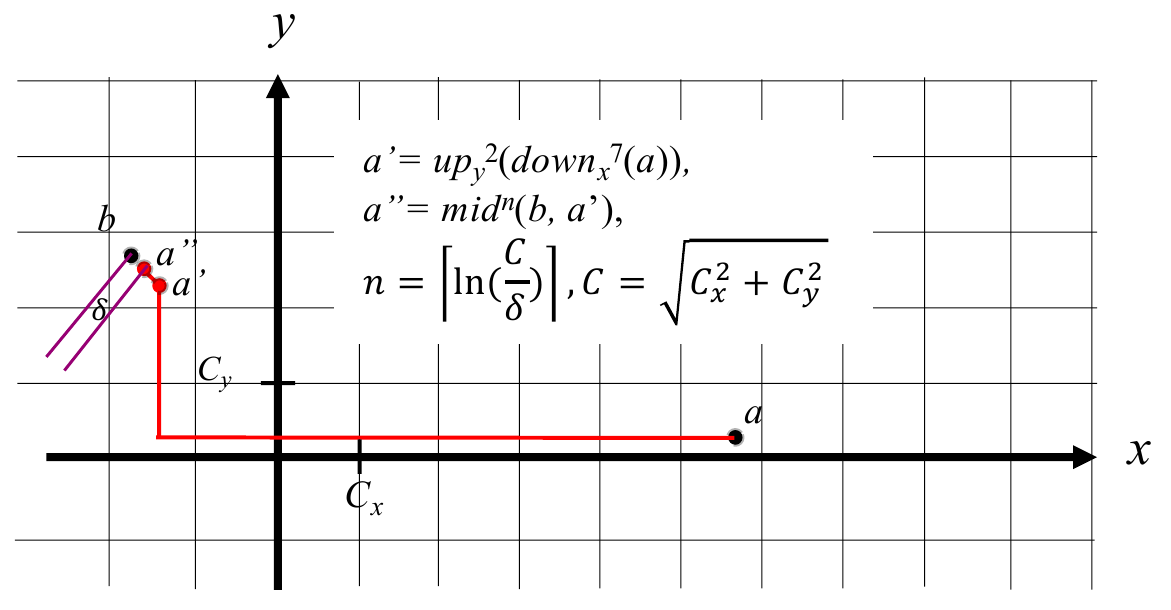}
	\caption{Construction of the Walk Path in the Running Example}
	\label{fig:WalkingPath}
\end{figure}
\end{Example}

\subsection{Discrete Non-Numerical Classifiers.}

If the classifier $P$ is a discrete non-numerical feature based classifier then for each $i=1, \cdots, K$, $D_i$ is a non-empty finite set. Let $D_i = \{v_{i,1}, v_{i,2}, \cdots, v_{i,n_i}\}$, where $n_i >0$. We define two unary datamorphisms $up_i(x)$ and $down_i(x)$ as the traversal methods as follows.

\begin{equation}
up_i(\left<x_1, \cdots, x_K\right>)=\left<x_1, \cdots, x'_i, \cdots, x_K\right>, 
\textrm{ where }
x'_i = \left\{\begin{array}{ll}
v_{i,j+1} & \textrm{if $x_i = v_{i,j}$ and $j < n_i$}\\
v_{i,n_i} & \textrm{if $x_i = v_{i,n_i}$}
\end{array} \right.
\label{eqn:e2_1}
\end{equation}

\begin{equation}
down_i(\left<x_i, \cdots, x_K\right>)=\left<x_i, \cdots, x'_i, \cdots, x_K\right>, 
\textrm{where }
x'_i = \left\{\begin{array}{ll}
v_{i,j-1} & \textrm{if $x'_i = v_{i,j}$ and $j>1$}\\
v_{i,1} & \textrm{if $x'_i = v_{i,1}$}
\end{array} \right.
\label{eqn:e2_2}
\end{equation}

Let $x, y \in D$, $x=\left<x_1, \cdots, x_K\right>$ and $y=\left<y_1, \cdots, y_K\right>$. The distance between $x$ and $y$, written $\| x, y \|_D$, is defined as the number of elements in $x$ and $y$ that are different. Let $\Delta(x,y)=\left<d_1, \cdots, d_k\right>$, $0\leq k \leq K$, be the sequence of elements in $x$ that are different from the corresponding elements in $y$. Therefore, we have that $\| x, y \|_D = k$. 

The following Lemma states that the function $\| \cdot, \cdot \|_D : D \times D \rightarrow N$ satisfies the conditions of distance metrics. The proof is straightforward, and thus is omitted for the sake of space. 

\begin{Lemma}\label{lmm:Lmm2_1}
The function $\|\cdot,\cdot \|_D: D \times D \rightarrow N$ defined above satisfies the conditions of distance metrics. That is, for all $x,y,z \in D$, we have that $\|x, x\|_D=0$, $\|x,y\|_D\geq 0$, $\|x,y\|_D = \|y,x\|_D$, and $\|x,y\|_D + \|y,z\|_D \geq \|x,z\|_D$. \qed
\end{Lemma}

We now define a binary datamorphism $mid_D(x,y)$ as the midpoint method as follows.

\begin{equation}
mid_D(x,y)= \left<z_1,\cdots, z_K\right>,
\end{equation}
\noindent{where}
\begin{equation}
z_i=\left\{\begin{array}{ll}
 x_i & \textrm{if $x_i = y_i$}\\
 x_i & \textrm{if $x_i \neq y_i$ and $x_i$ is an odd-indexed element in $\Delta(x,y)$} \\
 y_i & \textrm{if $x_i \neq y_i$ and $x_i$ is an even-indexed element in $\Delta(x,y)$}
\end{array} \right.
\label{eqn:e2_3}
\end{equation}

The following theorem gives some useful special properties of the distance metrics $\| ~ \|_D$ and midpoint datamorphism $mid_D$ on discrete data space. These properties are easy to prove by using the definitions of the distance function and discrete non-numerical data space. Details are omitted for the sake of space. 

\begin{Lemma}\label{lmm:Lmm2_2}
For all $x,y \in D$, we have that 
\begin{enumerate}
\item $x \neq y \Rightarrow \|x,y\|_D \geq 1$;
\item $\|x,y\|_D \leq K$;
\item $mid_D(x,x)=x$;
\item $\|x,y\|_D = 1 \Rightarrow (mid(x,y)=x) \vee (mid_D(x,y)=y)$;
\item $\|x,y\|_D > 1 \Rightarrow \|x,z\|_D < \|x,y\|_D \wedge \|z,y\|_D < \|x,y\|_D$, where $z=mid(x,y)$. \qed
\end{enumerate}
\end{Lemma}

Let $W_D = \{ up_i(x) ~|~ i=1, \cdots, K\} \cup \{down_i(x) ~|~ i=1, \cdots, K\} \cup \{mid_D(x, y)\}$. 

\begin{Theorem}\label{thm:Thm2_1}
$W_D$ and the distance metrics $\|\cdot,\cdot\|_D$ together satisfy the requirements of exploratory test systems on datamorphisms. 
\end{Theorem}
\noindent{\emph{Proof}.} By Lemma \ref{lmm:Lmm2_2}(1), $\delta_m = Min_{ x\neq y\in D}\{\|x,y\|_D\} = 1$. By Lemma \ref{lmm:Lmm2_2}(5), $mid_D(x,y)$ and $\|x,y\|_D$ meet the condition on the midpoint method given in Equation (\ref{eqn:E2_0}). Thus, the theorem is true. \qed

The following theorem states that the set of datamorphisms constructed above is complete. 

\begin{Theorem}\label{thm:Thm2_2}
The set $W_D$ of datamorphisms is complete for a discrete non-numerical feature based classifier $P$ defined on the data space $D = D_1 \times \cdots \times D_K$, $K>0$.  

\noindent{\emph{Proof}.} 
For any given points $a, b \in D$, we construct a composition of datamorphisms $\phi(x)$ such that $\phi(a) = b$. 
We define 
\begin{equation}
\phi(x) = ud^{n_1}_1 \circ \cdots \circ ud^{n_K}_K(x), 
\end{equation}
\noindent{where}
\begin{equation}
ud^{n-i}_i(x) = \left\{ \begin{array}{ll}
up^{n_i}_i(x) & \textrm{if  $a_i=v_{i,c_a}\in D_i$, $b_i=v_{i,c_b} \in D_i$, and $c_b \geq c_a$}\\ \label{eq8}
down^{n_i}_i(x) & \textrm{if $a_i=v_{i,c_a}\in D_i$, $b_i=v_{i,c_b} \in D_i$ and $c_b < c_a$}
\end{array} \right., n_i = |c_x -c_y|.
\end{equation}

By (\ref{eq8}), $ud_i(x)$ is either $up_i(x)$ or $down_i(x)$ depending on the difference between $a$ and $b$ on the $i'$th element, and $n_i$ is the distance between the $i'$th elements of $a$ and $b$. By the definitions of $up_i(x)$ and $down_i(x)$, we have that $\phi(a)=b$. Therefore, by Definition \ref{def:Completeness} of completeness, the theorem is true. \qed
\end{Theorem}

\subsection{Discrete Numerical Classifiers}

For a discrete numerical classifier, we also define two unary datamorphisms $up_i(x)$ and $down_i(x)$ for each feature $f_i$ as the traversal methods. The $up_i(x)$ datamorphism on feature $f_i$ is defined as follows. 
\begin{equation}
up_i(\left<x_1, \cdots, x_K\right>)=\left<x_1, \cdots, x'_i, \cdots, x_K\right>,
\textrm{ where } x'_i = x_i + 1. 
\label{eqn:e2_4}
\end{equation} 

The datamorphism $down_i(x)$ is defined as follows.
\begin{equation}
down_i(\left<x_1, \cdots, x_K\right>)=\left<x_1, \cdots, x'_i, \cdots, x_K\right>
\label{eqn:e2_5}
\end{equation}
where $x'_i = x_i - 1$, if the data set $D_i$ is the set of integer values; and 
%\begin{displaymath}
$x'_i = \left\{ \begin{array}{ll} 
x_i - 1 & \textrm{if $x_i>0$}\\
0 & \textrm{if $x_i=0$}
\end{array}
\right.$, 
%\end{displaymath}
if the data set $D_i$ is the set of natural numbers. 

The midpoint datamorphism $mid_N(x,y)$ is defined as follows. 
\begin{equation}
mid_N(\left<x_1, \cdots, x_K\right>, \left<y_1, \cdots, y_K\right>) = \left<\lfloor \frac{|x_1 - y_1|}{2}\rfloor, \cdots, \lfloor \frac{|x_K - y_K|}{2}\rfloor\right>
\end{equation}

Now, we define the distance metric $\|\cdot,\cdot\|_N$ on the data space, as follows. 
\begin{equation}
\|\left<x_1, \cdots, x_K\right>, \left<y_1, \cdots, y_K\right>\|_N = \sum^K_{i=1}{|y_i - x_i|}
\end{equation}

Similar to Lemma \ref{lmm:Lmm2_1}, we can prove that the function $\| \cdot, \cdot \|_N : D \times D \rightarrow N$ satisfies the conditions of distance metrics. The proof is straightforward, and thus is omitted for the sake of space. 
\begin{Lemma}
The function $\|\cdot,\cdot \|_N: D \times D \rightarrow N$ defined above satisfies the condition of distance metrics. That is, for all $x,y,z \in D$, we have that $\|x, x\|_N=0$, $\|x,y\|_N\geq 0$, $\|x,y\|_N = \|y,x\|_N$, and $\|x,y\|_N + \|y,z\|_N \geq \|x,z\|_N$. \qed
\end{Lemma}

The midpoint datamorphism $mid_N(x,y)$ and the distance metrics $\|x,y\|_N$ have the following properties. Again, they are easy to prove by the definitions of the distance function and discrete numerical data space. Details are omitted for the sake of space.
\begin{Lemma}\label{lmm:Lmm2_3}
For all $x,y \in D$, we have that 
\begin{enumerate}
\item $x \neq y \Rightarrow \|x,y\|_N \geq 1$;
\item $mid_N(x,x) = x$;
\item $\|x,y\|_N = 1 \Rightarrow (mid_N(x,y)=x) \vee (mid_N(x,y)=y)$;
\item $\|x,y\|_N > 1 \Rightarrow \|x,z\|_N < \|x,y\|_N \wedge \|z,y\|_N < \|x,y\|_N$, where $z=mid(x,y)$. \qed
\end{enumerate} 
\end{Lemma}

Let  $W_N = \{ up_i(x) ~|~ i=1, \cdots, K\} \cup \{down_i(x) ~|~ i=1, \cdots, K\} \cup \{mid(x, y)\}$. The following theorem states that the set $W_N$ of datamorphisms constructed above satisfies the conditions of exploratory test systems. The proof is very similar to that of Theorem \ref{thm:Thm2_1} so the details are omitted for the sake of space.

\begin{Theorem}\label{thm:Thm2_3}
$W_N$ and the distance metrics $\|\cdot,\cdot\|_N$ together satisfy the requirements of exploratory test systems on datamorphisms. \qed
\end{Theorem} 

The following theorem states that the set $W_N$ of datamorphisms constructed above is complete. 

\begin{Theorem}\label{thm:Thm2_4}
The set $W_N$ of datamorphisms is complete for a discrete numerical feature based classifier $P$ defined on the data space $D = D_1 \times \cdots \times D_K$, $K>0$, 

\noindent{\emph{Proof}.} 
For any given points $a, b \in D$, we construct a composition $\phi(x)$ of datamorphisms such that $\phi(a) = b$. 
We define 
\begin{equation}
\phi(x) = ud^{n_1}_1 \circ \cdots \circ ud^{n_K}_K(x),
\end{equation}
\noindent{where } 
\begin{equation}
ud^{n_i}_i(x) = \left\{ \begin{array}{ll}
up^{n_i}_i(x) & \textrm{if $a_i \geq b_i$}\\ \label{eq14}
down^{n_i}_i(x) & \textrm{if $a_i < b_i$}
\end{array} \right., ~~n_i = |a_i -b_i|. 
\end{equation}

By (\ref{eq14}), $ud_i(x)$ is either $up_i(x)$ or $down_i(x)$ depending on the difference between $a$ and $b$ on the $i'$th element, and $n_i$ is the absolute value of the distance between the $i'$th elements of $a$ and $b$. By the definitions of $up_i(x)$ and $down_i(x)$, we have that $\phi(x) = b$. Therefore, by Definition \ref{def:Completeness} of completeness, the theorem is true. \qed
\end{Theorem}

\subsection{Hybrid Feature Based Classifiers}

Let $P: D \rightarrow C$ be a hybrid feature based classifier. Without lost of generality, we assume that $D = D_1 \times \cdots \times D_u \times N_1 \times \cdots \times N_v \times R_1 \times \cdots \times R_w$, where $D_1, \cdots, D_u$ are discrete non-numerical features, $N_1, \cdots, N_v$ are discrete numerical features, and $R_1, \cdots, R_w$ are continuous numerical features, and at least two of $u, v$ and $w$ are greater than zero. 

We now define unary datamorphisms $up_i(x)$ and $down_i(x)$ as the traversal methods as follows. 
\begin{enumerate}
\item If feature $f_i$ is discrete non-numerical, we use Equation (\ref{eqn:e2_1}) to define $up_i(x)$. 
\item If feature $f_i$ is discrete numerical, we use Equation (\ref{eqn:e2_4}) to define $up_i(x)$. 
\item If feature $f_i$ is continuous numerical, we use Equation (\ref{eqn:e2_6}) to define $up_i(x)$. 
\end{enumerate}

Similarly, we define $down_i(x)$ depending on the type of features and using the Equations (\ref{eqn:e2_2}), (\ref{eqn:e2_5}) and (\ref{eqn:e2_7}), accordingly. 

Before we formally define a binary datamorphism as the midpoint method and a distance metric, let us first introduce some notation. 

Let $x=\left<d_1, \cdots, d_u, n_1, \cdots, n_v, r_1, \cdots, r_w\right> \in D$. We write $x_D = \left<d_1,\cdots, d_u\right>$, $x_N =\left<n_1, \cdots, n_v\right>$, and $x_E=\left<r_1, \cdots, r_w\right>$. We also write $x = x_D \oplus x_N \oplus x_E$. In general, $\oplus$ is an operator on vectors defined as follows. 
\[\left<x_1, \cdots, x_n\right> \oplus \left<y_1, \cdots, y_m\right> = \left<x_1, \cdots, x_n, y_1, \cdots, y_m\right>\]
 
Now, we define a binary datamorphism $mid_H(x,x')$ as follows.
\begin{equation}
mid_H(x,x')=mid_D(x_D,x'_D) \oplus mid_N(x_N,x'_N) \oplus mid_E(x_E,x'_E)
\end{equation}

We now define $\|\cdot ,\cdot \|_H :D\times D \rightarrow \mathbb{R}^+$ as follows. 
\begin{equation}
\|x, x'\|_H= \|d, d'\|_D + \|n,n'\|_N + \|r,r'\|_E. \label{eqn:e2_10}\\
\end{equation}

The following lemma states that the above equation defines a distance metric. It follows immediately the properties of $\|\cdot,\cdot\|_D$, $\|\cdot,\cdot\|_N$ and $\|\cdot,\cdot\|_E$. Details are omitted.

\begin{Lemma}
Function $\|\cdot,\cdot \|_H$ satisfies the conditions of distance metrics. \qed
\end{Lemma}

Let $W_H = \{up_i(x)\}_i \cup \{down_i(x)\}_i \cup \{mid_H(x)\}$, where $up_i(x)$, $down_i(x)$ and $mid_H(x,y)$ are defined as above. 

\begin{Theorem}
The set of datamorphisms $W_H$ and the distance metrics $\|x,y\|_H$ together satisfy the conditions of exploratory test systems. 
\end{Theorem}
\noindent{\emph{Proof}.}
First, from the definition of $\|x,y\|_H$, we have that $\delta_m = Min_{x\neq y\in D}\{\|x,y\|_H\}$. If there is at least one feature in the data space $D$ that is a continuous numerical feature, then it is easy to see that $\delta_m = 0$. Otherwise, all features are either discrete non-numerical or discrete numerical so we have $\delta_m = 1$. 

Second, let $x,y \in D$ and $\|x,y\|_H > \delta_m$, and $z=mid_H(x,y)$. By the definitions of $mid_H(x,y)$ and $\|x,y\|_H$, we have that 
\begin{eqnarray*}
\lefteqn{\|x,z\|_H = \|x_D, z_D\|_D + \|x_N, z_N\|_N + \|x_E, z_E\|_E}\\
&& = \|x_D, mid_D(x_D, y_D)\|_D + \|x_N, mid_N(x_N, y_N)\|_N + \|x_E, mid_E(x_E, y_E)\|_E\\
&& < \|x_D, y_D\|_D + \|x_N, y_N\|_N + \|x_E, y_E\|_E \\
&& = \|x,y\|_H
\end{eqnarray*}

Similarly, we have $\|y,z\|_H < \|x,y\|_H$. Therefore, the theorem is true. \qed

\begin{Theorem}
Let $P$ be a hybrid feature based classifier, and $W_H$ be the set of datamorphisms defined above.  
\begin{enumerate}
\item If there is a continuous numerical feature in the data space of $P$, $W_H$ is approximately complete. 
\item If there is no continuous numerical feature in the data space of $P$, the set $W_H$ of datamorphisms is complete. 
\end{enumerate}
\end{Theorem}

\noindent{\emph{Proof}.} 

Similar to the proofs of Theorem \ref{thm:Thm2_2}, \ref{thm:Thm2_4} and \ref{thm:Thm2_6}, for any given points $a$ and $b$ in the data space, and any given real number $\delta>0$, we construct a composition $\phi(x)$ of datamorphisms such that $\|b, \phi(a)\|_H \leq \delta$. 

Let $a = a_D \oplus a_N \oplus a_E$ and $b=b_D \oplus b_N \oplus b_E$. 

By the proof of Theorem \ref{thm:Thm2_2}, there is a composition of datamorphisms $\phi_D(x)$ such that $b_D = \phi_D(a_D)$. 

By the Theorem \ref{thm:Thm2_4}, there is a composition $\phi_N(x)$ of datamorphisms such that $b_n = \phi_N(a_N)$. 

By Theorem \ref{thm:Thm2_6}, there is a composition $\phi_E(x)$ of datamorphisms such that $\|b_E, \phi_E(a_E)\| \leq \delta$. 

By the definition of the datamorphisms for hybrid feature based classifier, $\phi_D(x)$, $\phi_N(x)$ and $\phi_E(x)$ are also compositions of the datamorphisms in $W_H$. Therefore, $\phi(x) = \phi_E \circ \phi_N \phi_D(x)$ is a composition of datamorphisms in $W_H$. 

Let $a' = \phi(a)$. It is easy to see that $\phi(a) = \phi_D(a_D) \oplus \phi_N(a_N) \oplus \phi_E(a_E)$. Therefore, 
\begin{eqnarray*}
\lefteqn{\|b, a'\|_H = \|b,\phi(a)\|_H}\\
&&= \|b_D \oplus b_N \oplus b_E, \phi_D(a_D) \oplus \phi_N(a_N) \oplus \phi_E(a_E)\|_H \\
&&= \|b_D, \phi_D(a_D)\|_D + \|b_N, \phi_N(a_N)\|_N + \|b_E, \phi_E(a_E)\|_E\\
&&= 0 + 0 + \|b_E, \phi_E(a_E)\|_E \leq \delta 
\end{eqnarray*}

Therefore, statement (2) of the theorem is true. 

If there is no continuous numerical feature in the data space, i.e. $b_E$ and $a_E$ are empty, then $\|b_E, \phi_E(a_E)\|_E = 0$. Therefore, in such a case, statement (1) is true. 
\qed

\section{Exploration Strategies} \label{sect:StrategyDefinitions}

This section presents the algorithms of three different exploratory strategies for testing clustering and classification applications. We also prove their correctness and illustrate their behaviour by using the running example given in the previous section. 

\subsection{Random Target Strategy}

Let's start with a simple exploration strategy based on random selection of two test cases in order to find the Pareto front of the classification groups between these two test cases. We call this strategy \emph{random target strategy}.  

The strategy starts by selecting a pair of two test cases $x$ and $y$ at random. If the outputs of the program $P$ under test on these test cases are different, i.e. $P(x) \neq P(y)$, then a point $z_1$ between $x$ and $y$ is generated by using the binary datamorphism of the midpoint method $mid(x,y)$, i.e. $z_1 = mid(x,y)$. The program $P$ is executed on this mutant test case $z_1$ to classify it. The classification of $z_1$ must be different from one of the original pair of test cases; say $P(z_1) \neq P(x)$. Thus, we can repeat the above steps with $x$ and $z_1$ as the pair of test cases, and a further mutant $z_2$ can be generated. This process is repeated a number of times to ensure the distance between the final pair of points is small enough. See Algorithm \ref{alg:AimedWalk}. 

\begin{algorithm}[h] 
\caption{(Random Target Strategy)}\label{alg:AimedWalk}
\begin{small}
\begin{algorithmic}
\Require 
        $testSet$: Test Pool; 
        $steps$: Integer; 
        $mid(x,y)$: Binary datamorphism; 
\Ensure  
        $a$, $b$: Test Case; \\
\hspace{-0.25cm}{\textbf{Begin}}
\State 1: Select two different test cases $x$ and $y$ in $testSet$ at random; 
\State 2: Execute program $P$ on test cases $x$ and $y$;
\State 3: Check if a pair of the Pareto front exists between $x$ to $y$:
\If {($x.output = y.output$)} \Return $\left< null, null \right>$ \EndIf
\State 4: Refinement:
\For {$i \gets 1$ to $steps$} 
	\State $z = mid(x,y)$;
	\If {($x.output \neq z.output$)} { $y=z$}
	\Else { $x = z$;}
	\EndIf
\EndFor ;
\State $a = x$;  $b = y$; 
\State \Return $\left< a, b \right>$;  \\
\hspace{-0.25cm}{\textbf{End}}
\end{algorithmic}
\end{small}
\end{algorithm}

Let $n>0$ be any given natural number. We write $RT(n)=\left< a,b \right>$ to denote the results of executing Algorithm \ref{alg:AimedWalk} with $n$ as the parameter $steps$ and $\left< a, b \right>$ as the output. 

Assume that the exploratory test system has the following properties.
\begin{enumerate}
\item There is a constant $c>1$ such that  
\begin{eqnarray}
\forall x,y \in D.\left(\frac{Max\{\|x,z\|,\|z,y\|\}}{\|x,y\|}\right) \leq 1/c, \label{eqn:Eqn1}
\end{eqnarray} 
where $z=mid(x,y)$. 
\item There is a constant $d_m>0$ such that
\begin{eqnarray}
\forall x,y \in D.(\|x,y\| \leq d_m). \label{enq:Enq2}
\end{eqnarray}
\end{enumerate}

Then, we have the following theorem about the correctness of the random target strategy algorithm. 

\begin{Theorem}\label{thm:Thm1}
If $RT(n)=\left< a, b \right> \neq \left< null, null \right>$, then $\left< a ,b \right>$ is a pair in the Pareto front according to $P$ with respect to $\|\cdot,\cdot\|$ and  $\delta$, if $d_m/c^n < \delta$. 
\end{Theorem}
\noindent{\emph{Proof.}}
If $RT(n)=\left<a, b \right> \neq \left< null, null \right>$ then the condition of the If-statement in step (3) is \textbf{false}. Thus, the loop is executed. It is easy to see that the For-loop in \emph{Step 4} in the algorithm terminates. 

We now prove that the following is a loop invariant by induction on the number $i$ of iterations of the loop body. 
\[\|x,y\| \leq \frac{d_m}{c^i} \wedge P(x) \neq P(y).\] 

When entering the loop, by assumption (\ref{enq:Enq2}), the distance between the data points stored in variable $x$ and $y$ satisfies the following inequality.
\[\|x,y\| \leq d_m\]
Since the condition of the If-statement is false, we have that 
\[P(x)=x.output \neq y.output=P(y).\] 
Therefore, the loop invariant is true for $i=0$. 

Assume that the loop invariant is true for $i=n \geq 0$. 

After the execution of the loop body one more time (i.e. $i=n+1$), by applying the Hoare logic of the If-statements in the loop body, the distance $d_x'$ between the data points stored in variables $x$ and $y$ will become either $\|x,z\|$ or $\|z,y\|$, where $z=mid(x,y)$. By assumption (\ref{eqn:Eqn1}), in both cases we have that 
\[d_x' \leq Max\{\|x,z\|,\|z,y\|\} \leq {\|x,y\|}/c \leq d_m /c^{n+1}.\]
By the condition of the If-statement in the loop body and the property (\ref{eqn:Eqn0}), applying Hoare logic we have that, after the execution of the loop body, the data points stored in variables $x$ and $y$ have the property that $P(x) \neq P(y)$.
Therefore, the condition is a loop invariant according to Hoare logic. 

When the loop exits, $i = steps=n$. By Hoare logic, after executing the assignment statements $a=x$ and $b=y$, we have that 
\[\|a,b\| \leq d_m/c^n \wedge P(a) \neq P(b).\] 
Therefore, the theorem is true by Definition \ref{def:Pareto}. 
\qed

The algorithm of random target strategy can be run multiple times to generate a number of pairs for the Pareto front. 
\begin{Example}\label{exm:Exm1}
For example, applying the random target strategy to the running example, we can obtain a test set shown in Figure \ref{fig:AimedWalkResult} when 1000 pairs of test cases are selected at random from a test set of 300 random test cases. A total of 641 pairs of Pareto front test cases were generated. The success rate in  generating a pair for the Pareto front is 64.1\%. The set of Pareto front pairs shows clearly the boundary between the subdomains classified by the software.  

\begin{figure}[htbp]
	\centering
	\includegraphics[width=7.5cm]{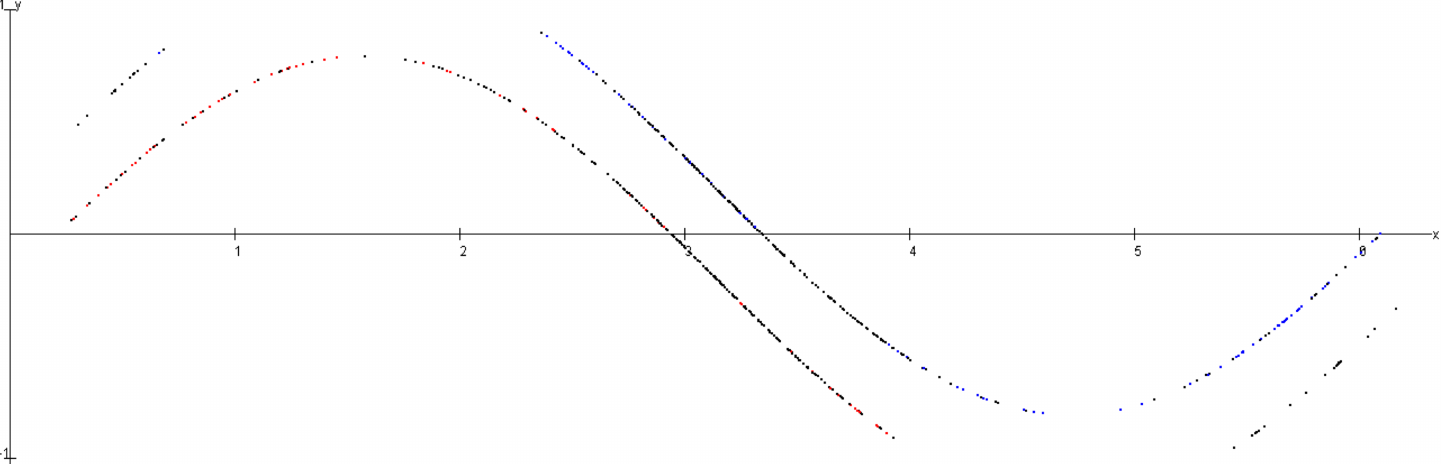}\\
	\caption{Pareto Front Generated by Random Target}
	\label{fig:AimedWalkResult}
\end{figure}

In this example, the number of steps $n$ is 20. Since the data space $D=[0, 2\pi]\times[-1,1]$, if the distance function $\|x,y\|$ is $Eucl(x,y)$, we have that $d_m=2\sqrt{\pi^2+1}$. By the definition of $mid(x,y)$, we have that 
\[\frac{Max(\{\|x,z\|, \|y,z\|\})}{\|x,y\|}=1/2.\] 
So, $c=2$. By Theorem \ref{thm:Thm1}, for the distance $\delta$ between each pair in the Pareto front, we have that 
\[\delta \leq \frac{d_m}{c^{20}} = \frac{\sqrt{\pi^2+1}}{2^{19}}.\] 

Note that the pairs of test cases in the Pareto front are so close together that they are visually indistinguishable.
\qed
\end{Example}

\subsection{Directed Walk Strategy}

A variation of the random target strategy is to start with one test case (rather than a pair) and apply a unary datamorphism repeatedly until a test case of different classification is found. Then, the Pareto front between these two test cases is searched for in the same way as for the random target strategy. In this strategy, the unary datamorphism (i.e. a mutation operator) is the traversal method. The repeated application of the mutation operator makes a `walk' in one direction until a test case in a different class is found or too many iterations have been carried out and the exploration has gone too far.

\begin{algorithm}[h] 
\caption{(Directed Walk)}
\label{alg:DirectedWalk}
\begin{small}
\begin{algorithmic}
\Require 
      $TestSet$: test set; 
      $walkDistance$: integer; 
      $steps$: Integer; \\
      $d(x)$: Unary datamorphism;
      $mid(x,y)$: Binary datamorphism; 
\Ensure 
      $a$, $b$: Test Case; \\
\hspace{-0.25cm}{\textbf{Begin}}
\State 1: Select a test cases $x$ in $testSet$ at random; 
\State 2: Execute program $P$ on test case $x$;
\State 3: Walk in one direction as follows:    
\State \textbf{Bool} found = \textbf{false}; 
\For {$i \gets 1$ to $walkingDistance$} 
	\State $y=d(x)$;
	\State Execute software on test case $y$;   
	\If {($x.output \neq y.output$)} 
		 {$found$ = \textbf{true};  break; }
	\Else { $x=y$}; 
	\EndIf
\EndFor
\State 4: Check if a Pareto front can be found:
\If {($\neg found$)}  
	\Return $\left< null, null \right>$; 
\EndIf
\State 5: Refinement:
\For {$i \gets 1$ to $steps$} 
	\State $z = mid(x,y)$;
	\If {($x.output \neq z.ouptut$)} { y = z; }
	\Else { x = z;}
	\EndIf;
\EndFor
\State  $a = x$; $b = y$; 
\State \Return $\left< a ,b \right>$; \\
\hspace{-0.25cm}{\textbf{End}}
\end{algorithmic}
\end{small}
\end{algorithm}

Note that, a walk in one direction may not be able to find a data point in a different class. In that case, the algorithm returns $\left< null, null \right>$. Let $m, n>0$ be any given natural numbers. We write $DW(m,n)=\left< a,b \right>$ to denote the results of executing Algorithm \ref{alg:DirectedWalk} with $m$ as the walking distance and $n$ as the number of $steps$ and $\left< a, b \right>$ as the output. Assume that the exploratory test system satisfies assumption (\ref{eqn:Eqn1}) and has the following property. 

There is a constant $d_s >0$ such that 
\begin{eqnarray}
\forall x \in D.\left( \|x, d(x)\| \leq d_s \right). \label{eqn:Eqn3}
\end{eqnarray}
where  $d_{s}$ is called the step size of the traversal method $d(x)$. 
Then, we have the following correctness theorem for the directed walk algorithm. 

\begin{Theorem}\label{thm:Thm2}
If $DW(m,n)=\left< a, b \right> \neq \left< null, null \right>$ then $\left< a ,b \right>$ is a pair in the Pareto front according to $P$ with respect to $\|\cdot,\cdot\|$ and  $\delta$, if $d_s/c^n < \delta$, where $n$ is the number of steps. 
\end{Theorem}
\noindent{\emph{Proof.}}
If $DW(m,n)=\left<a, b \right> \neq \left< null, null \right>$, then the condition of the If-statement in step (4) is \textbf{false}. Thus, the For-loop of Step (5) is executed. It is easy to see that the For-loop in \emph{Step 5 Refinement} in the algorithm terminates. 

Similar to the proof of Theorem \ref{thm:Thm1}, by the definiton of $d_s$ and assumption (\ref{eqn:Eqn3}), the following is a loop invariant of the loop by induction on the number $i$ of iterations of the loop body. 
\[\|x,y\| \leq \frac{d_s}{c^i} \wedge P(x) \neq P(y).\] 

When the loop exits, $i = steps=n$. By Hoare logic, after executing the assignment statements $a=x$ and $b=y$, we have that 
\[\|a,b\| \leq d_s/c^n \wedge P(a) \neq P(b).\] 
Therefore, the theorem is true by Definition \ref{def:Pareto}. 
\qed

\begin{figure}[htbp]
	\centering
	\includegraphics[width=7.5cm]{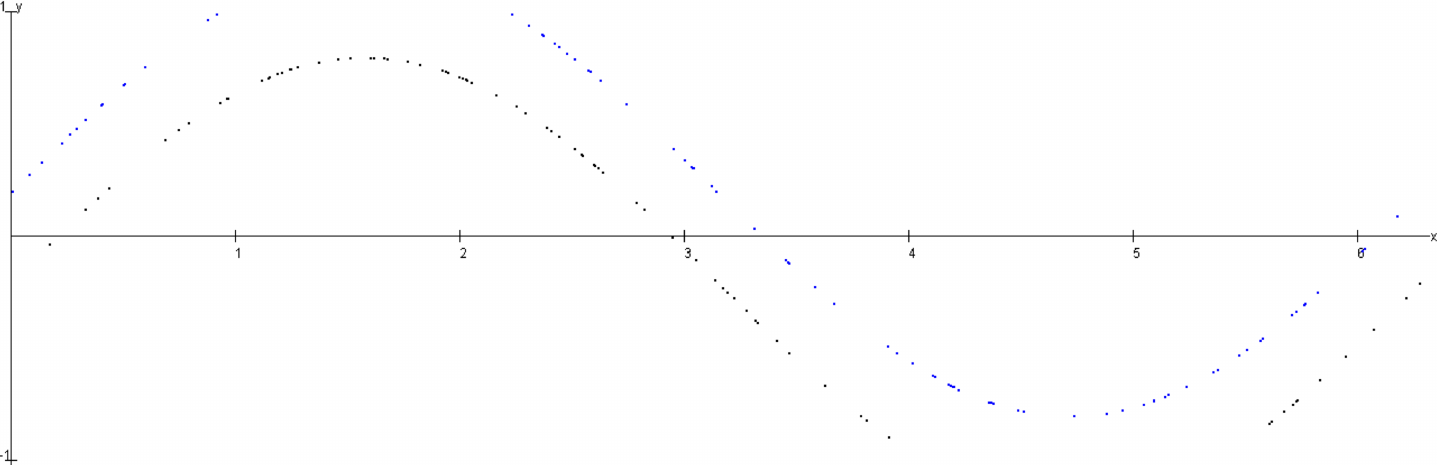}\\
	\caption{Pareto Fronts Generated by Directed Walk}
	\label{fig:DirectedWalkInputAndResult}
\end{figure}

\begin{Example}\label{exm:Exm2}
For example, starting from 1000 random test cases using the directed walk strategy with the $upward(x)$ datamorphism as the unary traversal method, a set of 161 Pareto front pairs were generated; shown in Figure \ref{fig:DirectedWalkInputAndResult}. The set of Pareto front pairs also shows clearly parts of the boundaries between classes. The success rate of finding a pair of Pareto front on one test case is 16.1\%. 

In this example, the number $n$ of steps is also 20. By the definition of $upward(x)$ traversal method, we have that $d_s=0.2$, if the distance function $\|x,y\|$ is $Eucl(x,y)$. As in Example \ref{exm:Exm1}, by the definition of $mid(x,y)$, we have that $c=2$. By Theorem \ref{thm:Thm2}, for the distance $\delta$ between each Pareto front pair, we have that 
\[\delta \leq \frac{d_s}{c^{20}} = 0.2 \times \frac{1}{2^{20}}.\] 
Again, the distance between the test cases in each Pareto front pair is so small that they are not visually distinguishable, so they appear as one dot in Figure \ref{fig:DirectedWalkInputAndResult}.
\qed
\end{Example}

\subsection{Random Walk Strategy}

If multiple traversal methods are available, a random walk can be performed by selecting the direction of the next step at random. This is similar to the random walk testing in a web GUI hyperlink test. The algorithm is given below. 

\begin{algorithm}[h] 
\caption{(Random Walk Strategy)}
\label{alg:RandomWalk}
\begin{small}
\begin{algorithmic}
\Require  
      $testSet$: Test Set; 
      $walkingDistance$: Integer; 
      $steps$: Integer; \\
      $d_1(x),\cdots,d_k(x)$: Unary datamorphism ($k>1$); 
      $mid(x,y)$: Binary datamorphism; 
\Ensure 
      $a, b$: Test Case; \\
\hspace{-0.25cm}\textbf{Begin}
	\State 1: Select a test case $x$ in $testSet$ at random;
	\State 2: Execute program $P$ on test case $x$;
	\State 3: Walking at random to search for test case in a different class:
	\State \textbf{Bool} $found$ = \textbf{false}; 
	\For {$i \gets 1$ to $walkingDistance$} 
		\State Get a random integer $r$ in the range $[1,k]$
		\State $y=d_r (x)$;
		\State Execute program $P$ on test case $y$;       
		\If {($x.output \neq y.output$)}
			{$found$ = \textbf{true};  \textbf{break};}
		\Else { x=y;} 
		\EndIf
    \EndFor
    \State 4: Check if a Pareto front can be found:
	\If {($\neg found$)} { \Return $\left<null, null \right>$; } 
	\EndIf
	\State 5: Refinement:
	\For {$i \gets 1$ to $steps$}
		\State $z = mid(x,y)$;
		\If {($x.output \neq z.ouptut$)} { $y = z$; }
		\Else { x = z; }
  		\EndIf
	\EndFor
	\State $a = x$; $b = y$;
	\State \Return $\left<a,b \right>$;  \\
\hspace{-0.25cm}{\textbf{End}}
\end{algorithmic}
\end{small}
\end{algorithm}

We write $RW(m,n)=\left< a,b \right>$ to denote the results of executing Algorithm \ref{alg:RandomWalk} with $m$ as the walking distance and $n$ as the $steps$ and $\left< a, b \right>$ as the output.
Assume that the exploratory test system satisfies assumption (\ref{eqn:Eqn1}) and has the following property. 
There is a constant $d_{sm} >0$ such that 
\begin{eqnarray}
\forall x \in D. \forall d_i \in W. (\|x, d_i(x)\| \leq d_{sm}). \label{eqn:Eqn4}
\end{eqnarray}
where  $d_{sm}$ is called the maximal step size of the traversal methods $d_i(x) \in W$. 
Then, we have the following correctness theorem for the algorithm of random walk strategy.

\begin{Theorem}\label{thm:Thm3}
If $RW(m,n)=\left< a, b \right> \neq \left< null, null \right>$ then $\left< a ,b \right>$ is a Pareto front pair according to $P$ with respect to $\|\cdot,\cdot\|$ and  $\delta$, if $d_{sm}/c^n < \delta$, where $n$ is the number of steps. 
\end{Theorem}
\noindent{\emph{Proof.}}
If $RW(m,n)=\left<a, b \right> \neq \left< null, null \right>$ then the condition of the If-statement in step (4) is \textbf{false}. Thus, the For-loop of Step (5) is executed. It is easy to see that the For-loop in \emph{Step 5 Refinement} in the algorithm terminates. 

Similar to the proof of Theorem \ref{thm:Thm1}, by the definition of $d_{sm}$ and assumption (\ref{eqn:Eqn4}), we can prove that the following is a loop invariant of the loop by induction on the number $i$ of iterations of the loop body. 
\[\|x,y\| \leq \frac{d_{sm}}{c^i} \wedge P(x) \neq P(y).\] 

When the loop exits, $i = steps=n$. After executing the assignment statements $a=x$ and $b=y$, the following is true by Hoare logic. 
\[\|a,b\| \leq d_{sm}/c^n \wedge P(a) \neq P(b).\] 
Therefore, the theorem is true by Definition \ref{def:Pareto}. 
\qed

\begin{Example}\label{exm:Exm3}
For example, by applying the random walk strategy on a test set containing 300 random test cases, 1000 random walks generated 805 pairs of Pareto front test cases, as shown in Figure \ref{fig:RandomWalkResult}, where the walking distance was 20 steps. 
 
\begin{figure}[htbp]
	\centering
	\includegraphics[width=7.5cm]{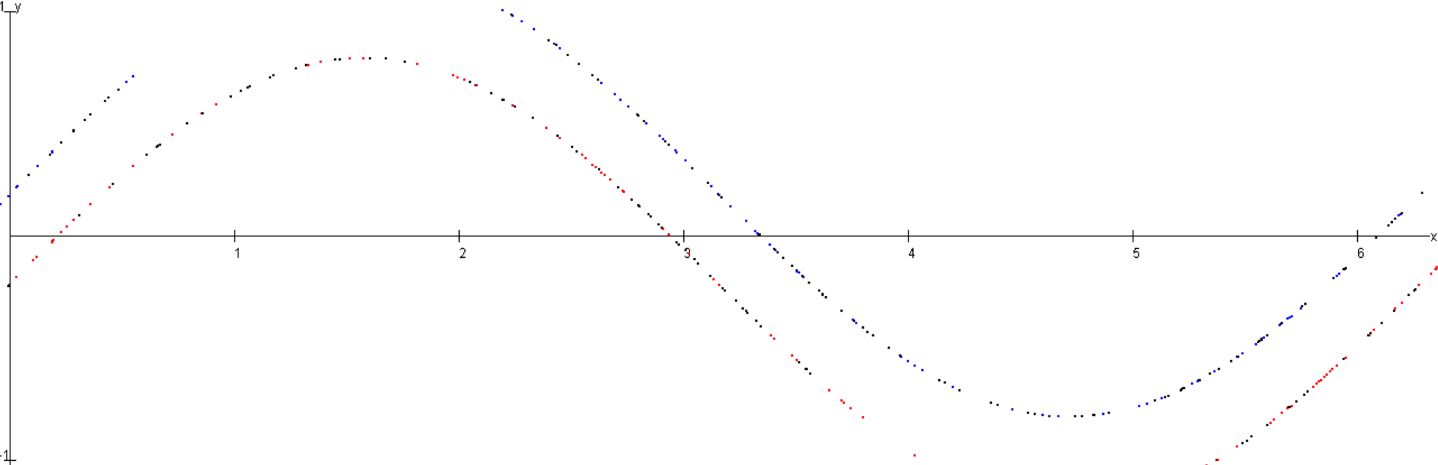}\\
	\caption{The Pareto Fronts Generated by Random Walk}
	\label{fig:RandomWalkResult}
\end{figure}

In this example, the number $n$ of steps is also 20. By the definition of $upward(x)$, $downward(x)$, $leftward(x)$ and $rightward(x)$ traversal methods, we have that $d_s=0.2$, if the distance function $\|x,y\|$ is $Eucl(x,y)$. As in Example \ref{exm:Exm1} and \ref{exm:Exm2}, by the definition of $mid(x,y)$, we have that $c=2$. By Theorem \ref{thm:Thm3}, the distance $\delta$ between each pair in the Pareto front satisfies the following inequality.  

\[\delta \leq \frac{d_s}{c^{20}} = 0.2 \times \frac{1}{2^{20}}. \] 

\end{Example}

\section{Empirical Evaluation}\label{sec:Evaluation}

We have conducted empirical evaluations of the proposed test strategies to determine their practical applicability for detecting borders between subdomains. In particular, we answer the following two research questions:

\begin{itemize}
\item \emph{RQ1: Capability}. Are the exploratory strategies \emph{capable} of discovering the borders between subdomains? 
\item \emph{RQ2: Cost}. Are the exploratory strategies \emph{costly} for discovering the borders between subdomains? 
\end{itemize}

Capability is the probability of a test strategy returning a Pareto front pair when executed. The expected size of a Pareto front set produced by a strategy can then be calculated as $C_m \times W$ pairs, where $C_m$ is the strategy's capability for testing classifier $m$ and $W$ is the number of invocations of the strategy, called the \emph{number of walks} in the sequel.

Cost is related to the amount of computational resources needed to find a Pareto pair. We measure the cost using the average number of test executions of the classifier for discovering each Pareto pair, since the specific time and storage space depends on the classifier. Note that the strategies do not require manual labelling of the test cases or any form of test oracle. Therefore, the time taken to complete the testing process can be estimated as 
\begin{equation}
Time = W \times C_m \times E_m \times s \label{equ:Time}
\end{equation}
where $C_m$ and $E_m$ denotes the strategy's capability and cost for testing the model $m$, and $W$ is the number of walks and $s$ the average time taken by each invocation of the classifier. 

We have conducted two empirical evaluations of the proposed test strategies. The first is a set of controlled experiments with 10 hand-coded classifiers on two-dimensional continuous numerical features. The second is a set of case studies with 16 machine learning models built by training on three real-world datasets. Both evaluations were conducted using the automated datamorphic testing tool Morphy. The raw data collected, source code of the test systems, test scripts, etc. are all available on GitHub repository together with the executable code of the automated testing tool Morphy for download.
\footnote{The URL of the GitHub repository is https://github.com/hongzhu6129/ExploratoryTestAI.git\label{ftn:GitHubURL}} 
Summary data can be found in the Appendix. This section reports the results of these empirical studies. 

\subsection{Controlled Experiments}\label{sec:Experiments}

%\subsection{Design of the Experiments}

\subsubsection{Design and Conduct of the Experiments}

The goal of the controlled experiments is to study the factors that affect the cost and capability of these test strategies in finding Pareto front pairs between subdomains. In doing so, we demonstrate that Pareto front pairs can represent borders between subdomains; the aim is not to compare the strategies, however.

\begin{figure}[htbp]
\centering

\includegraphics[width=4cm]{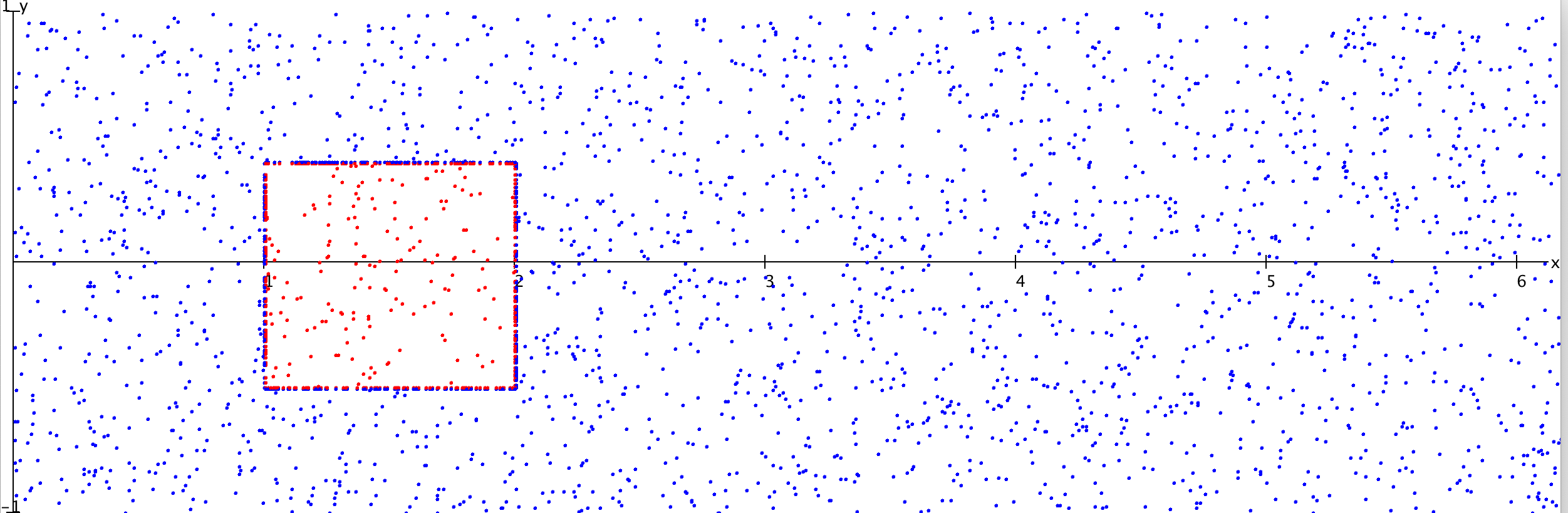} \hspace{0.5cm}
\includegraphics[width=4cm]{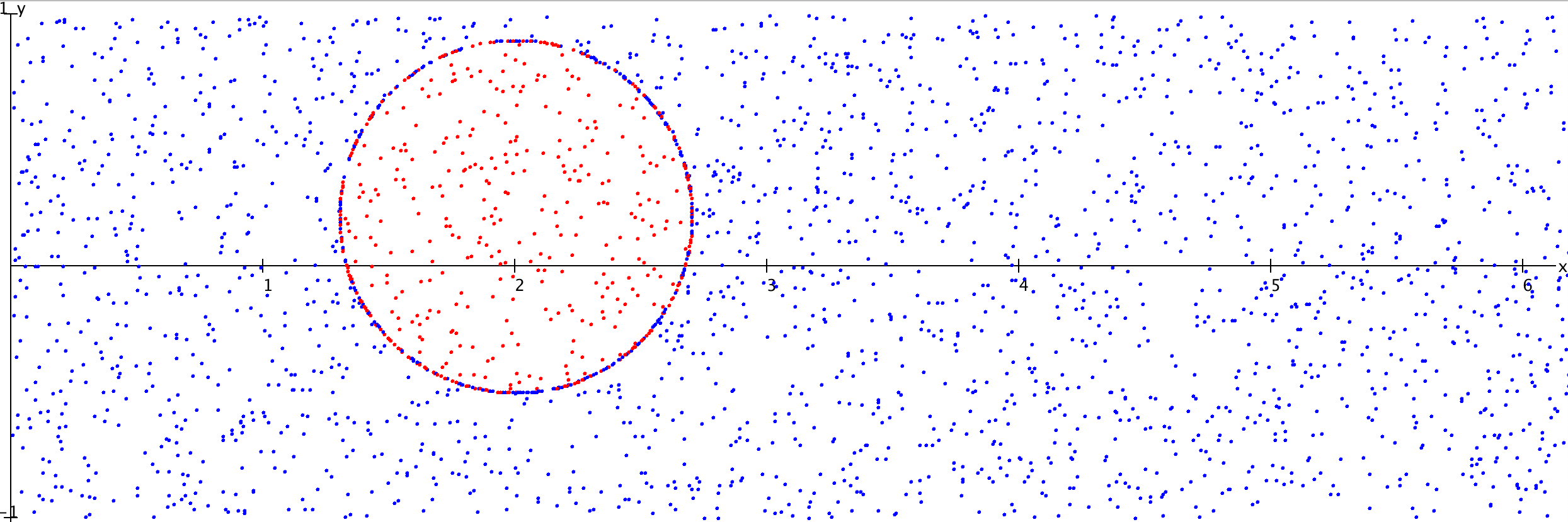} \hspace{0.5cm}
\includegraphics[width=4cm]{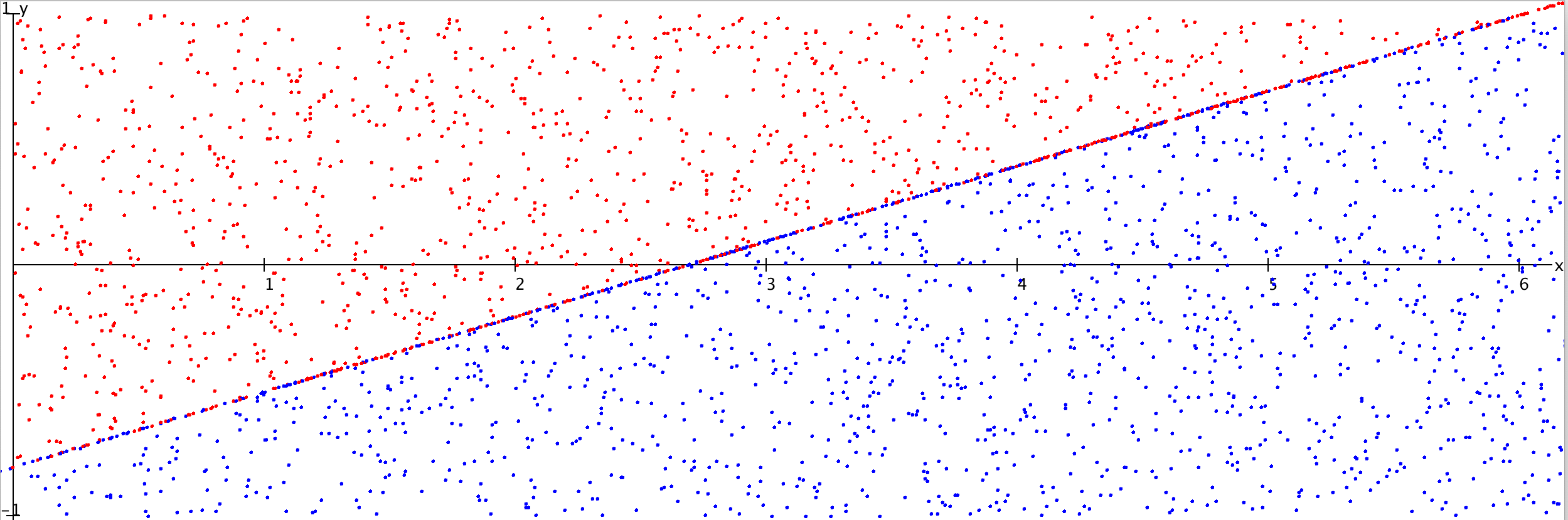}\\
\scriptsize{(1) Box 1} \hspace{3.5cm} \scriptsize{(3) Circle 1} \hspace{3.5cm} \scriptsize{(5) Line 1} \\

\includegraphics[width=4cm]{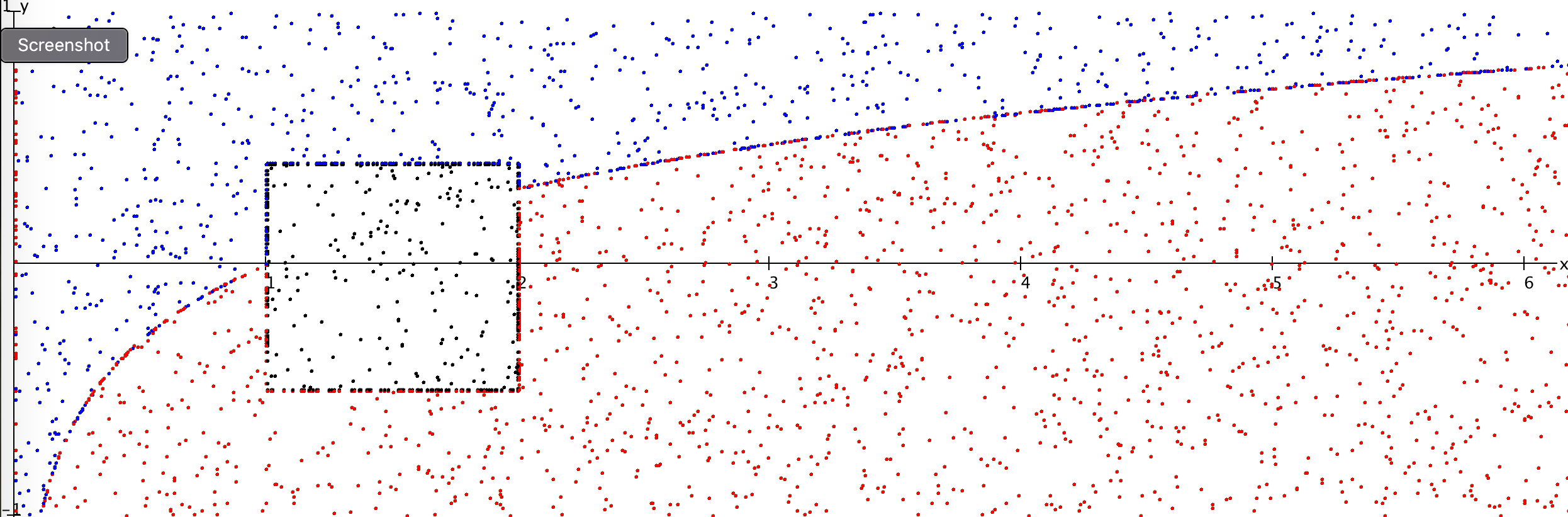} \hspace{0.5cm}
\includegraphics[width=4cm]{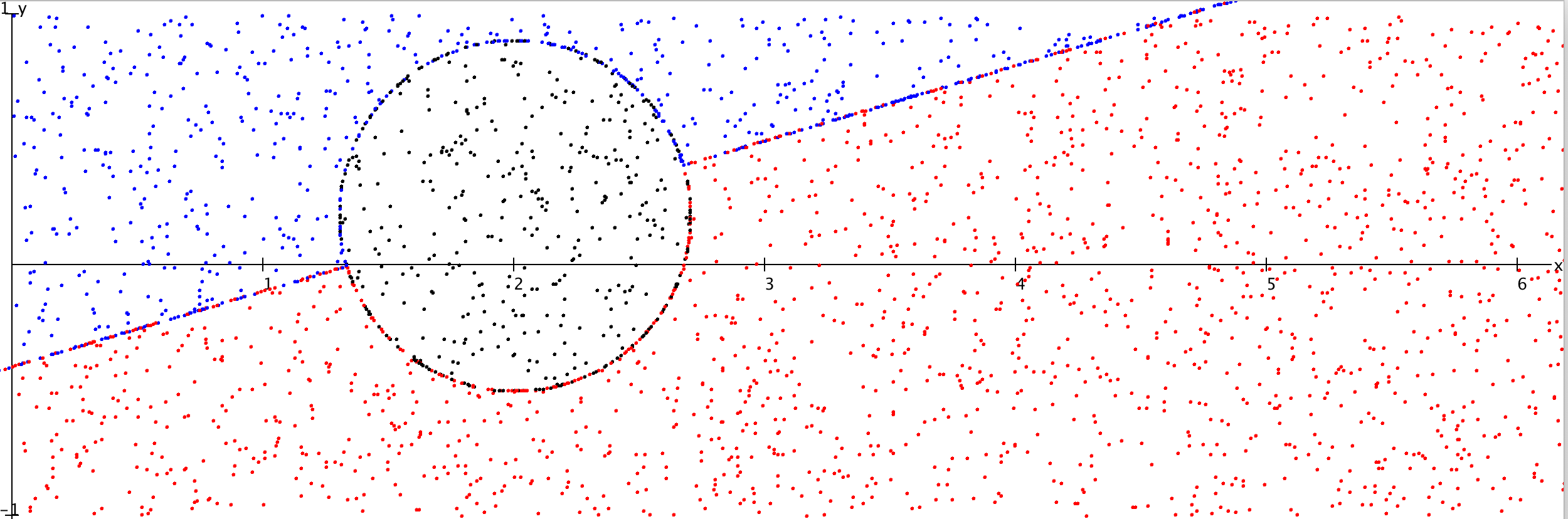} \hspace{0.5cm}
\includegraphics[width=4cm]{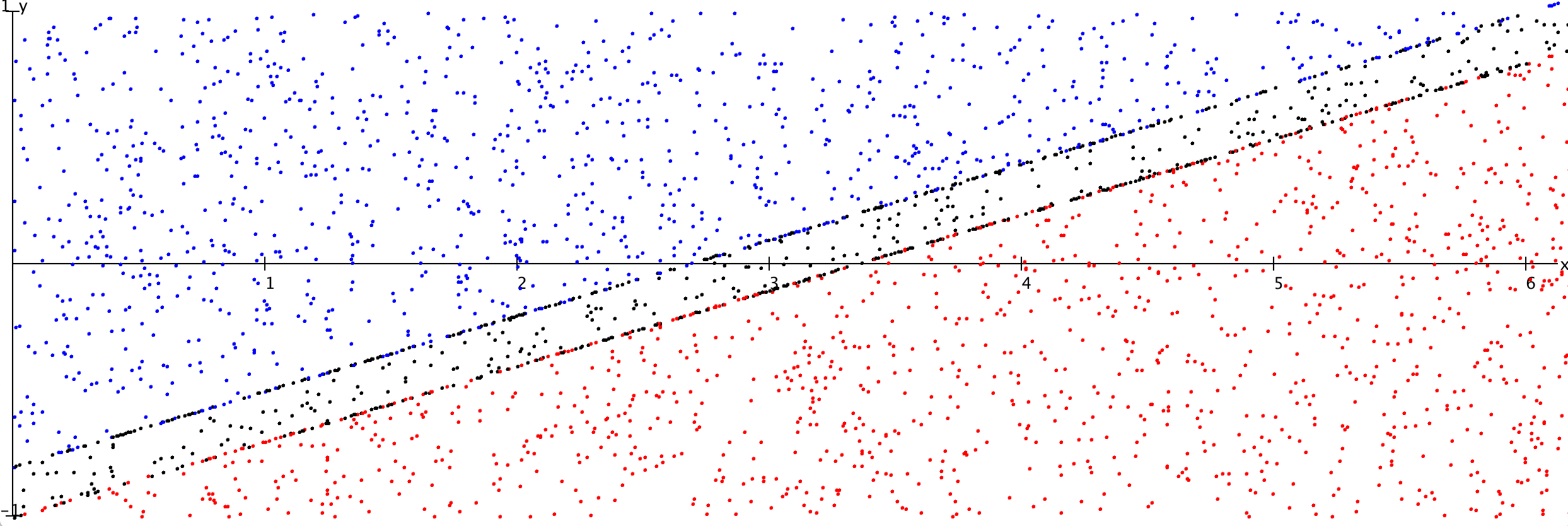}\\
\scriptsize{(2) Box 2} \hspace{3.5cm} \scriptsize{(4) Circle 2} \hspace{3.5cm} \scriptsize{(6) Line 2}\\

\includegraphics[width=4cm]{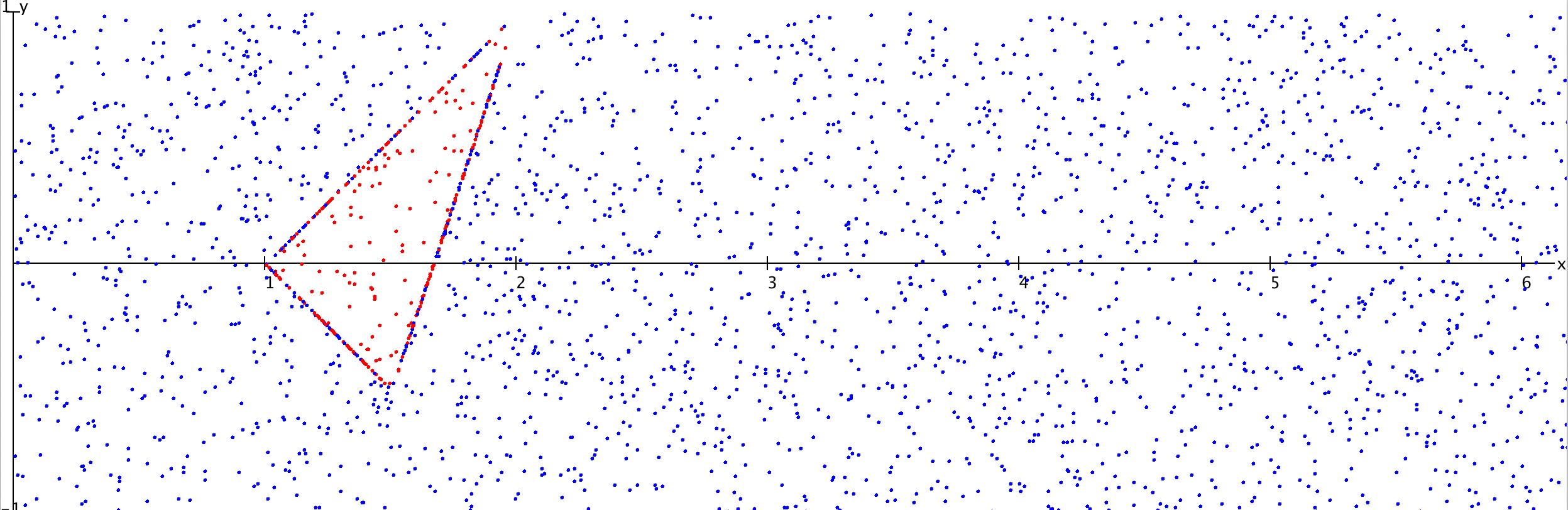} \hspace{1cm}
\includegraphics[width=4cm]{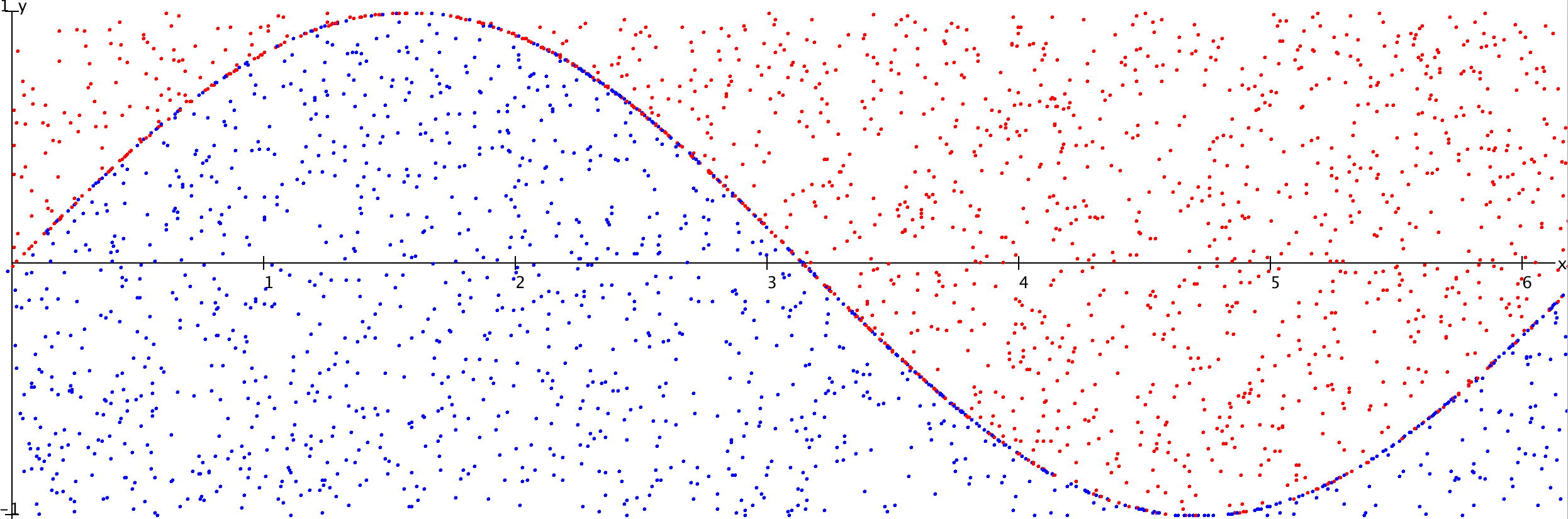}\\
\scriptsize{(7) Triangle 1} \hspace{4cm} \scriptsize{(9) Sin 1} \\

\includegraphics[width=4cm]{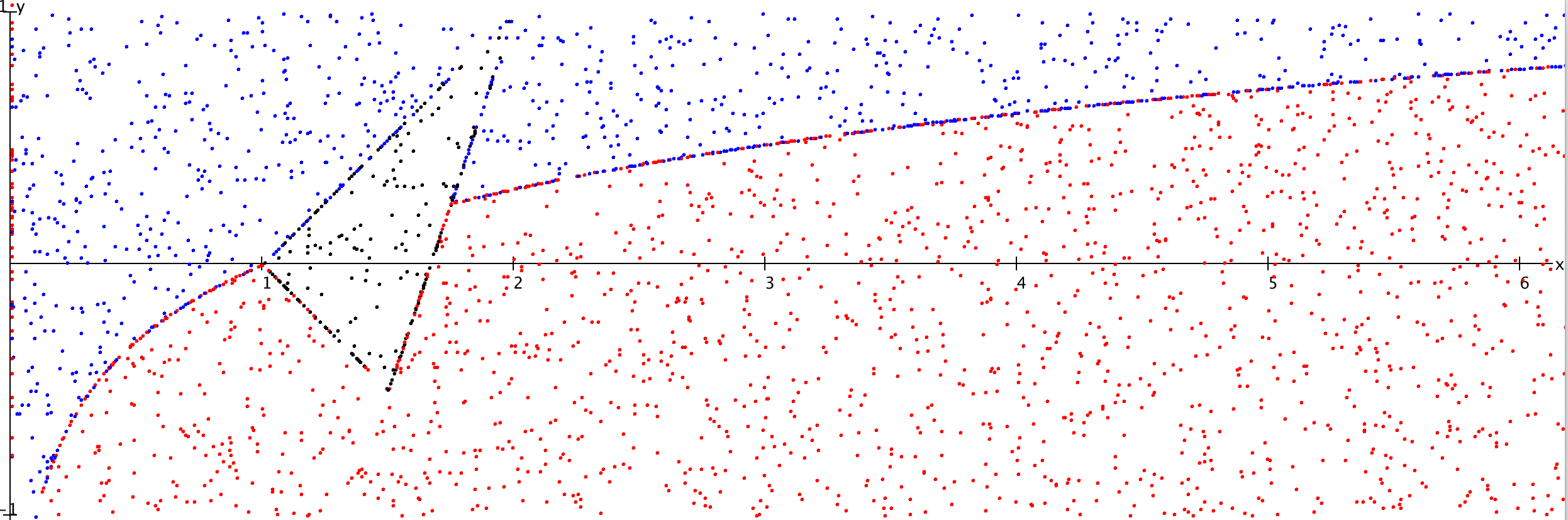} \hspace{1cm}
\includegraphics[width=4cm]{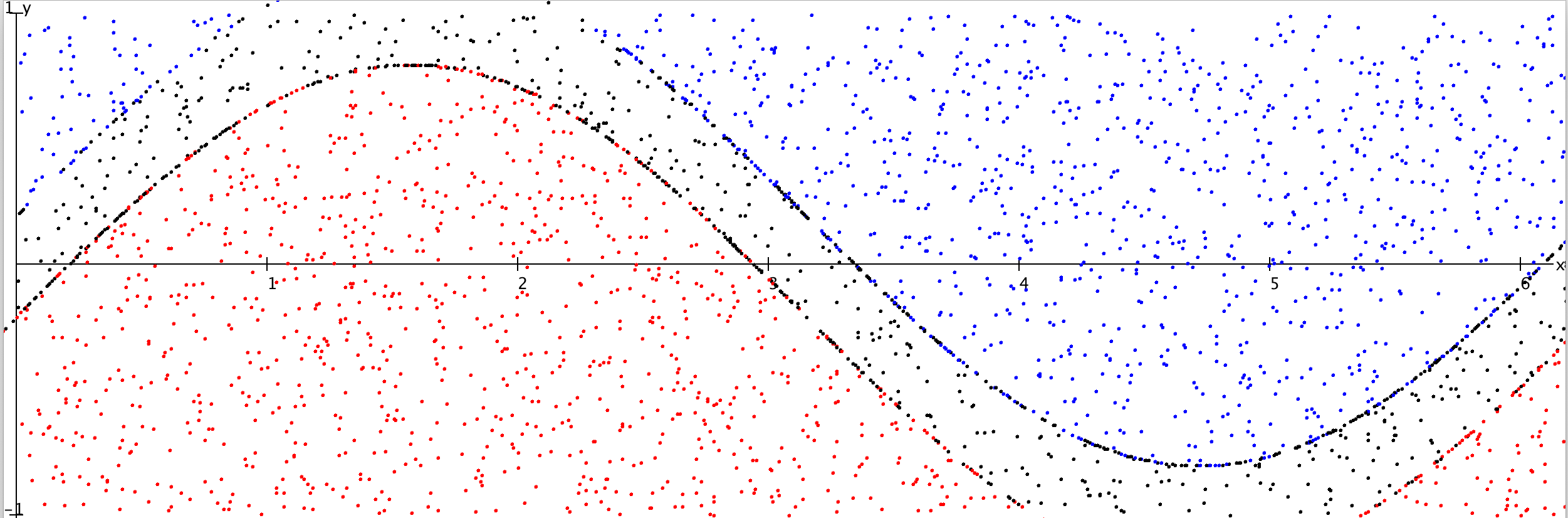}\\
\scriptsize{(8) Triangle 2} \hspace{4cm} \scriptsize{(10) Sin 2}
\caption{Illustration of the sample applications}
\label{fig:SampleApplications}
\end{figure}

The experiments are carried out with the ten classifiers shown in Figure \ref{fig:SampleApplications}. These classifiers are all on the same input domain of two-dimensional real numbers in the range of $[0,2\pi]\times [-1,1]$. As shown in Figure \ref{fig:SampleApplications}, they are continuous numerical feature based classifiers. 
   
The choice of the subjects enables us to visually display the Pareto fronts obtained from executing the test strategies so that we can verify the results against the theoretical borders between the subdomains. This has been done visually for a large number of random samples taken from the Pareto fronts and all have been found to be correct. For example, Figure \ref{fig:SampleApplications} shows some example screen snapshots of the visualisations of these test results. Each figure contains both the random test cases from which the starting points were selected and the test cases generated through testing. Figures \ref{fig:AimedWalkResult}, \ref{fig:DirectedWalkInputAndResult} and \ref{fig:RandomWalkResult} contain only the latter. 

In addition to the visual validation of the outputs of the tests, the strategies are executed repeatedly 10 times for each number of walks. The number of executions of the classifiers and the number of mutants generated were collected for statistical analysis of the capability and cost of the strategies. The following subsections reports this analysis. 

\subsubsection{Main Results}

\begin{itemize}
\item \emph{Results of experiments with the directed walk strategy}
\end{itemize}

The controlled experiments on the directed walk strategy consisted of randomly selecting a number of test cases from the uniform distribution and walking 20 steps in one direction using the upward datamorphism. Both the average number of test executions of the subject program under test and the average number of mutant test cases generated (i.e. the number of Pareto front pairs) are recorded.

\begin{figure}[htbp]
	\centering
	\includegraphics[width=5cm,height=2.5cm]{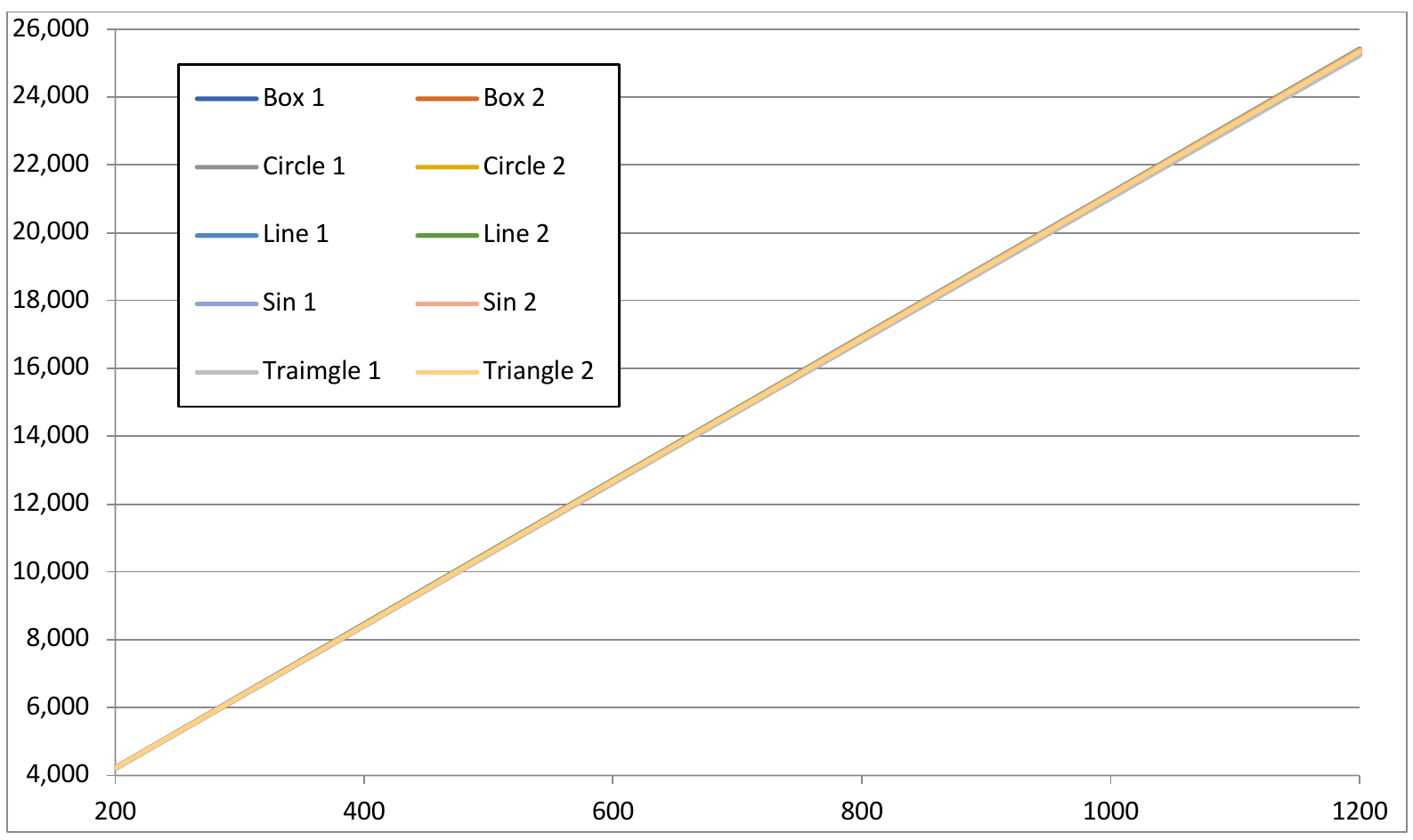}
	\includegraphics[width=5cm,height=2.5cm]{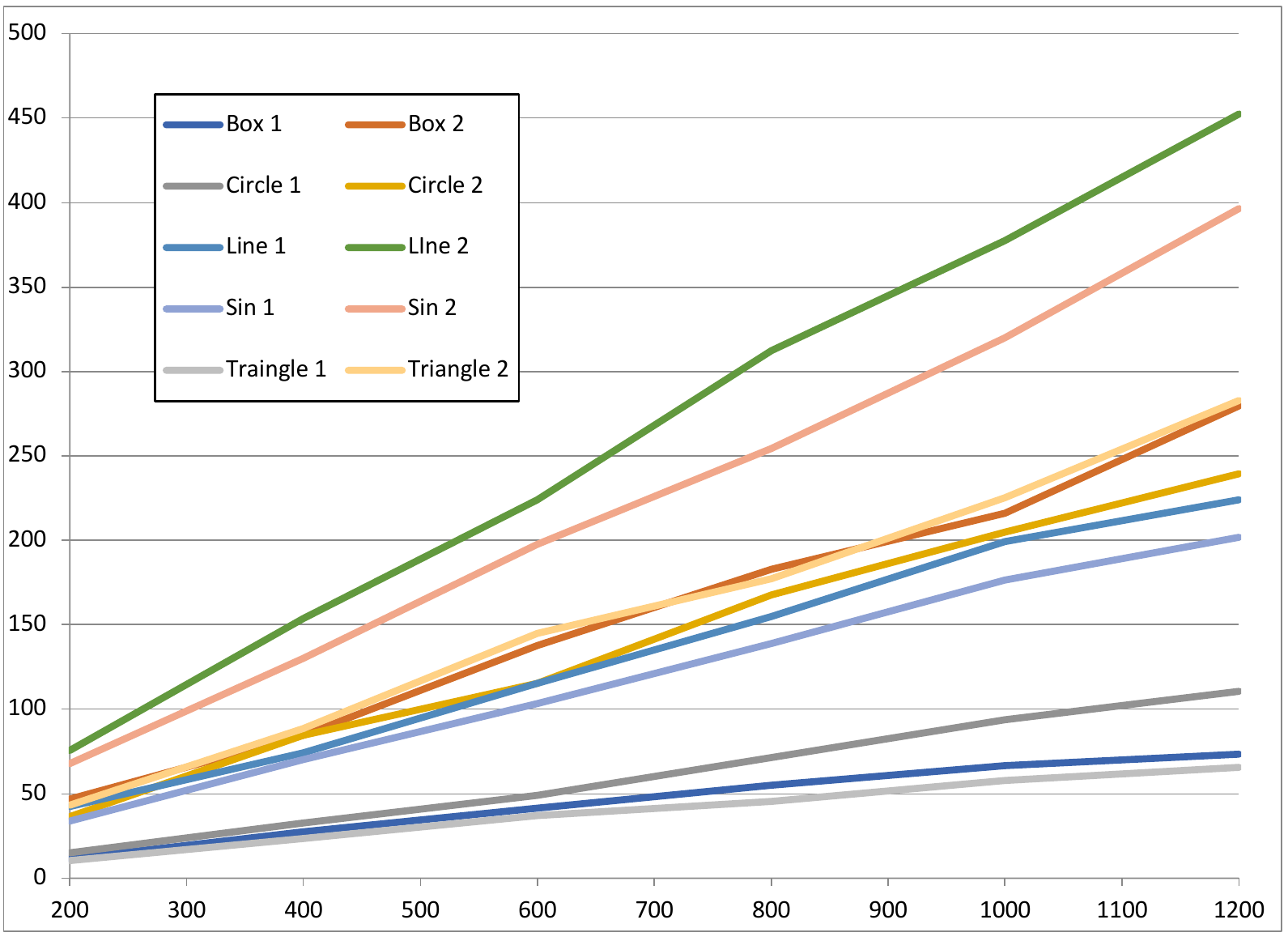}\\
	\scriptsize{(a) Average Number of Executions  \hspace{2cm} (b) Average Number of Mutants} \\
	\includegraphics[width=5cm,height=2.5cm]{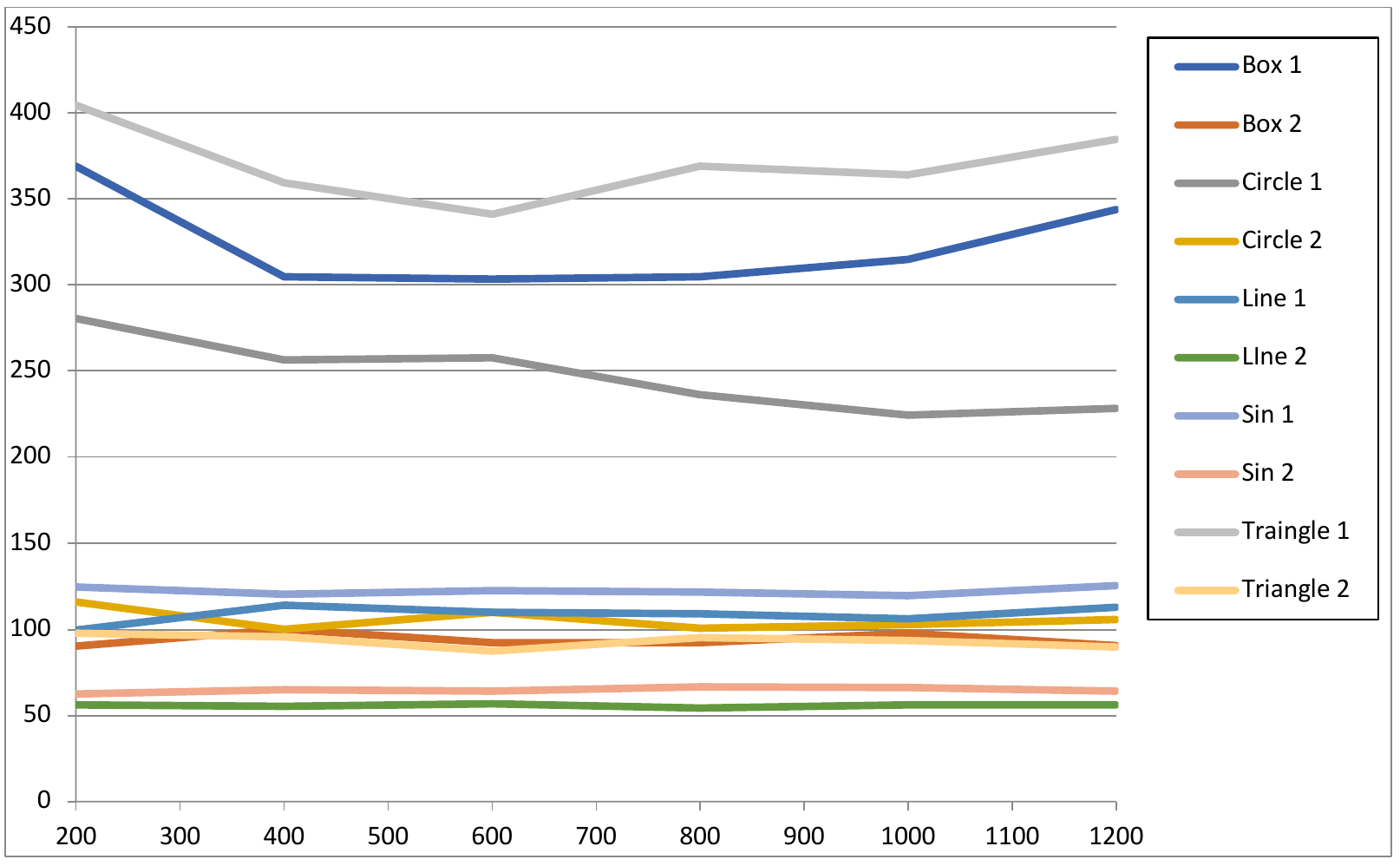}
	\includegraphics[width=5cm,height=2.5cm]{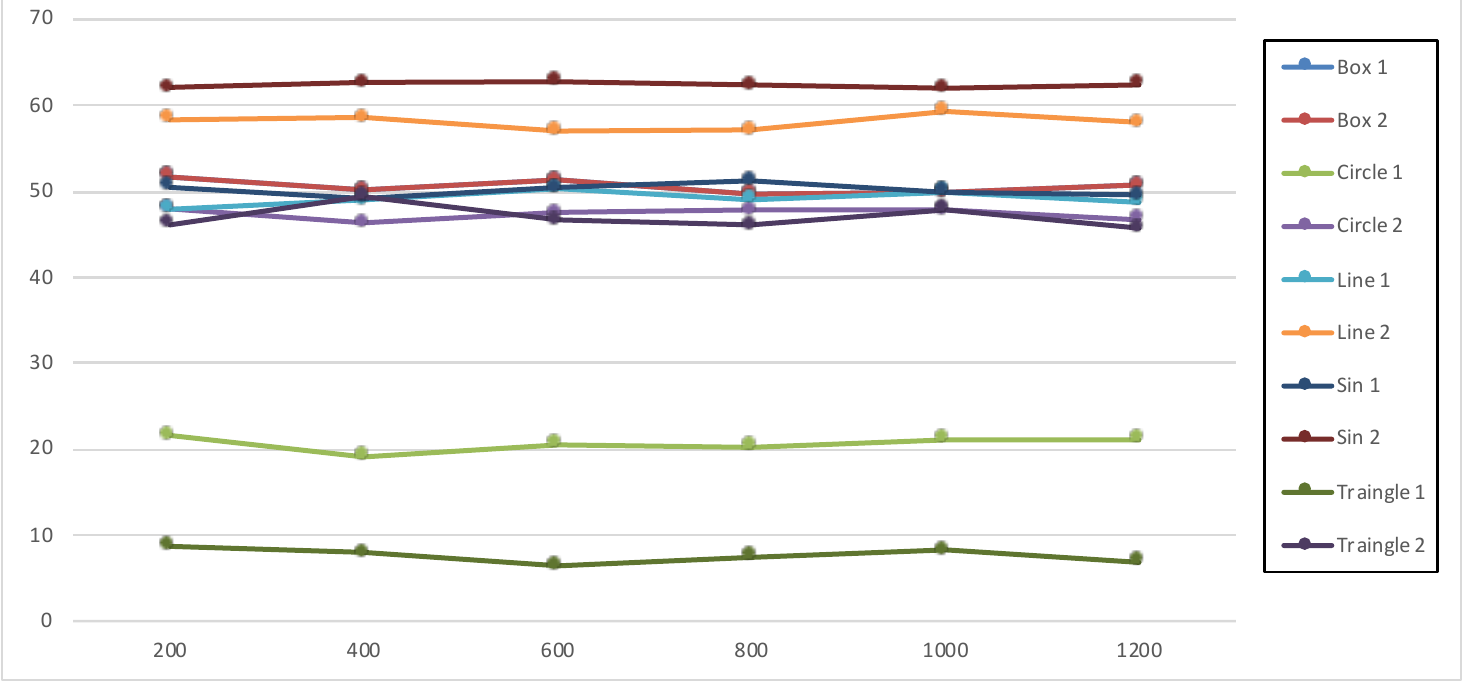} \\
	\scriptsize {(c) Average Cost \hspace{2.7cm} (d) Average Capability}\\
	\caption{Results of The Directed Walk Strategy with Variable Number of Walks}
	\label{fig:DirectedWalkResultChart}
\end{figure}

The experimental data shows that the number of mutant test cases generated with the directed walk strategy increases linearly with the number of walks; see Figure \ref{fig:DirectedWalkResultChart}. Similarly, the number of test executions is also linear with respect to the number of walks. In Figure \ref{fig:DirectedWalkResultChart}, the x-axis is the number of random seed test cases, which equals the number of walks, and the y-axes of (a) and (b) are the average numbers of test executions and mutant test cases, respectively. In (a), the average numbers of test executions on various subject programs are so close to each other that they are not visually separable. The y-axis of (c) measures the average cost as the number of test executions per test case in the generated Pareto front. We can see that this is fairly invariant for each subject as the former ranges from 200 to 1200. Similarly, (d) shows the average capability remains invariant when the number of walks increases. 

\begin{itemize}
\item \emph{Results of experiments with the random walk strategy}
\end{itemize}

\begin{figure}[htbp]
	\centering
	\includegraphics[width=5cm, height=2.5cm]{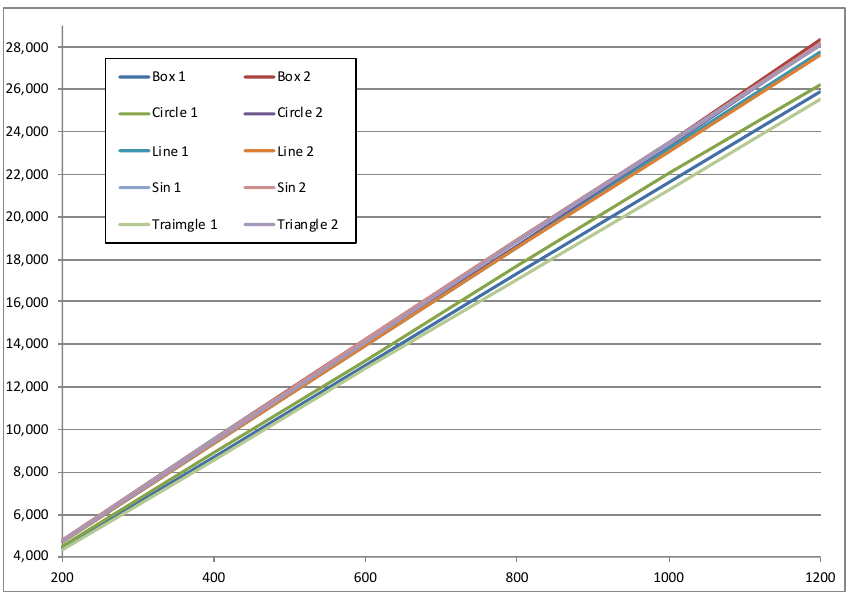}
	\includegraphics[width=5cm, height=2.5cm]{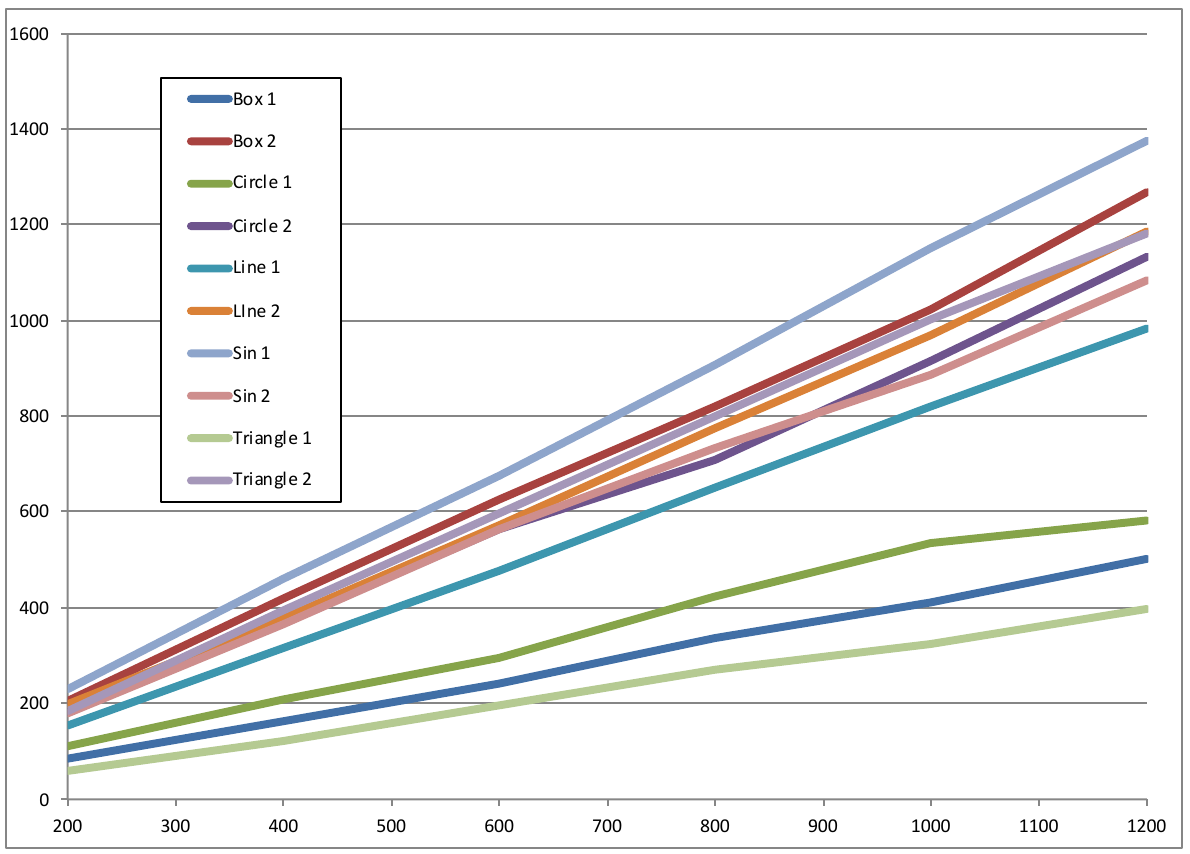}\\
		\scriptsize{(a) Average Number of Executions \hspace{2cm} (b) Average Number of Mutants}\\
	\includegraphics[width=5cm, height=2.5cm]{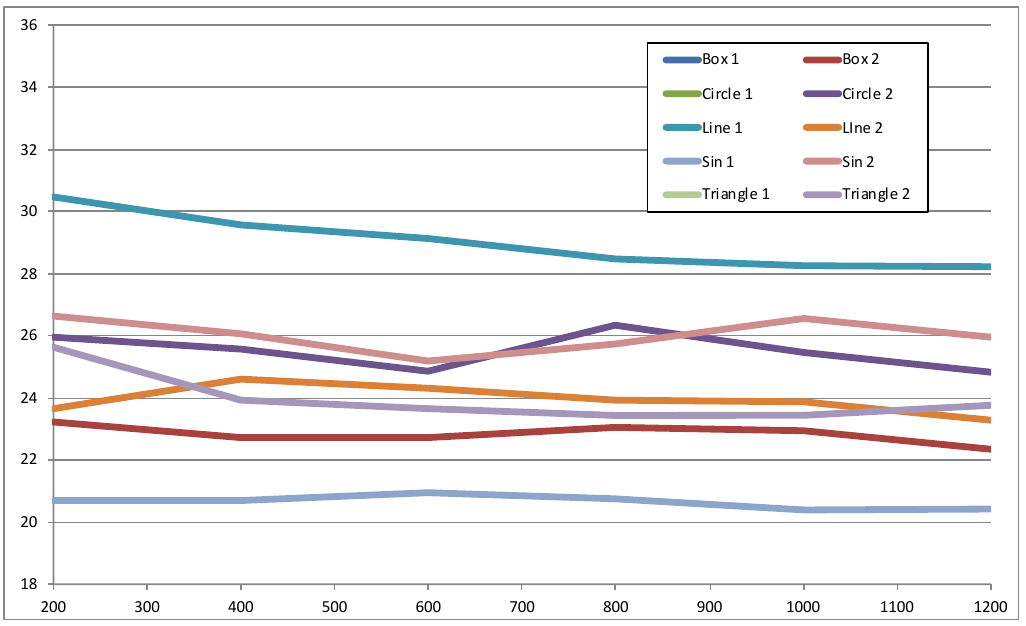}
	\includegraphics[width=5cm, height=2.5cm]{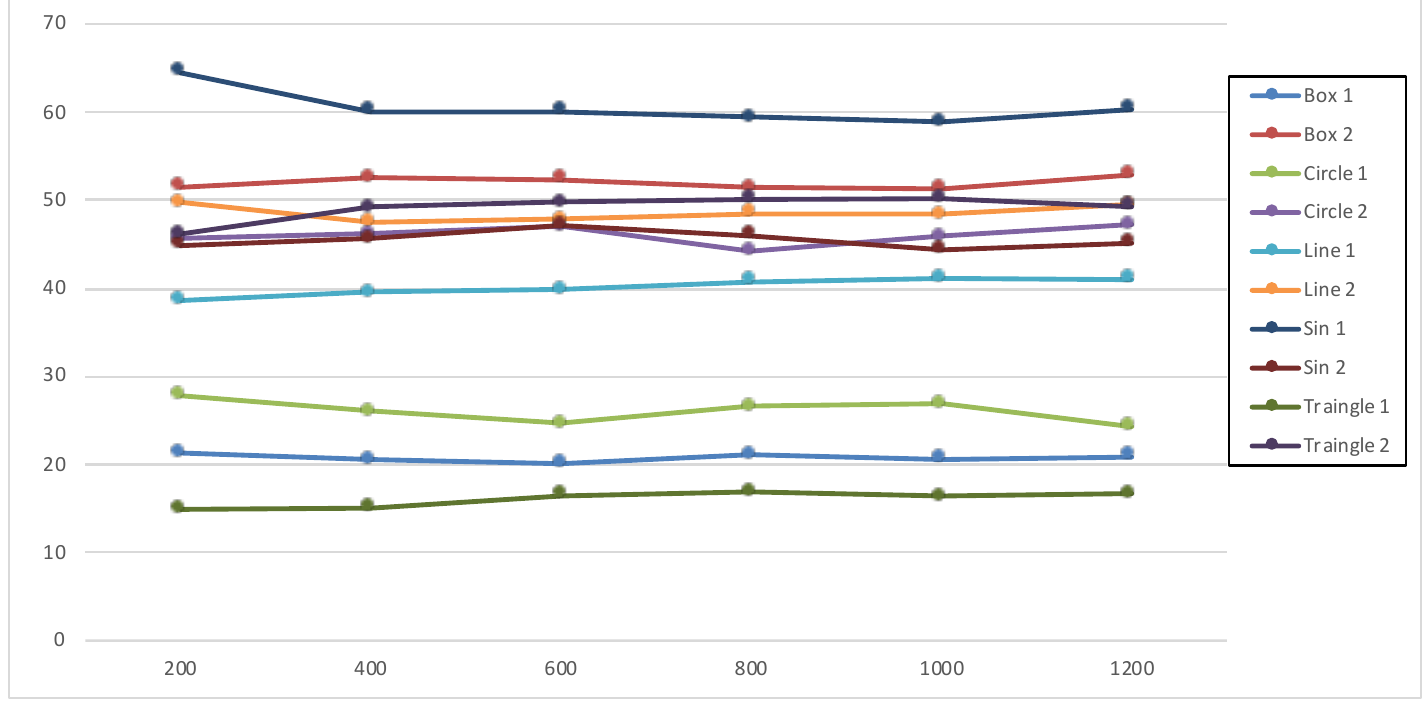}\\
	\scriptsize{(c) Average Cost  \hspace{3cm} (d) Average Capability} \\ 
	\caption{Results of The Random Walk Strategy with Variable Numbers of Walks}
	\label{fig:RandomWalkResultChart}
\end{figure}

The random walk strategy is parameterised by the number of seed test cases and the number of walks starting from them. So, we fix the first parameter at 200 seeds and vary the number of walks, and then we fix the second parameter at 800 walks and vary the number of seeds. Figure \ref{fig:RandomWalkResultChart} shows the results of the first set of experiments with the random walk strategy. Figure \ref{fig:RandomWalkResultChart}(a) and (b) clearly shows that the number of runs and the size of Pareto fronts increase linearly with the number of walks, while the cost and capability remains mostly invariant as shown in (c) and (d). 

\begin{figure}[htbp]
	\centering
	\includegraphics[width=5cm, height=2.5cm]{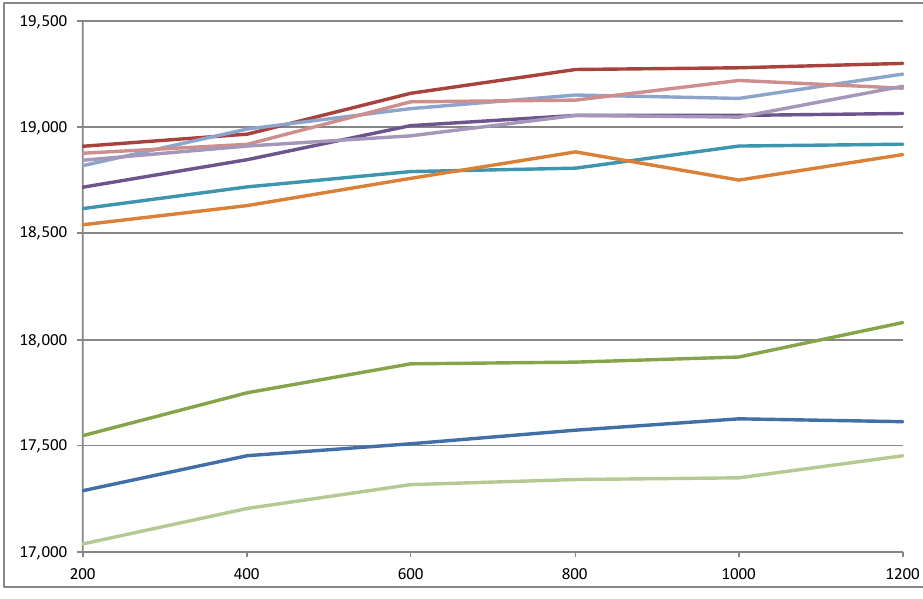}
	\includegraphics[width=5cm, height=2.5cm]{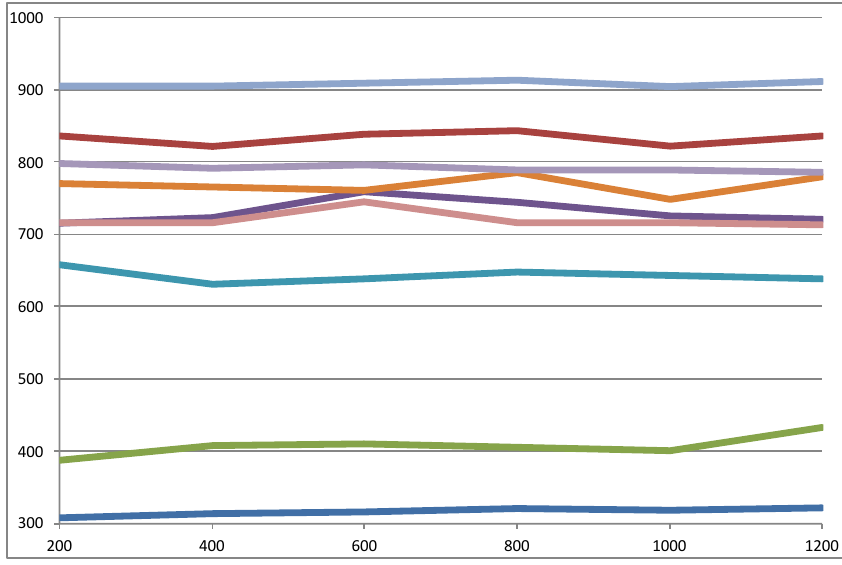}\\
	\scriptsize{(a) Average Number of Runs \hspace{2cm} (b) Average Number of Mutants} \\
	\includegraphics[width=5cm, height=2.5cm]{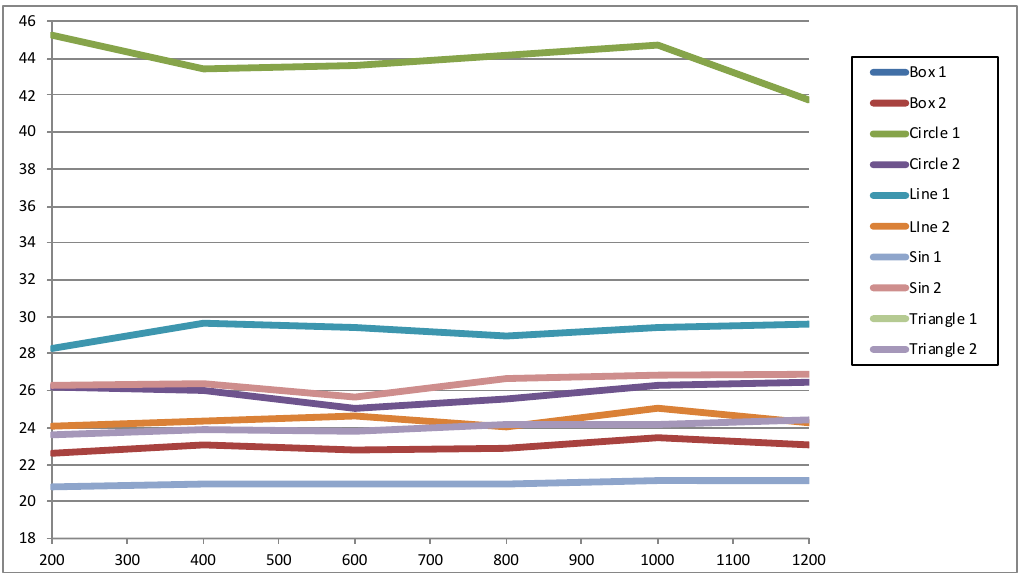}			\includegraphics[width=5cm, height=2.5cm]{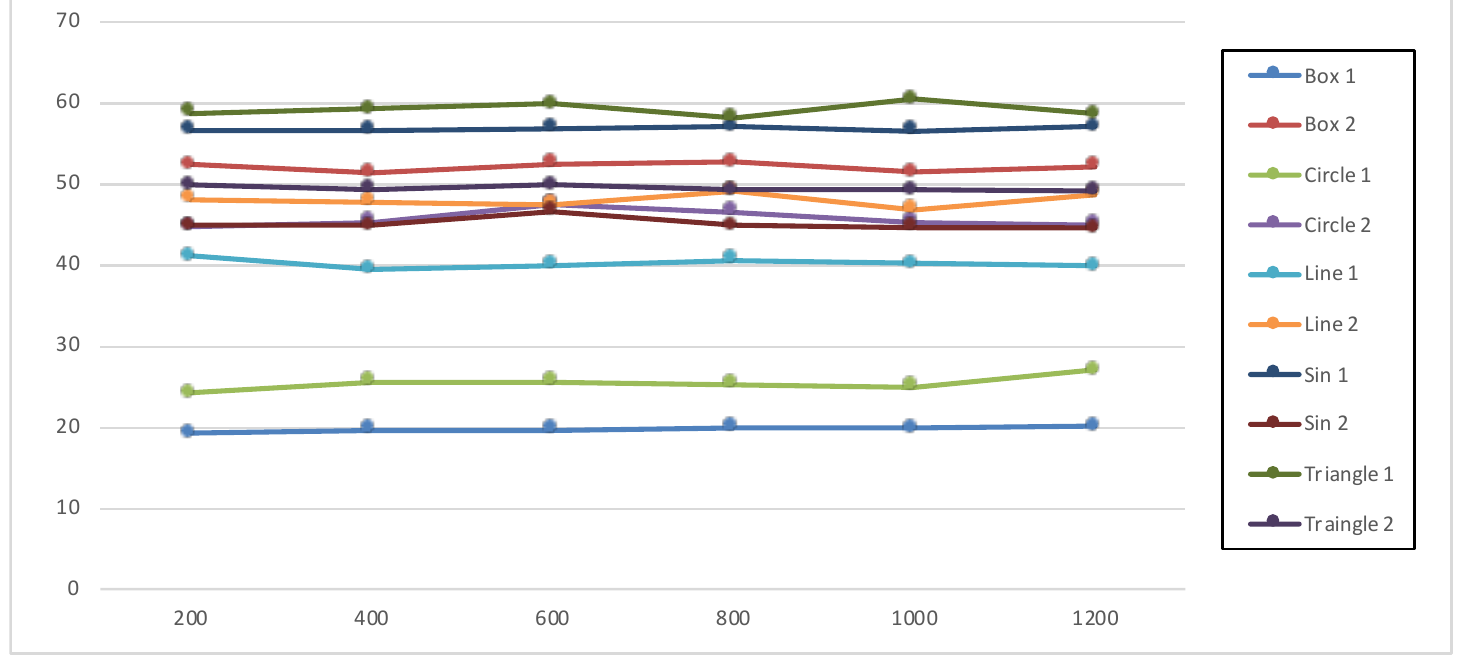}\\
	\scriptsize{(c) Average Cost \hspace{3cm} (d) Average Capability} 
	\caption{Results of the Random Walk Strategy with Variable Number of Seeds}
	\label{fig:RandomWalkTCResultChart}
\end{figure}

Similarly, Figure \ref{fig:RandomWalkTCResultChart}(a) and (b) shows that the number of runs increases slightly as the number of seed test cases increases, while the size of generated Pareto front remains almost invariant. Moreover, the cost and the capability remain almost invariant as the number of seed test cases increases as shown in Figure \ref{fig:RandomWalkTCResultChart}(c) and (d), respectively.  

\begin{itemize}
\item \emph{Results of the experiments with the random target strategy}
\end{itemize}

\begin{figure}[htbp]
	\centering
	\includegraphics[width=5cm, height=2.5cm]{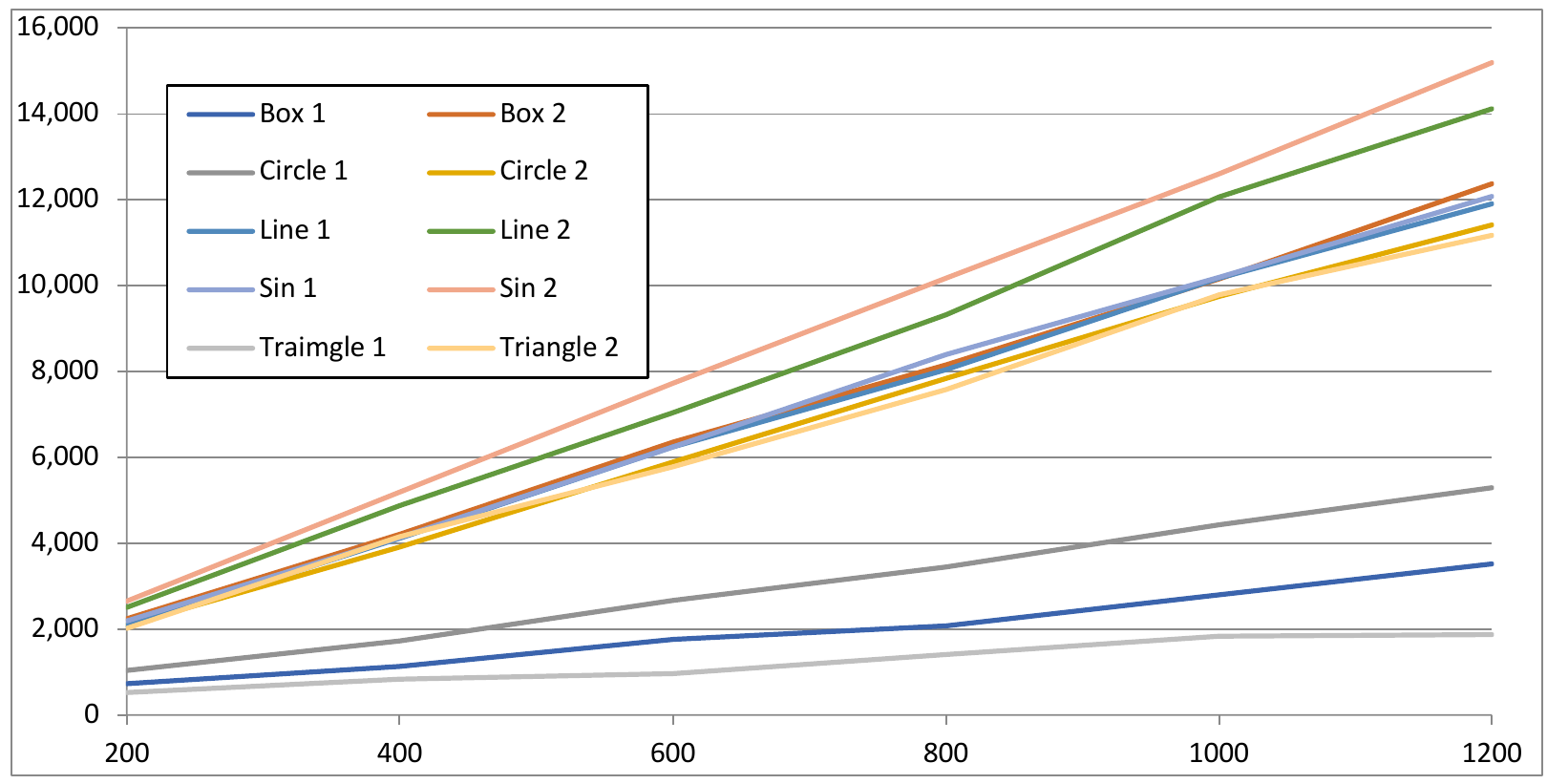}
	\includegraphics[width=5cm, height=2.5cm]{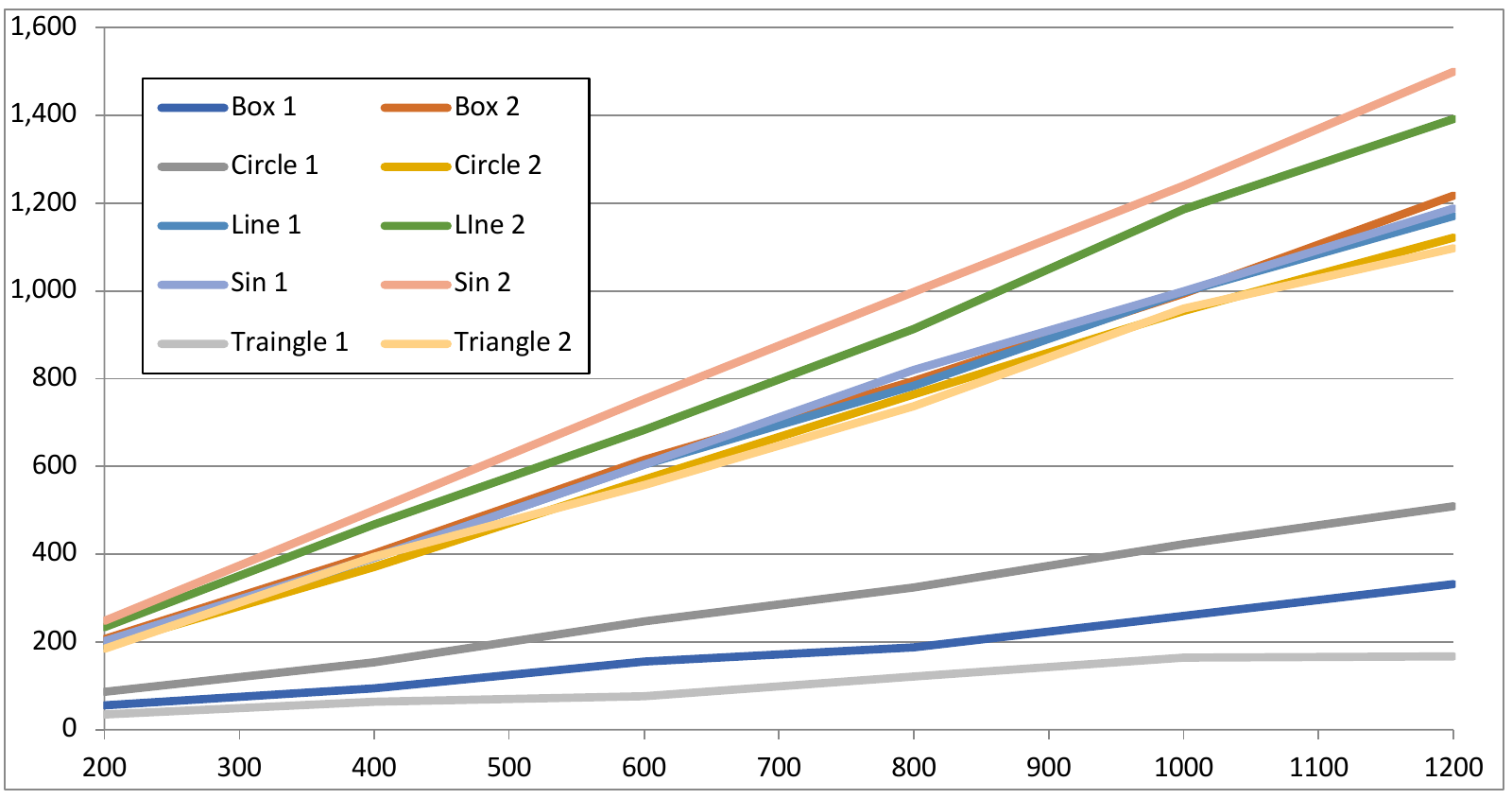}\\
	\scriptsize{(a) Average number of runs \hspace{2cm} (b) Average number of mutants}\\			
	\includegraphics[width=5cm, height=2.5cm]{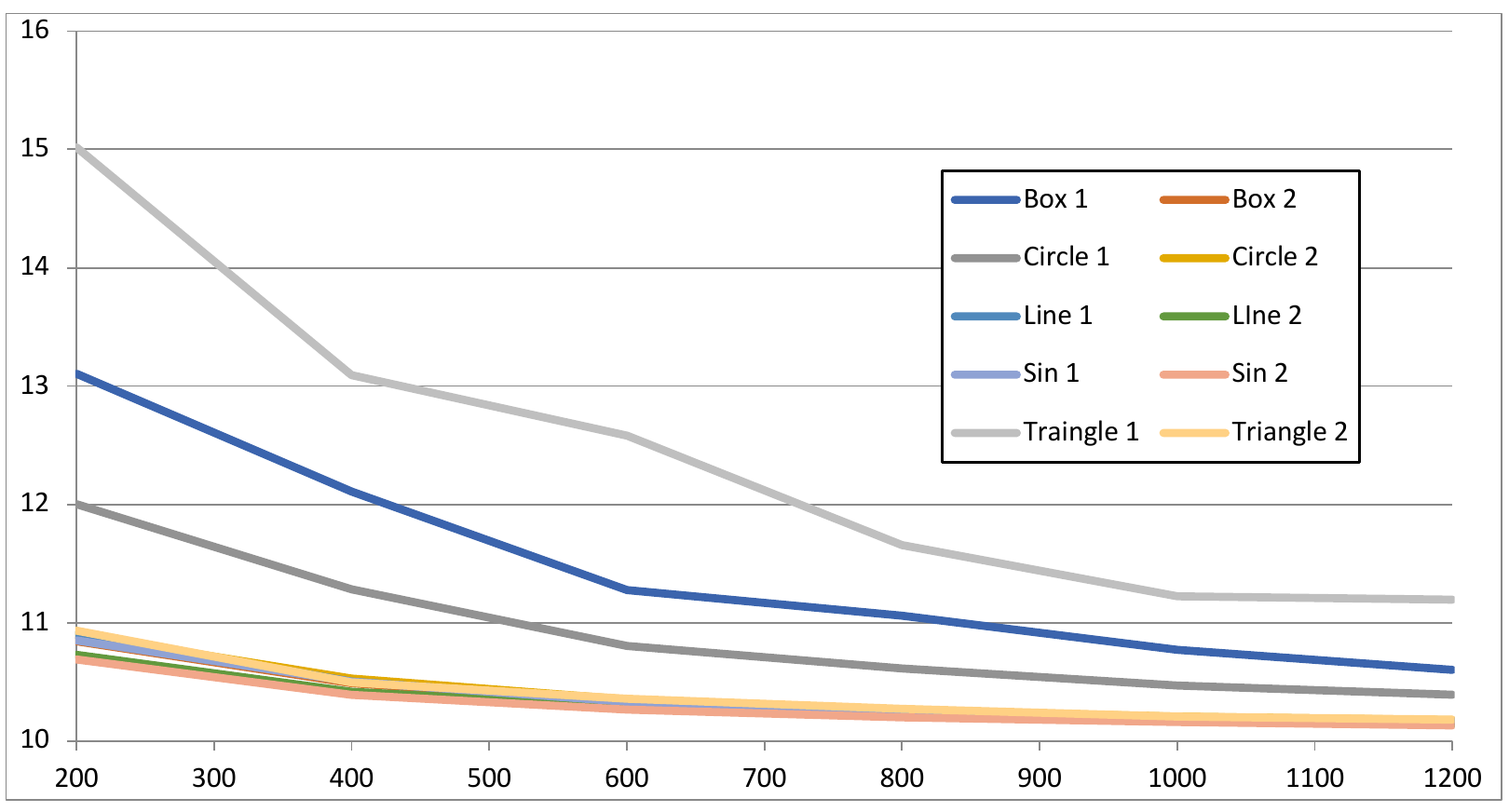}
	\includegraphics[width=5cm, height=2.5cm]{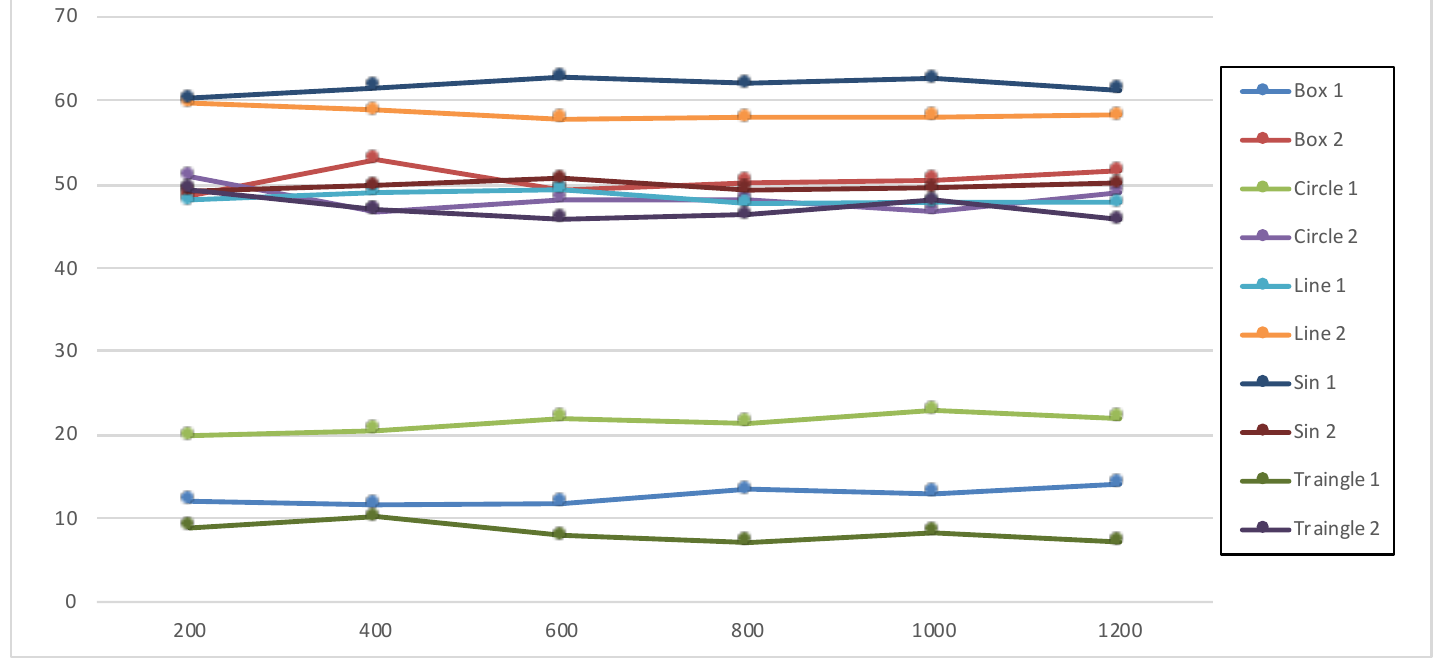}\\
	\scriptsize{(c) Average Cost \hspace{3cm} (d) Average Capability} 
	\caption{Results of the Random Target Strategy}
	\label{fig:AimedWalkResultChart}
\end{figure}

The random target strategy only has one parameter: the number of pairs of test cases selected at random. The experiments are conducted with this parameter ranging from 200 to 1200. The results, as shown in Figure \ref{fig:AimedWalkResultChart}(a) and (b), are that the average number of test executions and the average size of generated Pareto front are linear in the number of walks for all subject programs. The test cost, as shown in Figure \ref{fig:AimedWalkResultChart}(c), increases slightly with the number of walks since the average number of test executions needed to generate a test case in the Pareto front decreases as the number of walks increases. However, the capability remains invariant with the number of walks as shown in (d). 

\subsubsection{Discussion} \label{sec:Discussion}

From the experiments, we observed the following phenomena in addition to the results stated above. 

\begin{itemize}
\item \emph{Factors influencing cost and capability}
\end{itemize}

\begin{figure}[htbp]
	\centering
	\includegraphics[width=6cm, height=3.5cm]{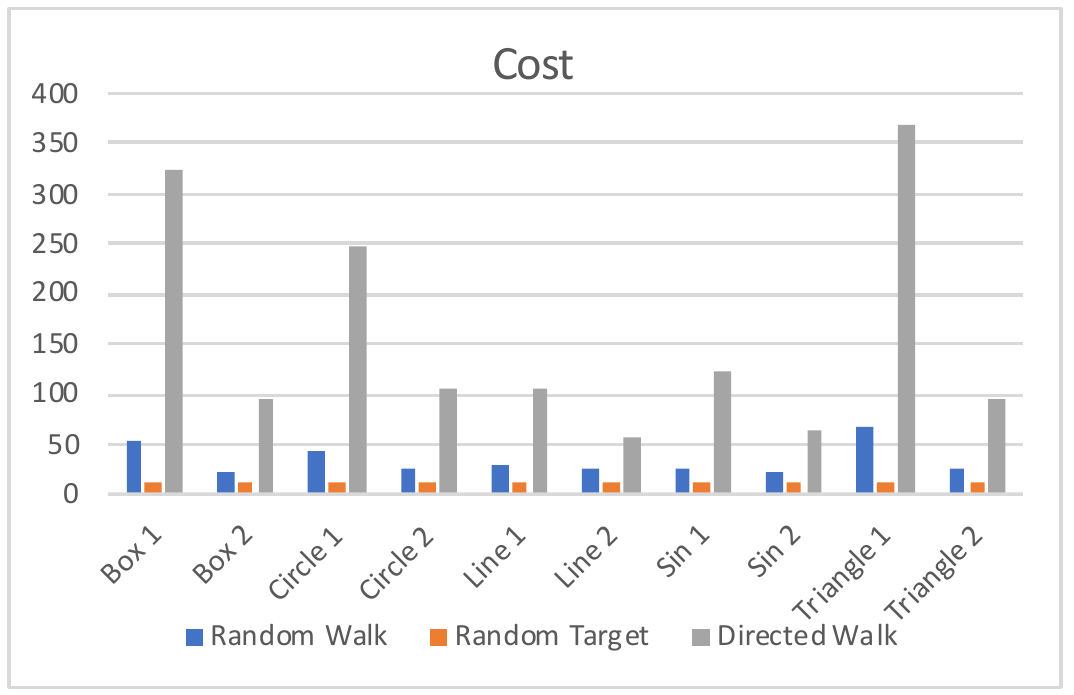}
	\includegraphics[width=6cm, height=3.5cm]{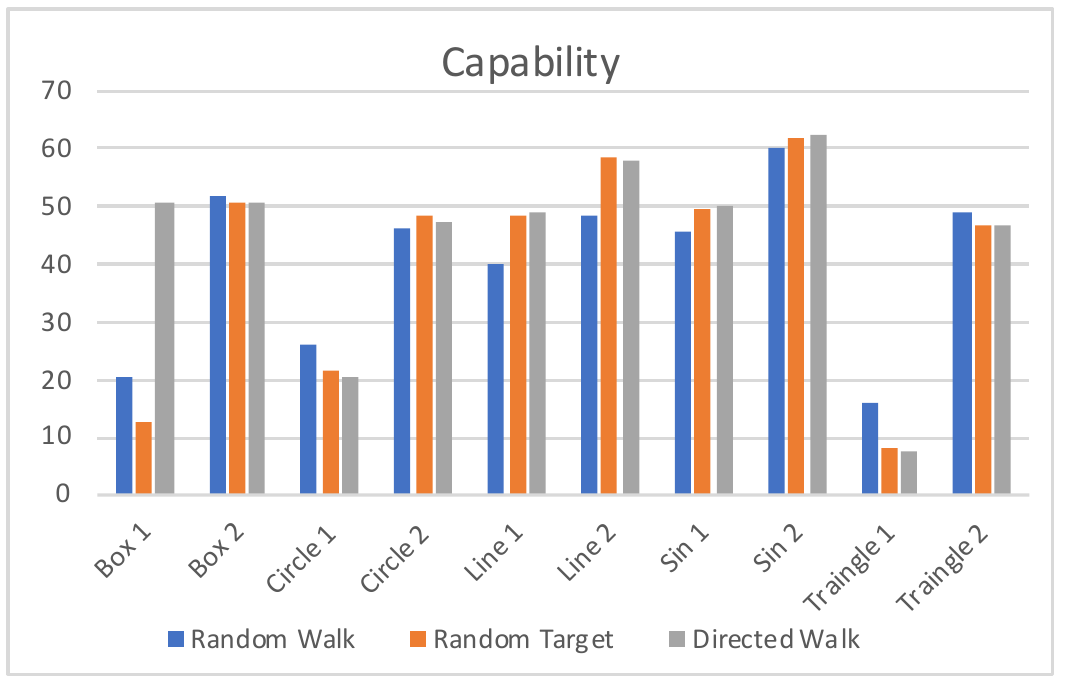}
	\caption{Test Cost and Capability on Subject Programs}
	\label{fig:ResultChartCompareStrategyEffectiveness}
\end{figure}

The test cost of the strategies on various subject programs are summarised in Table \ref{tab:ComparisonOfStrategies} and depicted in Figure \ref{fig:ResultChartCompareStrategyEffectiveness}, where larger numbers indicate higher test cost. 

% the experiments demonstrated that the random target strategy is the most effective strategy while the directed walk is the most ineffective one consistently on all subjects. 

\begin{table}[htbp]
\caption{Summary of Test Cost and Capability}
\begin{scriptsize}
\begin{center}
\begin{tabular}{|p{1.4cm}|c|c|c|c|c|c|}
\hline
\multirow{2}{*}{\textbf{Subject}} &\multicolumn{2}{c|}{\textbf{Directed Walk}} &\multicolumn{2}{c|}{\textbf{Random Walk}} &\multicolumn{2}{c|}{\textbf{Random Target }}\\ \cline{2-7}
& \textbf{Cost} &\textbf{Cap} &\textbf{Cost} &\textbf{Cap} &\textbf{Cost} &\textbf{Cap}\\ \hline
Box 1 		&323.45 	&50.53 	&52.46 	&20.72 	&11.49 	&12.69\\ \hline
Box 2 		&93.85 	&50.53 	&22.83 	&51.59 	&10.38 	&50.53\\ \hline
Circle 1 	&247.32 	&20.67 	&42.59 	&26.03 	&10.93 	&21.49\\ \hline
Circle 2 	&105.82 	&47.32 	&25.50 	&46.01 	&10.41 	&48.31\\ \hline
Line 1 		&105.82 	&49.15 	&29.02 	&40.13 	&10.41 	&48.25\\ \hline
Line 2 		&55.76 	&58.03 	&23.94 	&48.56 	&10.33 	&58.40\\ \hline
Sin 1 		&122.35 	&50.10 	&20.65 	&45.51 	&10.38 	&49.76\\ \hline
Sin 2 		&64.75 	&62.34 	&26.03 	&60.54 	&10.31 	&61.76\\ \hline
Triangle 1 	&370.38 	&7.62 	&66.79 	&16.06 	&12.46 	&8.33\\ \hline
Triangle 2 	&93.19 	&46.96 	&23.98 	&49.08 	&10.41 	&47.01\\ \hline
\textbf{Avg}  &\textbf{158.27} &\textbf{44.32} &\textbf{33.38} &\textbf{40.46} &\textbf{10.75} &\textbf{40.65}\\ \hline
\end{tabular}
\end{center}
\end{scriptsize}
\label{tab:ComparisonOfStrategies}
\end{table}

The data show that for each strategy, the test cost and capability vary significantly according to the subject programs. However, for each strategy, test cost and capability of Box 1 are lower than Box 2, Circle 1 is lower than Circle 2, and so on. This phenomenon is not a coincidence.

From the theorems given in Section \ref{sect:StrategyDefinitions}, we can see that the capability for the directed walk strategy is determined by the probability that there is a border between two subdomains in the right direction from a test case and within the walking distance. For the random target strategy, it is determined by the probability that two random test cases fall in two different subdomains, and for the random walk strategy, it is determined by the probability that there is a border near to a randomly selected test case. For test cost, the more Pareto front pairs found, the more runs of the classifier will be to refine the pairs of test cases in order to reduce the distance between each pair. 

Two implications follow from these properties. First of all, given a classification application, one should select the most cost efficient strategy to explore the Pareto fronts between subdomains based on the understanding of the application. The data obtained from our experiments are not sufficient to compare the strategies on their cost. This is because the probability of finding a pair in the Pareto front heavily depends on the size and location of the subdomains of the classification application. Our subjects in the experiments may not be representative of the distribution of the parameters in real applications. Secondly, we now have an explanation why the number of pairs generated for the Pareto front is a linear function of the number of walks since the results of a walk is independent of the results of its predecessors.

Moreover, although the cost is mostly determined by the size, shape and location of the subdomains that the program classifies, for directed walk and random walk strategies, it is also affected by the number of steps walked and the number of iterations in the refinement. The number of steps walked influences the probability of finding two points in different subdomains and also the total number of test executions. The longer the walk, the more likely one is to find two points in different subdomains, but this requires more test executions. Thus, a balance between these two contradictory factors of cost must be made to achieve the best test effectiveness. 

Finally, the number of iterations in the refinement loop controls the distance between the pairs of test cases in the Pareto fronts generated. It has no impact on capability, i.e. the probability of finding two data points in different subdomains, but it does have an effect on test cost. The shorter distance requires more iterations, and thus more test executions, and therefore, it is more costly. For random walk and directed walk strategies, the number of iterations can be selected according to the formula given in the correctness theorems given in Section \ref{sect:StrategyDefinitions}. For the random target strategy, usually more iterations are required than the other two strategies. 

\begin{itemize}
\item \emph{Validity of the experiments}
\end{itemize}

As pointed out at the beginning of the section, the experiments are designed to determine which factors have an effect on the capability and cost of the strategies. The subject programs used in the controlled experiments are manually coded by the authors. They have been designed in such a way that their subdomains are of typical shapes in data mining and machine learning applications \citep{DataMiningTextBook, FoundationsOfMLBook, UnderstandingMLBook}. As discussed above, they provide insight into the factors that affect capability and cost. 

The manual examinations of the Pareto fronts generated by the test strategies confirmed that they are indeed test cases very close to the borders of subdomains. The phenomena observed from the experiments is consistent with the predictions made from the theorems. However, the specific data about cost and capability obtained from the experiments depends on the specific features of the subdomains such as their sizes and locations. Therefore, the experiment data do not answer the question whether the test strategies are applicable to testing real machine learning applications. This issue is addressed in the case studies reported in the next subsection. 

\subsection{Case Studies}\label{sec:CaseStudy}

This subsection reports a set of case studies with the exploratory testing of machine learning and data analytics applications using the test strategies. 

\subsubsection{Design and Conduct of the Case Studies}

The procedure for the case study consists of the following steps:

\begin{enumerate}
  \item Select sample applications of classifiers. 
  \item For each selected sample, 
  \begin{enumerate}
  	\item Download the dataset.
    \item Construct classifiers by applying machine learning techniques on the dataset. 
  	\item Develop test system according to the specification of the test systems defined in Section \ref{sec:TestSystem}. 
  	\item Write test scripts in Morphy's test scripting language for repeated executions of the experiments and collection of data.
  	\item Execute test strategies on the test classifiers by running the test scripts. 
  \end{enumerate}
\end{enumerate}

The following describes each step in detail. 

\begin{itemize}
\item \emph{Sample datasets}
\end{itemize}

The following three datasets were selected at random from the well-known Kaggle collection of datasets for machine learning and data analytics applications. They were as follows:
\begin{description}
  \item [(1)] \emph{Red Wine Quality}. This dataset concerns red varieties of the Portuguese ``Vinho Verde" wine \citep{RedWinePaper}. There are 11 physicochemical variables as inputs (i.e. there is no data about grape types, wine brand, wine selling price, etc.) and the output is a classification of wine quality as a number from 1 to 10. The classes are ordered but not balanced in that there are many more normal wines than excellent or poor ones. 
  \item [(2)] \emph{Mushroom Edibility.} This dataset concerns hypothetical samples of 23 species of gilled mushrooms in the Agaricus and Lepiota family drawn from The Audubon Society Field Guide to North American Mushrooms \cite{MushroomBook}. Each species is identified as definitely edible, definitely poisonous, or of unknown edibility and not recommended. This latter class was combined with the poisonous one in the dataset. The Guide clearly states that there is no simple rule for determining the edibility of a mushroom, i.e. no rule like ``leaflets three, let it be'' for poison oak and poison ivy. The dataset has been available to researchers on data mining and machine learning for 30 years. 
  \item [(3)] \emph{Bank Churners.} This dataset concerns credit card customers and can be used to predict churners, who are bank customers who leave the credit card service. It consists of more than 10,000 real data items with 19 features about customer's age, salary, marital status, credit card limit, credit card category, etc. It is, however, considered to be a difficult task to train a model to predict churning customers. 
\end{description}

All three datasets are available from the Kaggle repository
\footnote{https://www.kaggle.com/uciml/red-wine-quality-cortez-et-al-2009}
\footnote{https://www.kaggle.com/uciml/mushroom-classification}
\footnote{https://www.kaggle.com/sakshigoyal7/credit-card-customers}.
The first two datasets are commonly used in research on machine learning and data analytics to determine which physiochemical properties make a wine good and which features are most indicative of a poisonous mushroom, respectively. As well as Kaggle, they can also be found at the UCI machine learning repository. \footnote{https://archive.ics.uci.edu/ml/datasets/wine+quality}
\footnote{https://archive.ics.uci.edu/ml/datasets/Mushroom} The Bank Churners dataset originates from a LEAPS website \footnote{https://leaps.analyttica.com/home}, which specialises in application of data analytics and machine learning techniques to solve business problems. 

Table \ref{tab:SummaryDatasets} summarises the datasets used in the case study. The column \emph{Records} gives the number of records in the dataset and \emph{Classes} is the number of classes (subdomains) in the classification. Columns \emph{DF}, \emph{NF} and \emph{CF} are the numbers of discrete non-numerical features, discrete numerical features and continuous numerical features, respectively. The column \emph{Features} shows the total number of features. We can see that the dataset \emph{Red Wine Quality} is a continuous numerical data space, whereas the dataset \emph{Mushroom Edibility} is a discrete non-numerical data space, and the \emph{Bank Churners} dataset is a hybrid data space. 
 
\begin{table}[htbp]
\caption{Summary of Datasets}
\begin{small}
\begin{center}
\begin{tabular}{|l|c|c|c|c|c|c|}
\hline
\textbf{Dataset} &\textbf{Records} &\textbf{Classes} &\textbf{DF} &\textbf{NF} &\textbf{CF} &\textbf{Features}  \\ \hline
Red Wine Quality 	&1599	&8	&0	&0	&11	&11\\ \hline
Mushroom Edibility	&8124	&2	&22	&0	&0	&22\\ \hline
Bank Churners	&10127	&2	&5	&11	&3	&19\\ \hline
\end{tabular}
\end{center}
\end{small}
\label{tab:SummaryDatasets}
\end{table}%

\begin{itemize}
\item \emph{Construction of machine learning models}
\end{itemize}

Since the goal of the case study is to demonstrate that our test strategies are applicable to real machine learning applications, we have used the datasets to train models that use a wide variety of machine learning techniques. This enables us to demonstrate that our testing techniques are effective on both low-quality and high-quality models as well as on different types of models.

The training consists of executing a Python program, adapted from code posted on the Kaggle website  and selected at random again. For each dataset, we  build 16 different models, as shown in Table \ref{tab:ModelInfo}. The Python programs for training and invoking the models as well as all datasets used in the case study can be found on the project's GitHub repository; see Footnote \ref{ftn:GitHubURL} for the URL.

\begin{table}[htbp]
\caption{Machine Learning Models Constructed for Each Dataset}
\begin{scriptsize}
\begin{center}
\begin{tabular}{|l|l|l|}
\hline
\textbf{Name}	&\textbf{Type}	&\textbf{Details}\\ \hline
LR	&Logistic Regression	&Trained on whole data set\\ \hline
LR2	&Logistic Regression	&Used train-test 90-10 split \\ \hline
KNN	&K-Nearest Neighbors	&Trained on whole data set\\ \hline
KNN2	&K-Nearest Neighbors	&Used train-test 90-10 split\\ \hline
DT	&Decision Tree	&Trained on whole data set\\ \hline
DT2	&Decision Tree	&Used train-test 90-10 split\\ \hline
%RF	&Random Forest	&Trained on whole data set\\ \hline
%RF2	&Random Forest	&Used train-test 90-10 split\\ \hline
NB	&Naive Bayes	&Trained on whole data set\\ \hline
NB2	&Naive Bayes	&Used train-test 90-10 split\\ \hline
SVM	&Surportting vector machine	&Trained on whole data set\\ \hline
SVM2	&Surportting vector machine	&Used train-test 90-10 split\\ \hline
SV	&Ensemble via Soft voting	&Trained on whole data set; LR+KNN+DT\\ \hline
SV2	&Ensemable via Soft Voting	&Used train-test 90-10 split; LR+KNN+DT\\ \hline
HV	&Ensemble via Hard Voting	&Trained on whole data set; LR+KNN+DT\\ \hline
HV2	&Ensemble via Hard Voting	&Used train-test 90-10 split; LR+KNN+DT\\ \hline
Stack1	&Ensemble via Stacking	&Used train-test 90-10 split; KNN as Meta; LR2+KNN2+DT2+HV2\\ \hline 
%Stack2	&Ensemble via Stacking	&Trained on whole data set; LR as Meta; KNN+DT+SV+RF \\ \hline
Stack3	&Ensemble via Stacking	&Used train-test 90-10 split;  LR as Meta; KNN2+DT+SV2+HV2\\ \hline
\end{tabular}
\end{center}
\end{scriptsize}
\label{tab:ModelInfo}
\end{table}%

A total of 48 models were constructed. Their accuracy varies from 49.9\% to 100\%; see Appendix B.1 for details. It is worth noting that no effort was spent to construct a model of high quality because the purpose of the experiment is to determine if the strategies are capable and cost efficient for models of all different kinds of quality.

\begin{itemize}
\item \emph{Development of test systems}
\end{itemize}

The test system for the Red Wine Quality dataset was a straightforward implementation of the appropriate algorithm in Section \ref{sec:TestSystem} and the code for the experiments was a clone of the code written for Section \ref{sec:Experiments}. The main difference in the test system is that the executions are performed by invoking programs in Python through executing test morphisms in Java.

The test system for Mushroom Edibility was made by refactoring the test system for Red Wine Quality to make the code common to both ready for reuse. Once again, the datamorphisms were a straightforward implementation of the definitions in Section \ref{sec:TestSystem}. Similarly, the test system for Bank Churners prediction is again a straightforward implementation of the algorithms given in Section \ref{sec:TestSystem}. 

\begin{itemize}
\item \emph{Executions of test strategies}
\end{itemize}

As with the controlled experiments in Section \ref{sec:Experiments}, the test strategies are applied to each classifier to generate the Pareto fronts and the same kinds of data are collected from their executions.

In particular, both the random target and random walk test strategies were executed with varying numbers of walks (10 times in each case) ranging from 100 to 1000 in order to calculate the average number of mutant test cases generated, i.e. the number of test cases in the generated Pareto front.  The directed walk strategy was executed with starting points of 100, 200, ..., 1000 test cases selected at random from the original dataset on all directions (i.e. each unary datamorphism) for 10 times; the average of these directions was calculated for each of the models. 

The repeated executions of the test strategies were conducted by invoking test scripts written in Morphy's test scripting facility. The test scripts can be found in the GitHub repository. 

\subsubsection{Main Results}

%In this section, we first present the basic data collected in the case study, then answer the research questions by analysing the data. 

\begin{itemize}
\item \emph{Numbers of runs and mutants}
\end{itemize}

\begin{figure}[h]
	\centering
	\includegraphics[width=4cm, height=2.5cm]{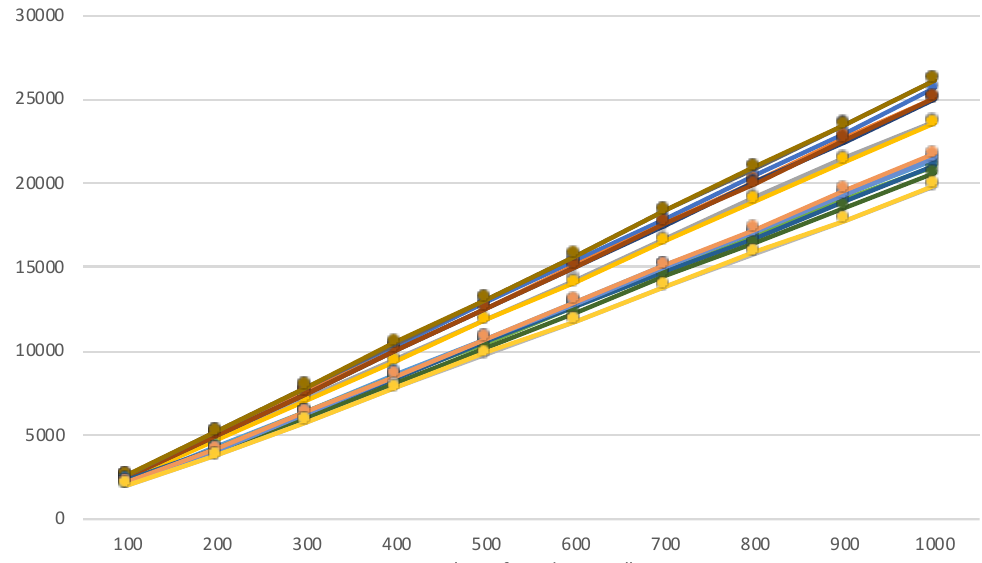}
%	\includegraphics[width=4cm, height=2.5cm]{figures/RedWine-DirectedWalk-Runs.pdf}\\
%	\vspace{-0.3cm}
%	\begin{flushleft}
%		\scriptsize {Red Wine: \hspace{0.7cm} (a) Random Target} \hspace{1.5cm} 
%		\scriptsize{(b) Random Walk} \hspace{2cm} 
%		\scriptsize{(c) Directed Walk} 
%	\end{flushleft}
	\includegraphics[width=4cm, height=2.5cm]{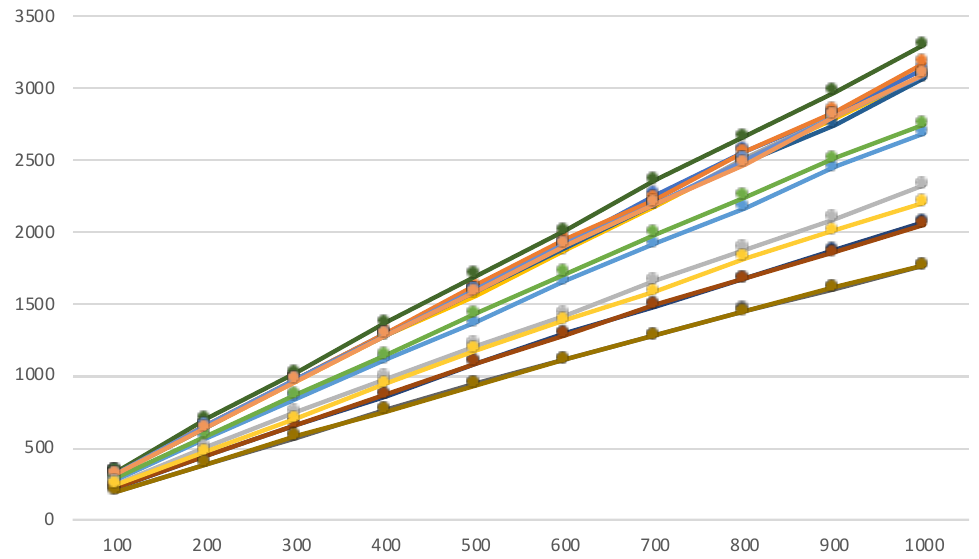}
%	\includegraphics[width=4cm, height=2.5cm]{figures/Mushroom-RandomWalk-Runs.pdf}
%	\includegraphics[width=4cm, height=2.5cm]{figures/Mushroom-DirectedWalk-Runs.pdf}\\
%	\vspace{-0.3cm}
%	\begin{flushleft}
%		\scriptsize {Mushroom: \hspace{0.7cm} (a) Random Target} \hspace{1.5cm} 
%		\scriptsize{(b) Random Walk} \hspace{2cm} 
%		\scriptsize{(c) Directed Walk} 
%	\end{flushleft}
%	\includegraphics[width=4cm, height=2.5cm]{figures/Bank-RandomTarget-Runs.pdf}
%	\includegraphics[width=4cm, height=2.5cm]{figures/Bank-RandomWalk-Runs.pdf}
	\includegraphics[width=4cm, height=2.5cm]{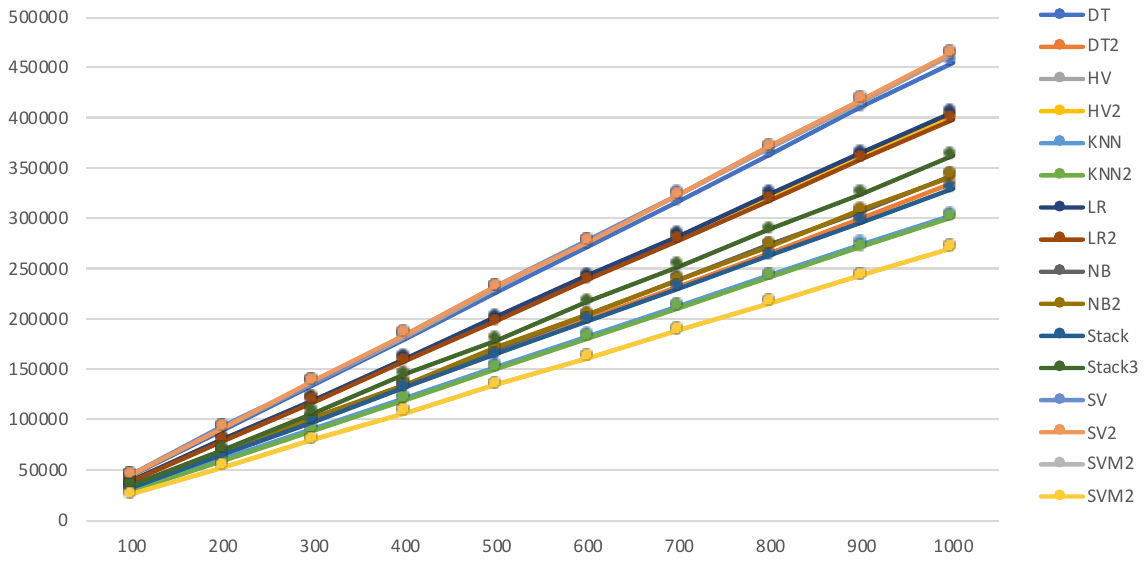}\\
%	\vspace{-0.3cm}
%	\begin{flushleft}
%		\scriptsize {Bank: \hspace{1.2cm} (a) Random Target} \hspace{1.5cm} 
%		\scriptsize{(b) Random Walk} \hspace{2cm} 
%		\scriptsize{(c) Directed Walk} 
%	\end{flushleft}
\scriptsize {(a) Random Walk on Red Wine \hspace{0.8cm} (b) Random Target on Mushroom \hspace{0.7cm} 
(c) Directed Walk on Bank} 
	\caption{Variation of the Number of Runs with the Number of Walks}
	\label{fig:CaseStudy-Runs}
\end{figure}

 The case studies clearly show that for all machine learning models, the average numbers of runs of the model increase linearly with the number of walks made when executing the test strategies. Figure \ref{fig:CaseStudy-Runs} shows some typical examples. 

\begin{figure}[h]
	\centering
%	\includegraphics[width=4cm, height=2.5cm]{figures/RedWine-RandomTarget-Mutants.pdf}
%	\includegraphics[width=4cm, height=2.5cm]{figures/RedWine-RandomWalk-Mutants.pdf}
%	\includegraphics[width=4cm, height=2.5cm]{figures/RedWine-DirectedWalk-Mutants.pdf}\\
%	\vspace{-0.3cm}
%	\begin{flushleft} 
%		\scriptsize{Red Wine: \hspace{0.7cm}(a) Random Target} \hspace{2cm} 
%		\scriptsize{(b) Random Walk} \hspace{2cm} 
%		\scriptsize{(c) Directed Walk} 
%	\end{flushleft}
	\includegraphics[width=4cm, height=2.5cm]{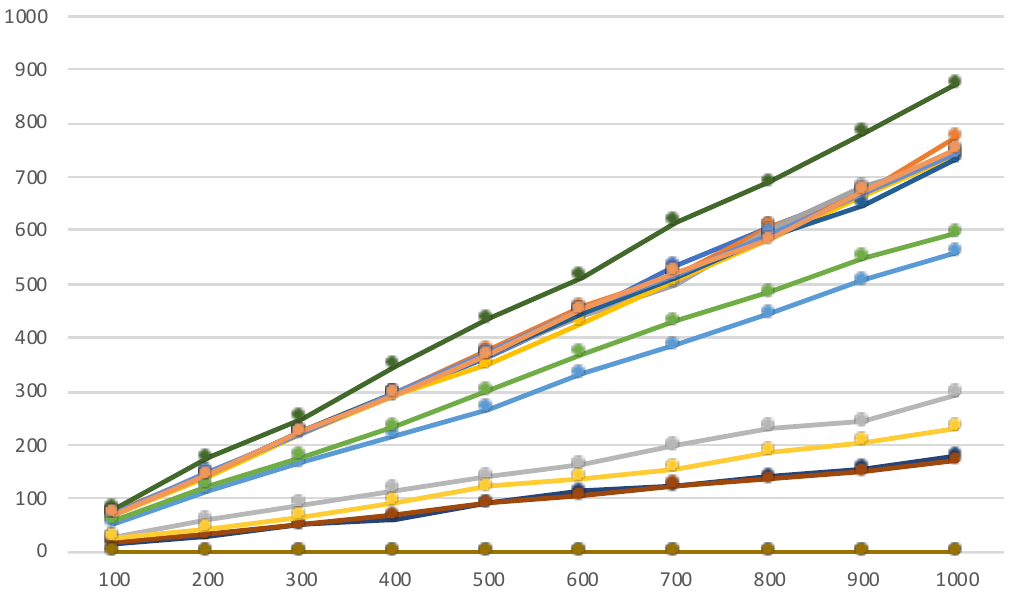}
%	\includegraphics[width=4cm, height=2.5cm]{figures/Mushroom-RandomWalk-Mutants.pdf}
%	\includegraphics[width=4cm, height=2.5cm]{figures/Mushroom-DirectedWalk-Mutants.pdf}\\
%	\vspace{-0.3cm}
%	\begin{flushleft} 
%		\scriptsize{Mushroom: \hspace{0.7cm} (a) Random Target} \hspace{1.5cm} 
%		\scriptsize{(b) Random Walk} \hspace{2cm} 
%		\scriptsize{(c) Directed Walk} 
%	\end{flushleft}
%	\includegraphics[width=4cm, height=2.5cm]{figures/Bank-RandomTarget-Mutants.pdf}
	\includegraphics[width=4cm, height=2.5cm]{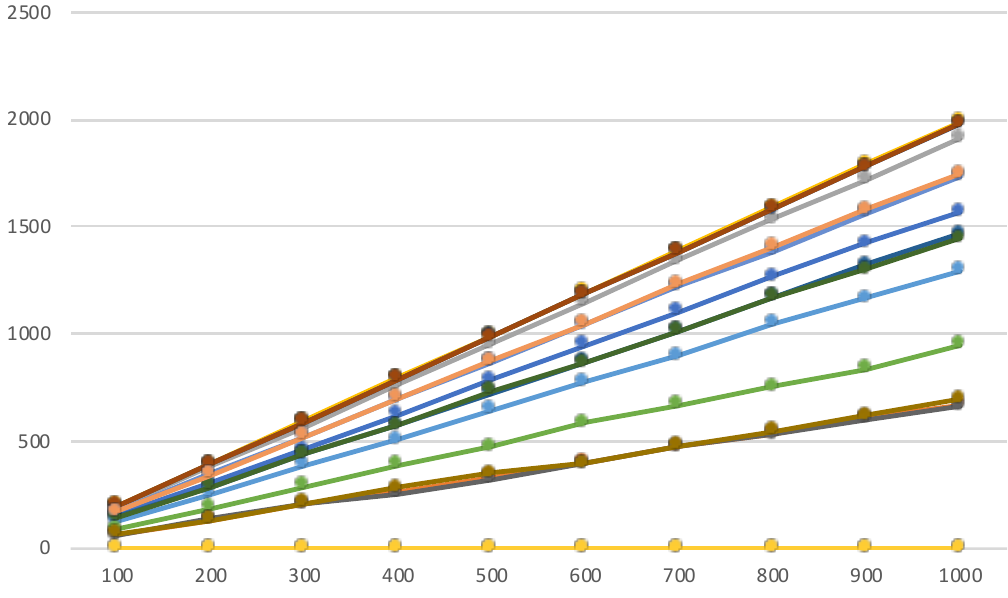}
%	\includegraphics[width=4cm, height=2.5cm]{figures/Bank-DirectedWalk-Mutants.pdf}\\
%	\vspace{-0.3cm}
%	\begin{flushleft} 
%		\scriptsize{Bank: \hspace{1.2cm} (a) Random Target} \hspace{1.5cm} 
%		\scriptsize{(b) Random Walk} \hspace{2cm} 
%		\scriptsize{(c) Directed Walk} 
%	\end{flushleft}
	\includegraphics[width=4cm, height=2.5cm]{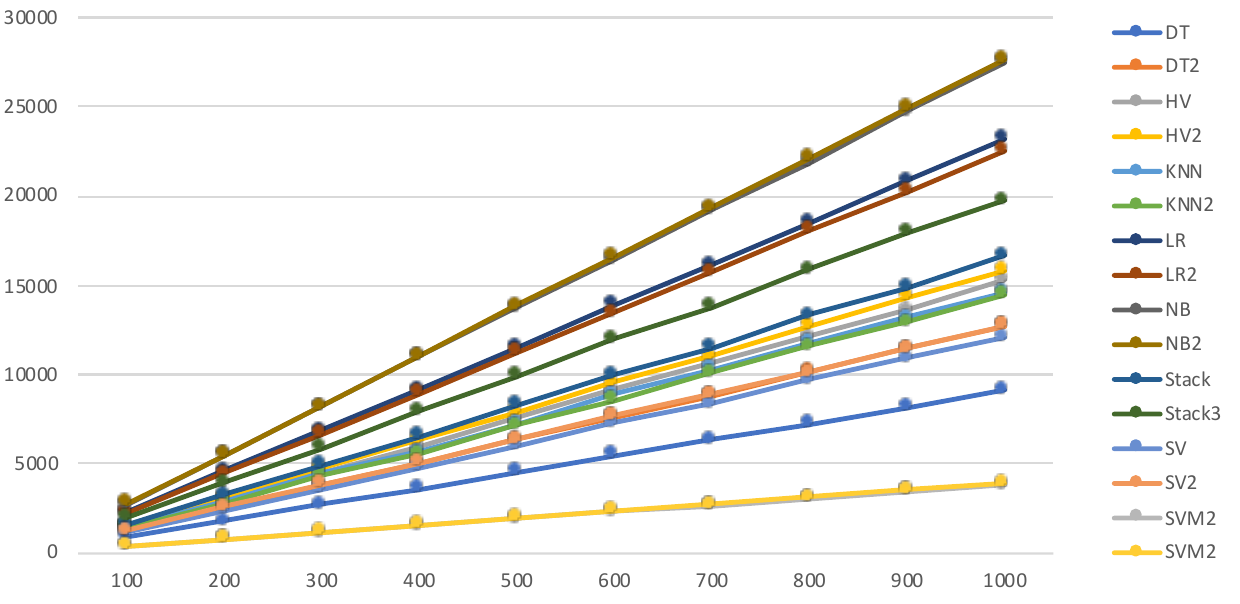}\\
	\scriptsize{(a) Random Target on Mushroom \hspace{0.8cm} (b) Random Walk on Bank \hspace{0.9cm} (c) Directed Walk on Red Wine} 
	\caption{Variation of The Number of Mutants with the Number of Walks}
	\label{fig:CaseStudyMutants}
\end{figure}

Similarly, the average numbers of mutant test cases (i.e. the points in Pareto fronts generated by strategies) increase linearly with the number of walks from 100 to 1000. Again, this is for all machine learning models. Three typical examples are shown in Figure \ref{fig:CaseStudyMutants}. 

The data of the case studies confirmed the observations made in the controlled experiments. 

\begin{itemize}
\item \emph{Capability of discovering borders} 
\end{itemize}

\begin{figure}[h]
	\centering
	\includegraphics[width=4cm, height=2.5cm]{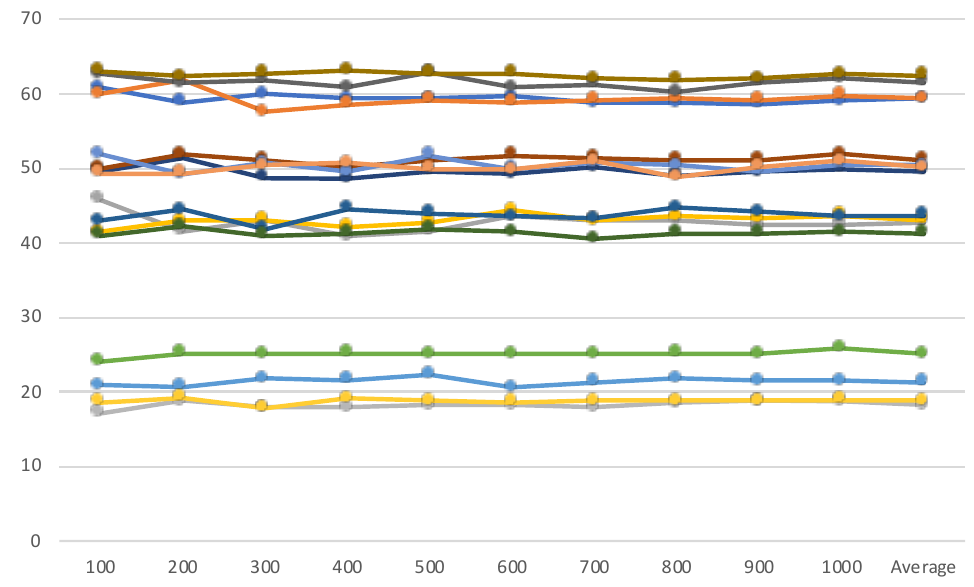}
%	\includegraphics[width=4cm, height=2.5cm]{figures/RedWine-RandomWalk-Capability.pdf}
%	\includegraphics[width=4cm, height=2.5cm]{figures/RedWine-DirectedWalk-Capability.pdf}\\
%	\vspace{-0.3cm}
%	\begin{flushleft}
%		\scriptsize{Red Wine: \hspace{0.7cm} (a) Random Target} \hspace{2cm} 
%		\scriptsize{(b) Random Walk} \hspace{2cm} 
%		\scriptsize{(c) Directed Walk} 
%	\end{flushleft}
%	\includegraphics[width=4cm, height=2.5cm]{figures/Mushroom-RandomTarget-Capability.pdf}
	\includegraphics[width=4cm, height=2.5cm]{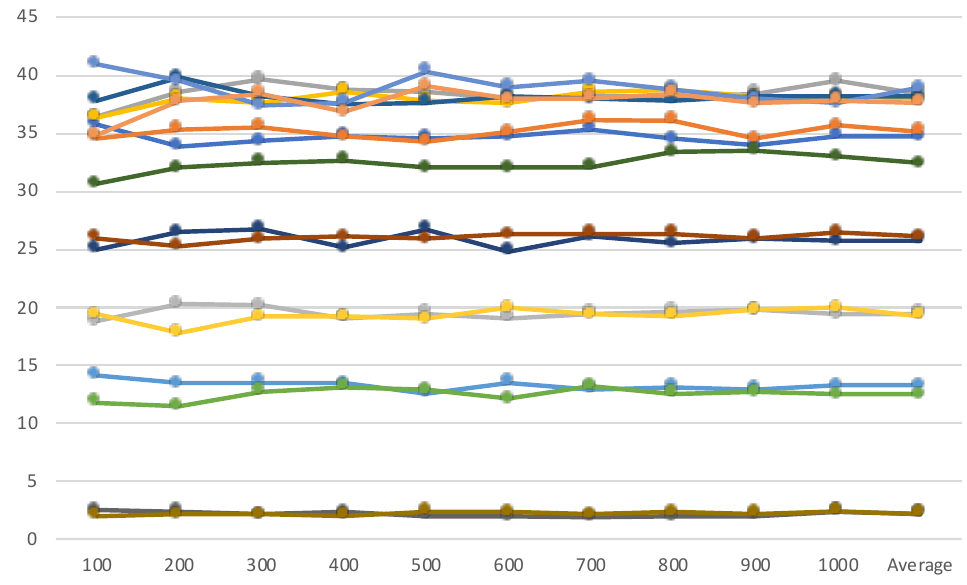}
%	\includegraphics[width=4cm, height=2.5cm]{figures/Mushroom-DirectedWalk-Capability.pdf}\\
%	\vspace{-0.3cm}
%	\begin{flushleft}
%		\scriptsize{Mushroom: \hspace{0.7cm} (a) Random Target} \hspace{2cm} 
%		\scriptsize{(b) Random Walk} \hspace{2cm} 
%		\scriptsize{(c) Directed Walk} 
%	\end{flushleft}
%	\includegraphics[width=4cm, height=2.5cm]{figures/Bank-RandomTarget-Capability.pdf}
%	\includegraphics[width=4cm, height=2.5cm]{figures/Bank-RandomWalk-Capability.pdf}
	\includegraphics[width=4cm, height=2.5cm]{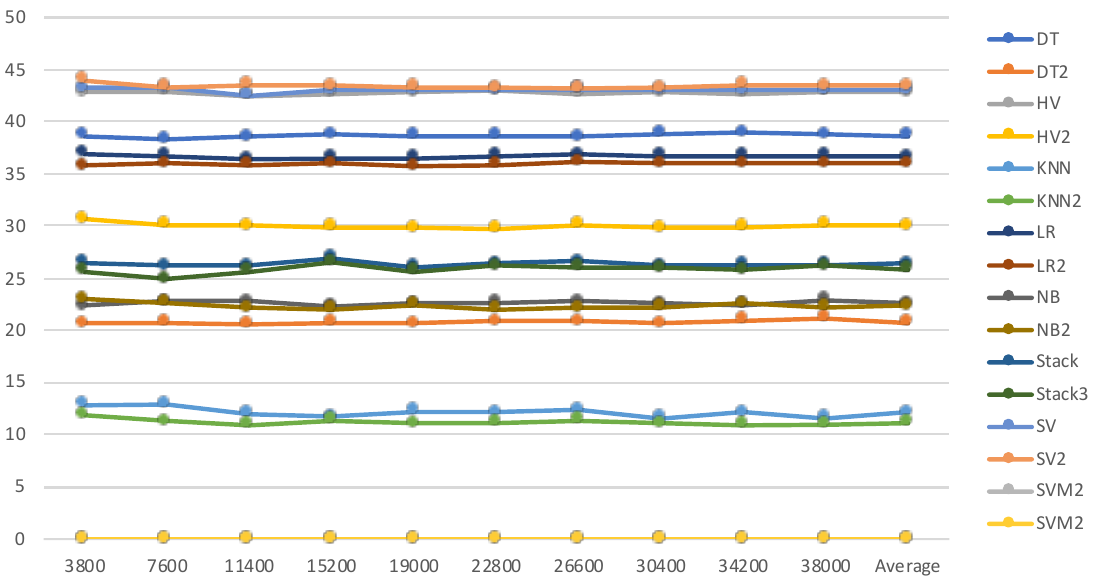}\\
%	\vspace{-0.3cm}
%	\begin{flushleft}
%		\scriptsize{Bank: \hspace{1.2cm} (a) Random Target} \hspace{2cm} 
%		\scriptsize{(b) Random Walk} \hspace{2cm} 
%		\scriptsize{(c) Directed Walk} 
%	\end{flushleft}
	\scriptsize{(a) Random Target on Red Wine \hspace{1cm} (b) Random Walk on Mushroom \hspace{1cm} 
(c) Directed Walk on Bank} 
	\caption{Variation of Capability with the Number of Walks}
	\label{fig:CaseStudyCapabilityVariation}
\end{figure}

\begin{figure}[h]
	\centering
	\includegraphics[width=13cm]{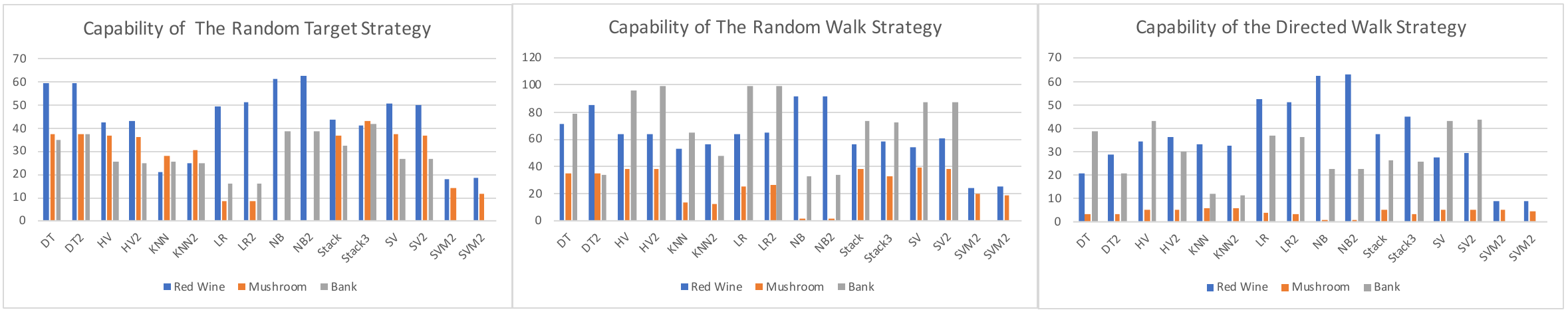}
	\caption{Capabilities of Testing Different ML Models}
	\label{fig:CaseStudyCapability}
\end{figure}

The capability of each test strategy in discovering border points for each machine model, measured as the probability of finding a border point via a walk, remains invariant in the number of walks as shown in Figure \ref{fig:CaseStudyCapabilityVariation}. However, the capability varies significantly over different machine learning models; see Figure \ref{fig:CaseStudyCapability}. 

\begin{itemize}
\item \emph{Cost}
\end{itemize}

\begin{figure}[h]
	\centering
%	\includegraphics[width=4cm, height=2.5cm]{figures/RedWine-RandomTarget-Effectiveness.pdf}
%	\includegraphics[width=4cm, height=2.5cm]{figures/RedWine-RandomWalk-Effectiveness.pdf}
%	\vspace{-0.3cm}
%	\begin{flushleft}
%		\scriptsize{Red Wine: \hspace{0.7cm} (a) Random Target} \hspace{2cm} 
%		\scriptsize{(b) Random Walk} \hspace{2cm} 
%		\scriptsize{(c) Directed Walk} 
%	\end{flushleft}
%	\includegraphics[width=4cm, height=2.5cm]{figures/Mushroom-RandomTarget-Effectiveness.pdf}
	\includegraphics[width=4cm, height=2.5cm]{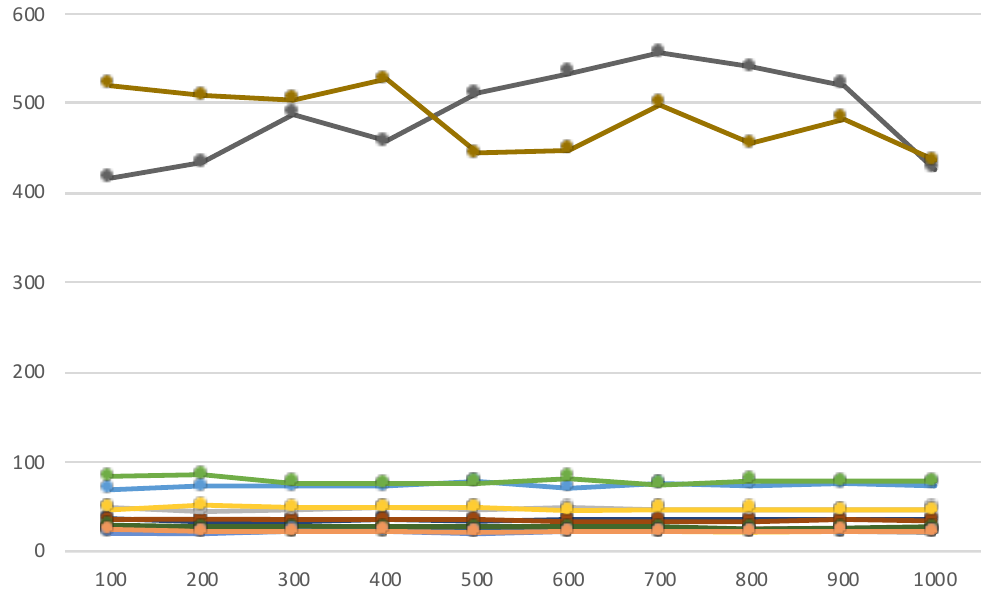}
%	\includegraphics[width=4cm, height=2.5cm]{figures/Mushroom-DirectedWalk-Effectiveness.pdf}\\
%	\vspace{-0.3cm}
%	\begin{flushleft}
%		\scriptsize{Mushroom: \hspace{0.7cm} (a) Random Target} \hspace{2cm} 
%		\scriptsize{(b) Random Walk} \hspace{2cm} 
%		\scriptsize{(c) Directed Walk} 
%	\end{flushleft}
	\includegraphics[width=4cm, height=2.5cm]{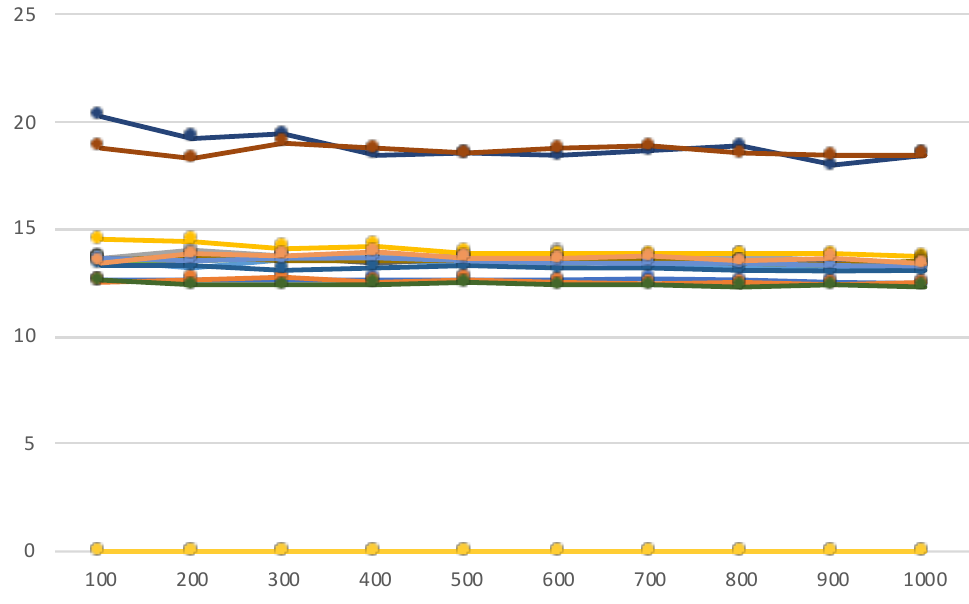}
%	\includegraphics[width=4cm, height=2.5cm]{figures/Bank-RandomWalk-Effectiveness.pdf}
%	\includegraphics[width=4cm, height=2.5cm]{figures/Bank-DirectedWalk-Effectiveness.pdf}\\
%	\vspace{-0.3cm}
%	\begin{flushleft}
%		\scriptsize{Bank: \hspace{1.2cm} (a) Random Target} \hspace{2cm} 
%		\scriptsize{(b) Random Walk} \hspace{2cm} 
%		\scriptsize{(c) Directed Walk} 
%	\end{flushleft}
	\includegraphics[width=4cm, height=2.5cm]{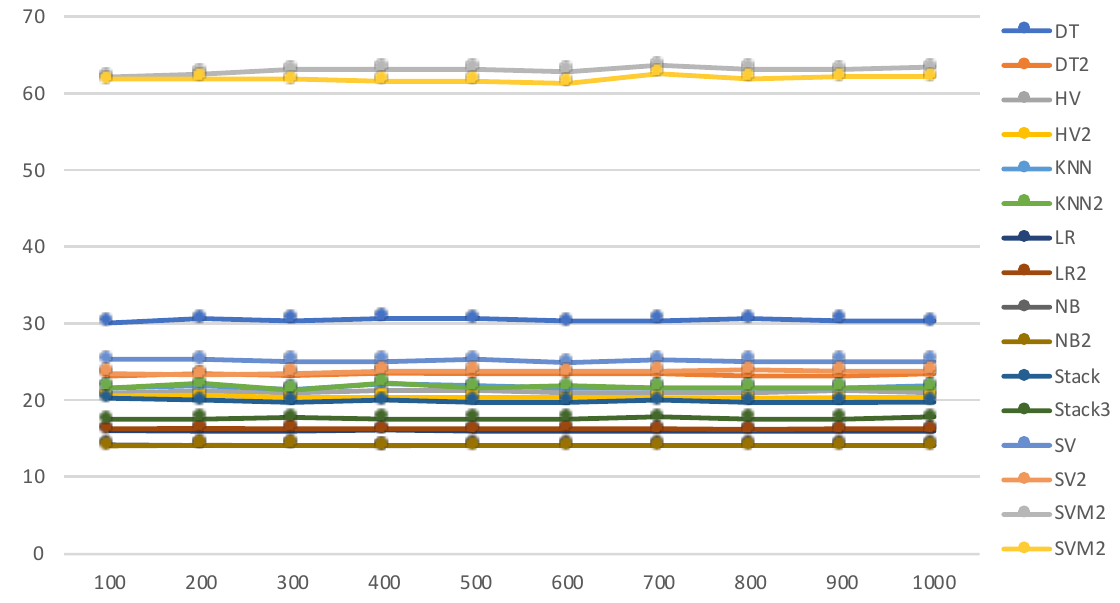}\\
	\scriptsize{(a) Random Walk on Mushroom \hspace{1cm} (b) Random Target on Bank \hspace{1cm} (c) Directed Walk on Red Wine}
	\caption{Variations of Cost with Numbers of Walks}
	\label{fig:CaseStudyEffectiveness}
\end{figure}

\begin{figure}[h]
	\centering
	\includegraphics[width=13cm]{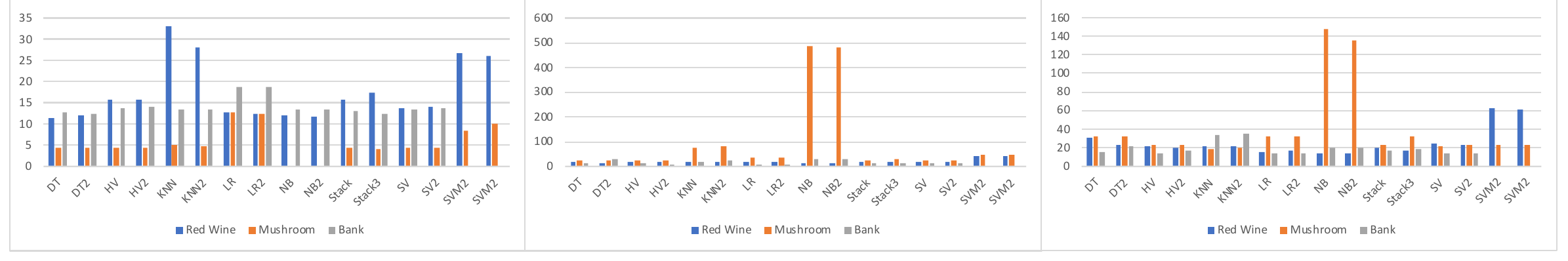}\\
	\scriptsize{(a) Random Target \hspace{2.5cm} (b) Random Walk \hspace{2.5cm} (c) Directed Walk}
	\caption{Cost of Testing Different ML Models}
	\label{fig:CaseStudyAvgEffectiveness}
\end{figure}

As was seen with the controlled experiments in Section \ref{sec:Experiments}, the case studies show that the cost of the strategies was mostly invariant as the number of walks increases; see Figure \ref{fig:CaseStudyEffectiveness} for some typical examples. The cost for each model is shown in Figure \ref{fig:CaseStudyAvgEffectiveness}. 

\subsubsection{Discussion}

\begin{itemize}
\item \emph{Answers to the research questions.}
\end{itemize}

From the data collected from the case studies, we can draw the following conclusions. 

First, the data of the case study are consistent with the observations made in the controlled experiments that both capability and cost of the strategies heavily depends on the model under test, but is invariant in the number of walks. In other words, both cost and capability are constants that only vary with the model under test. 

Second, the strategies are capable of discovering borders between subdomains. The overall average of the capabilities of all three strategies is 34.48\%. The average capabilities of the directed walk, random target and random walk strategies over three subjects are 21.86\%, 31.47\% and 50.10\%, respectively. The highest capability reached was 62.83\% in testing bank churner prediction using the random walk strategy. The average capabilities are almost all above 25\% except that the average capability of testing mushroom edibility models using the directed walk strategy is only 4.10\%. Table \ref{tab:CaseStudySummaryData} shows the maximal, minimal, and the average capability and cost of the test strategies over different models.\footnote{When no Pareto Front is found, the cost is infinite. In such cases, the numbers given in Table \ref{tab:CaseStudySummaryData} for the maximal, minimal and average cost have been calculated by excluding the infinite.} 

\begin{table}[h]
	\centering
	\caption{Summary of the Capability and Cost of the Strategies}
	\includegraphics[width=13cm]{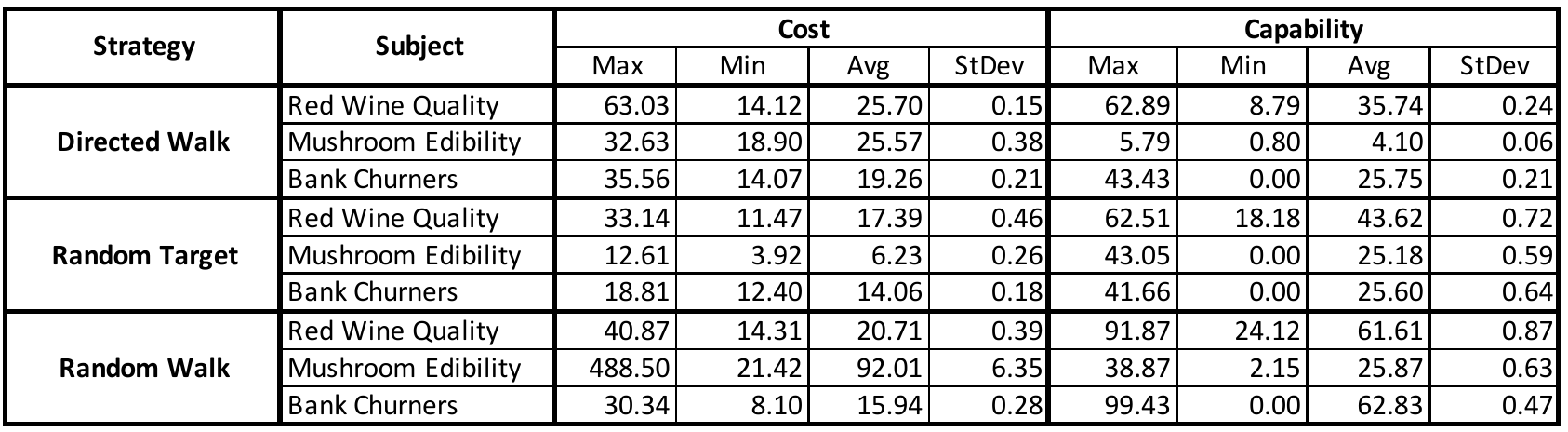}
	\label{tab:CaseStudySummaryData}
\end{table}

%\begin{figure}[h]
%	\centering
%	\includegraphics[width=6cm]{figures/Overall-Analysis-Capability.pdf}
%	\includegraphics[width=6cm]{figures/Overall-Analysis-Effectiveness.pdf}\\
%	\scriptsize{(a) Capability \hspace{5cm} (b) Effectiveness}
%	\caption{Average Capability of the Strategies}
%	\label{fig:CaseStudySummary}
%\end{figure}

Third, the case study also clearly demonstrated that applying exploratory strategies is cost efficient for discovering borders between classes; also see Table \ref{tab:CaseStudySummaryData}. The overall average cost of three strategies over all subjects is 26.32, which means that on average one would detect a border point by executing the machine learning model on about 27 test cases. In other words, within a fraction of second, a large number of border points can be found by applying these exploratory test strategies. The best cost efficiency was achieved in the testing of mushroom edibility models using the random target strategy, where the average cost over 16 models is 6.23. In contrast, the worst cost of 92.01 is observed also when testing mushroom edibility but using the random walk strategy. 

Fourth, comparing with the data of the controlled experiments, we observed that the costs and capabilities of the strategies in the case study are compatible to those of controlled experiments, although the dimensions of the input data spaces of the real-world examples are significantly larger than those coded classifiers. This indicates that the approach is scalable to high dimensional data spaces.

Moreover, the data of the case study provides some useful hint for the choice of strategies when testing a machine learning application. The data show that on average, the random walk strategy is the most capable in detecting borders. However, the walk may require many steps to find a border point. Thus, it could be slightly less cost efficient than the random target strategy in many cases. For the directed walk strategy, searching for borders in all directions is very much like a brute force search. Thus, it could be of higher cost in general. 

%\begin{figure}[h]
%	\centering
%	\includegraphics[width=7cm]{figures/Overall-Analysis-Effectiveness.pdf}
%	\caption{Average Effectiveness of the Strategies}
%	\label{fig:CaseStudySummaryEffectiveness}
%\end{figure}

Finally, in the case study, we observed a few cases where exploratory strategies performed poorly. These cases provide some insight for how to choose from the proposed strategies. 

Among the worst capabilities observed in the case studies is that of the directed walk strategy which performed poorly on testing mushroom edibility with an average capability of 4.10\% over 16 models. The reason why directed walk performed poorly on testing mushroom edibility models is as follows. 

The theorems proved in Section \ref{sect:StrategyDefinitions} imply that the capability of the directed walk on a given direction depends on the existence of a border in the direction from the randomly selected starting test case. If a border point is found, it only differs from the starting test case in one feature. This is a limitation of the capability of the strategy. This is the case for testing the mushroom edibility, where it is rare that changing just one feature of a mushroom variety will change its edibility; usually at least two features must change.

It was also observed that the random target strategy has zero capability when used for testing the NB and NB2 models of mushroom edibility, as does all three strategies when testing the SVM and SVM2 models of bank churners. The reason for the poor performances is as follows. 

The random target strategy discovers a border point when the two starting points are in different classes. If a subdomain is small, the probability of selecting a point inside it is correspondingly small. In the extreme case, when all test cases are in the same class, no border will be discovered. The NB and NB2 models of mushroom edibility classify all mushrooms in the training dataset as poisonous. Similarly, the SVM and SVM2 models of bank churners classify all credit card customers to be non-churners so no Pareto front can be discovered by any strategy. 

It is worth noting that the NB and NB2 models have the worst accuracy among all models of mushroom edibility, and SVM/SVM2 models are the worst on accuracy among the models of bank churners. They are underfit models, which means they are insufficient for classifying the input data space. Therefore, exploratory testing cannot detect the borders between subdomains. 

On average, the random walk strategy achieved the best performance on capability. It can discover a border point even if all start points are in the same class; it is only required that a border exists within walking distance from the starting point. Moreover, the Pareto front found may be different from the starting point on many features. Although its cost is not the lowest of the three, it balances capability and cost best of the three.

\begin{itemize}
\item \emph{Length of Execution Time.}
\end{itemize}

The real cost of the testing strategies in terms of the lengths of execution time required to generate a Pareto front for a classifier depends on the speed of the computer system, the time needed to invoke the classifier to classify an input data, and the number of walks to be executed. The measure of test cost in terms of the number of invocations of the classifier under test per pair of points in the Pareto front gives an abstract metric, which is independent of these factors while the real cost can be calculated with these factors as parameters by using equation (\ref{equ:Time}). To give an indication to the scale of real cost, we have run each strategy 10 times for each classifier, and each time we have executed 1000 walks and recorded the clock times spent and the sizes of Pareto fronts generated. The testing tool Morphy was run on a Windows PC with Intel Xeon x64 CPU E3-1230V5 \@ 3.40GHz and 32 GB memory. 

Table \ref{tab:TimeOfCodedClassifiers} below shows the average numbers of Pareto front pairs generated per second for various coded classifiers used in the controlled experiments. From these data, the average real cost of generating Pareto fronts of a certain size, such as 1000 pairs of points, can be easily calculated by the formula $Time = \frac{P}{RC}$, where $P$ is the size of Pareto front, $RC$ is the data of real cost given in Table \ref{tab:TimeOfCodedClassifiers}, i.e. the average number of Pareto front pairs per second. 

\begin{table}[h]
\centering
\caption{Average Number of Pairs Generated Per Second for Coded Classifiers}\label{tab:TimeOfCodedClassifiers}
\scriptsize
\begin{tabular}{|l|r|r|r|r|}
\hline 
Classifier &Directed Walk &Random Target &Random Walk	 &Average\\
\hline
Box 1 &645.93 &3059.24& 1919.92 &1875.03\\
Box 2 &2734.57 &3532.37 &2954.01 &3073.65\\
Circle 1 &1084.63 &3730.02 &2281.91 &2365.52\\
Circle 2	 &2956.22 &3591.11 &2909.88 &3152.40\\
Line 1 &2421.86 &3709.12 &2749.14 &2960.04\\
Line 2 &2434.26 &3610.46 &2985.71 &3010.14\\
Sin 1 &2133.10 &3733.23 &2880.03 &2915.45\\
Sin 2 &2500.64 &3653.87 &3090.02 &3081.51\\
Triangle 1 &601.53 &3104.84 &1853.88 &1853.42\\
Triangle 2 &2773.01 &3697.53 &2932.50 &3134.35\\
\hline
Average &2028.57 &3542.18 &2655.70 &2742.15\\
\hline 
\end{tabular} 
\end{table}

The data shows that, for coded classifiers, on average, generating a Pareto front consisting of 1000 pairs of points only took less than 0.4 seconds. The worst case, for directed walk strategy, for the same size of Pareto front was 1.66 seconds and the best case, for random target strategy, took 0.27 seconds. 

Table \ref{tab:TimeOfMLModels} shows the results of testing those real ML models used in our case studies. It gives the average number of pairs generated per second for various types of ML models using different exploratory strategies in the same experimental setup, where DW, RT and RW stand for Directed Walk, Random Target and Random Walk strategies, respectively. 

\begin{table}[h]
\centering
\caption{Average Number of Pairs Generated Per Second for Real ML Models}\label{tab:TimeOfMLModels}
\scriptsize
\begin{tabular}{|l|r|r|r|r|r|r|r|r|r|r|}
\hline
ML 	 &\multicolumn{3}{c|}{Red Wine Quality}	&\multicolumn{3}{c|}{Mushroom Edibility}	&\multicolumn{3}{c|}{Bank Churners}	&\\
Model &DW	&RT	&RW &DW	&RT	&RW &DW	&RT	&RW	&Average\\
\hline
DT	&84.46	&266.68	&203.21	&131.52	&329.42	&176.43	&278.04	&188.77	&327.33	&220.65\\
HV	&21.68	&32.13	&28.62	&22.56	&119.62	&23.43	&33.68	&35.09	&47.05	&40.43\\
KNN	&30.16	&21.95	&37.70	&36.73	&140.34	&9.23	&19.68	&47.57	&39.35	&42.52\\
LR	&272.68	&260.63	&202.64	&135.44	&307.29	&121.53	&276.27	&204.84	&526.73	&256.45\\
NB	&146.81	&156.79	&148.69	& -~-   	& -~-   	& -~-   	&166.61	&231.19	&113.20	&160.55\\
Stack	&9.03	&13.21	&10.86	&8.54	&46.29	&8.76	&10.66	&13.83	&14.53	&15.08\\
SV	&8.62	&20.31	&14.39	&11.52	&63.03	&12.09	&17.87	&18.92	&22.13	&20.99\\
SVM	&37.02	&68.94	&52.74	&120.37	&306.60	&61.98 	&107.94 &  -~- & 	-~-  & -~-\\
\hline
Avg	&76.31	&105.08	&87.36	&66.67	&187.51	&59.06	&114.69	&105.74	&155.76	&105.70\\
\hline
\end{tabular} 
\end{table}

The data shows that it took less than 10 seconds to generate Pareto fronts of 1000 pairs. In the worst case, which is when testing the Stack model of Mushroom Edibility using the directed walk strategy, it took an average of 117.11 second (less than 2 minutes) in 10 executions of the test strategies to generate 1000 pairs. The best case is when testing the Logistic Regression model LR of Bank Churners using the random walk strategy. This took less than 2 seconds to generate the same number of pairs. 

There are two machine learning models in our case study that do not have any border between classes as our exploratory testing discovered. They are the naïve Bayes model NB of mushroom edibility and the Support Vector Machine model SVM of bank churners prediction. Table \ref{tab:TimeOfNoBorder} shows the average lengths of time that the strategies completed the search for borders by taking 1000 walks to test these two models. In the worst case, it took less than 2 minutes, while in most cases it took between a fraction of a second and a few seconds. 

\begin{table}[h]
\centering
\caption{Lengths of Time (Second) to Complete Search when No Border in the Classifier}\label{tab:TimeOfNoBorder}
\scriptsize
\begin{tabular}{|l|c|c|c|}
\hline
ML Model	 &Directed Walk	&Random Target	&Random Walk\\
\hline
NB-Mushroom	&14.83	 &0.25	 &2.87\\
SVM-Bank	 &102.47 &	0.69 	&7.51\\
\hline
\end{tabular} 
\end{table}

In general, the time taken to execute a test strategy heavily depends on how fast the classifier under test is for classifying an input data. Our experiment data presented in the previous sections shows that the time to execute the strategies increases linearly with the number of walks and with the number of pairs of border points generated. Therefore, the experiment data with real machine learning models indicate that to generate a Pareto front containing 1000s of pairs, on average we only need 10s of seconds. It is highly efficient for practical uses of the strategies. 

\begin{itemize}
\item \emph{Validity of the Conclusions.}
\end{itemize}

The case studies have been conducted on datasets selected at random from a large library with each dataset representing a different type of classifier system. It is possible that the datasets chosen had special properties that had an impact of the results but this threat to validity can be eliminated by repeating the case studies on other datasets.

The case studies used a wide range of models of different types and of different quality (e.g. of different accuracy). They were constructed by using Python code selected from the Kaggle website at random. The distribution of the quality among these models may be not representative of the models in a real production environment. Thus, the statistics may be biased. However, due to the lack of data on the distributions of model quality, we are unable to eliminate such a potential bias. The way to improve this aspect is to use the test strategy in a real production environment. 

The test systems were implemented by the authors according to the formal definitions given in Section \ref{sec:TestSystem}. They were debugged and tested on a large number of test cases. A threat to the validity of the case study is the existence of bugs in the test system, which may have impact on the correctness of the data. The source code of the test systems is written in Java and freely available from GitHub for inspection. We are reasonably confident that the test system has no serious bugs. 

The process of the case studies is highly automated by executing test scripts written in Morphy's test scripting language. Manual operational errors in the conduct of the case studies can be eliminated to the highest extent. However, there may be bugs in the test script and in the Morphy testing tool. Such bugs form a threat to the validity of the conclusions drawn from the data. We believe that this threat should have a minimal impact, however, as the Morphy tool and test scripts have been tested, too. Morphy is available for download and use for free. The test scripts are also available on GitHub for download and inspection. The whole case study can be repeated easily. 

Finally, the observations made in the case studies and the conclusions drawn from the data are consistent with the observations made in the controlled experiments and what the formally proved theorems imply from the formal definitions and the algorithms. Therefore, we can confidently conclude that the conclusions drawn from the cases studies are valid and can be generalised to other machine learning models built via supervised training on datasets. 

\section{Related Work}\label{sec:RelatedWork}

The most closely related work is exploratory testing (ET). We will review the current state of research on this field, and summarise our contributions to it. We will also discuss the similarities and differences between our work and adaptive random testing (ART), fuzz and data mutation testing, metamorphic testing (MT), and search-based testing (SBT). Finally, since the work of this paper is partially inspired by the traditional testing method of domain testing, we will also briefly discuss the applicability of that method to machine learning models. 

\subsection{Exploratory Testing}

ET has been widely applied to many types of software systems, but most successfully to GUI-based systems; see, for example, \citep{whittaker2009exploratory}. \citet{Pfahl_et_al2014} reported an online survey of Estonian and Finnish software developers and testers on their uses of exploratory testing in practice, revealing that a majority used it intensively for usability-critical, performance-critical, security-critical and safety-critical software. However, as far as we know, there is no report on the systematic application of ET for testing AI applications.  

Research on ET exists that evaluates its fault detection effectiveness and efficiency, including reports on its effectiveness in practice. Itkonen et al. (\citeyear{ItkonenAndRautiainen2005, Itkonen_et_al_2007,  ItkonenAndMantyla2014}) were amongst the first. They used students as subjects to compare ET with traditional software testing techniques that based on pre-designed test cases (TCT). Through replicated experiments, they found that ET had the same effectiveness in fault detection but greater time-efficiency because less design effort was needed. Moreover, TCT produces more false-positive defect reports than ET.

\citet{Afzal_et_al2014} conducted a controlled experiment with 24 practitioners and 46 students who performed manual functional testing to compare the effectiveness of exploratory testing against traditional test techniques. Unlike Iktonen et al., they reported that ET found significantly more defects, including those at varying levels of difficulty, type and severity. Also unlike Iktonen et al., they did not report that ET reduced the number of false-positive defect reports.

However, both of these experiments were conducted on traditional software and since AI applications have different failure modes and faults, it is unclear whether ET is effective and efficient for machine learning applications. 

Since ET is used as a manual testing method, research on it has mostly focused on the human factors that alter effectiveness and efficiency. An industrial case study by \citet{GebizliAndSozer2017} with 19 practitioners of different educational backgrounds and experience levels show that both factors affects efficiency but only experience affects the number of critical failures detected. \citet{Micallef_et_al2016} found that trained testers employed different types of exploratory strategies than untrained testers. The trained testers were more effective at finding input validation errors, while untrained testers tended to uncover mostly content bugs. The two groups were however equally effective at detecting logical bugs or functional UI bugs.

\citet{Shoaib_et_al2009} found that people with extrovert personalities are more likely to be good exploratory testers. \citet{Itkonen_et_al2013} found that exploratory testers applied their knowledge for test design and failure recognition differently. \citet{Martensson_et_al2021}, after interviewing testers in six companies, identified nine key factors that determine the effectiveness of exploratory testing in an organisation and proposed a simple model for improving it.

On the automation of ET, \citet{Eidenbenz_et_al2021} employed artificial intelligence techniques to predict test cases that are likely to cause failure in testing an industry control software. \citet{Makondo_et_al2016} used neural networks to train test oracles to help the analysis of test results. Research has also been reported on the development of test environments to support exploratory testing. For example, ARME enables the automatic refinement of system models based on recorded testing activities of test engineers \citep{GebizliAndSozer2016}. Tapir supports team collaboration in exploratory testing and reconstruction of system models \citep{Bures_et_al2018}. However, as far as we know, there is no work in the literature that automates exploratory strategies as we have done, even though many exploratory strategies have been documented in the literature such as those by \citet{whittaker2009exploratory} and \citet{Hendrickson2013}. 

Our main contributions to exploratory testing are to apply it to machine learning applications and to automate it. By identifying the discovery of boundary values as a specific goal of testing, we demonstrated how the elements of ET can be formalised and implemented in the datamorphic testing framework to achieve test automation. Our approach can be summarised as follows.

Firstly, test design is formalised and implemented by a set of datamorphisms. Together with a test executor test morphism, these datamorphisms form a test system for exploratory testing. We introduced the notion of complete exploratory test systems, developed a systematic way to construct exploratory test systems for feature-based classifiers, and proved that such exploratory test systems are complete so that they ensure the whole data space of the model can be explored. 

Secondly, steering strategies are formalised and implemented as algorithms that invoke the datamorphisms and the test executor. We then formally proved that the strategies are correct; that is, they always terminate and produce Pareto fronts that represent the borders between classes. In other words, these strategies always achieve the goal of ET: to discover the borders between classes defined by the machine learning model under test. 

Finally, the strategies have been implemented in the automated datamorphic testing tool Morphy \citep{AITest2020, AITest2020TR}. We have also conducted empirical evaluations of the strategies to determine their capability and cost through controlled experiments and case studies. The results demonstrated that the approach can discover the borders between classes in a cost efficient way. The data also provide insight into the factors that affect capability and cost for each test strategy. This can be used to select appropriate parameters for appropriate strategies for each classification application. 

\subsection{Random and Adaptive Random Testing}

Generally speaking, random testing (RT) is a software testing method that selects or generates test cases through random sampling over the input domain or a profile of the software under test (SUT) according to a given probability distribution \citep{Hamlet1994}. As discussed in \citep{ZhuHallMay1997}, RT techniques can be classified into two types: representative and non-representative. 

The representative type uses the probabilistic distribution on the input domain as the input distribution for the SUT. One approach is to sample at random the operation profile of the software under test \citep{Myers2011}. Another approach is to develop a Markov model of the human computer interaction process and use it to generate random test cases \citep{Whittaker1994}. Although representative RT works well for fault detection in simulation-based experiments \citep{DuranAndNtafos1984, HamletAndTaylor1990, Tsoukalas_et_al1993, Ntafos1998}, its most compelling advantage is that test results naturally lead to an estimate of software reliability. However, such random testing requires a much larger number of test cases to achieve the same level of fault detection ability in comparison to test methods where the test cases are purposely designed. 

In contrast, the non-representative type of RT uses a distribution unconnected to the operation of the software. The major subtype of ART methods, for example, spread test cases evenly over the entire input space \citep{TYChen2001, TYChen2004, TYChen2007, TYChen_et_al2010} and experiments show that they improve both fault detection ability \citep{TYChen2007b} and reliability \citep{LiuAndZhu2008}. Even spread over the input space can be achieved by manipulating randomly generated test cases. Many such manipulation algorithms (called ``strategies" in the literature) have been developed and evaluated, including mirror \citep{TYChen2004MirrorAR}, balance \citep{TYChen2007BalancingART}, distance \citep{Huang2020}, filter \citep{KPChan2005FilterART}, lattice \citep{Mayer2005LatticebasedAR}, partition \citep{Mao2020}, etc. In a recent comprehensive survey of ART, \citet{Huang2020SurveyART} classified these techniques into Select-Test-From-Candidates Strategies, Partitioning- Based Strategies, Test-Profile-Based Strategies, Quasi-Random Strategies, and their combinations (called Hybrid-Based Strategies). Even spread can also be achieved with evolutionary computing algorithms, as discussed later in the subsection on search-based testing. 

A common feature of these ART algorithms for test case generation is that the new test cases are generated or selected based on the positions of existing test cases in the input space. This is similar to the so-called steering feature of exploratory testing. However, none of them uses the information revealed in test executions. In fact, the generation and/or selection of new test cases in ART strategies do not require the execution of the software under test at all. Of course, the most fundamental difference between ART and ET is that ART does not aim to discover the system's behaviour although evenly spreading the test cases may help indirectly. Another difference is in test design, which is the selection of a probability distribution on the input domain for ART and a decision on how to change the test cases for ET.

\subsection{Fuzz and Data Mutation Testing}

Datamorphic testing evolves from data mutation testing (DMT) \citep{DataMutationJournal2009} and its integration with metamorphic testing \citep{Zhu2015JFuzz}. Data mutation testing was proposed by \citet{DataMutationWorkshop2006} to generate realistic test cases that are structurally complex, such as those for software modelling tools. The basic idea is to develop a set of operators that transform existing test cases (called \emph{seed test cases}) to new test cases (called \emph{mutant test cases}). These operators were originally called data mutation operators, but were renamed as \emph{datamorphisms} in \citep{Datamorphic2019}. \citet{DataMutationJournal2009} also proposed that data mutation operators (i.e. datamorphisms) can indicate the correctness of the program on mutant test cases. Metamorphic relations associated with data mutation operators were formally defined in \citep{Zhu2015JFuzz} as \emph{mutational metamorphic relations}, and called \emph{metamorphisms} in datamorphic testing \citep{Datamorphic2019}. The uses of seed makers, datamorphisms and metamorphisms in one general purpose testing tool to achieve test automation was first reported in \citep{Zhu2015JFuzz}. 

Data mutation testing has similarity to \emph{fuzz testing}; see, for example, \citep{FuzzTestBook2007}. However, mutation testing emphasises an engineering process of developing data mutation operators that can be used to generate meaningful and realistic test cases for the software under test, while fuzz testing randomly makes a change without first determining whether the mutants are meaningful and realistic or not. A datamorphic test system can include either random or purposeful datamorphisms or even a combination of both. Most importantly, datamorphic testing recognises other types of test morphisms and uses them to achieve test automation at a high level of strategy and test process \citep{Datamorphic2019, AITest2020}. 

\subsection{Metamorphic Testing}

Metamorphic testing was proposed by \citet{MetamorphicTest1998} to use metamorphic relations to check test results and to generate test cases. A metamorphic relation is a relation on inputs and outputs of multiple test cases. Theoretically speaking, metamorphic relations are axioms about the software under test presented in a special form as axioms that contain multiple test cases. Such axioms can be specified in algebraic specification languages. For example, a metamorphic relation $\forall x, y.(x + y = y + x)$ on integer values of $x$ and $y$ can be written in all algebraic specification languages such as SOFIA \citep{SOFIA2014}. 

Algebraic specifications have been used for test automation since the early 1980s. They have been developed for testing procedural programs \citep{GMH81,BGM91}, object oriented programs \citep{DoF94, HuS96, TACCLE2001, BlackAndWhite98}, component-based systems \citep{KongZZ07,YuKZZ08}, and more recently for service-oriented systems \citep{MonicTest2016}. The research on metamorphic testing demonstrated that such axioms can be useful for testing software even if they do not form a complete set of axioms, though the latter are often required by test tools that automate software testing from algebraic specifications \citep{TACCLE2001, BlackAndWhite98}. 

The main difficulty of applying metamorphic testing is to define the metamorphic relations for the software under test. This is because metamorphic relations are in fact definitions of the semantics of the application. \citet{Zhu2015JFuzz} proposed a feasible engineering solution via the integration of data mutation testing with metamorphic testing through mutational metamorphic relations (i.e. metamorphisms). This approach is further developed into datamorphic testing \citep{DatamorphicTR2019,AITest2020TR}. The test automation environment Morphy shows that the approach can be efficiently implemented and applied. However, datamorphic testing is more general than metamorphic testing. It may contain test morphisms other than metamorphisms. It can also be applied without metamorphic relations as demonstrated by the case study reported in \citep{AITest2020} and the exploratory strategies studied in this paper, while metamorphic relations is essential for metamorphic testing \citep{MetamorphicTestingSurveyChen2018}. 

Research on testing AI applications has been active in recent years \citep{AST2018Proc, AITest2019Proc, AITest2020Proc}. Metamorphic testing is one of the most popular approaches to testing machine learning applications. The testing of driverless vehicles is one interest application; see for example, \citep{DeepTest2018, Zhou2019DriverlessCar}. These works demonstrate that synthetic test cases can find many erroneous behaviors under different realistic driving conditions, many of which led to potentially fatal crashes in three top performing DNNs in the Udacity self-driving car challenge. Most existing testing techniques for DNN-driven vehicles are heavily dependent on the manual collection of test data under different driving conditions. This is prohibitively expensive as the number of test conditions is huge. The works by \citet{DeepTest2018} and \citet{Zhou2019DriverlessCar} also show that the metamorphic approach can be cost efficient. 

Metamorphic testing has also been applied to testing clustering and classification algorithms. \citet{TestingClassifier2011} developed a set of metamorphic relations as test oracles for testing such machine learning alrotihms. \citet{Yang2019} reported a case study on the use of metamorphic relations to test a clustering function generated by the data mining tool Weka. 

It is interesting to observe that datamorphisms are actually used in these cases. For example, DeepTest automatically generates test cases that leverage real-world changes in driving conditions like rain, fog, lighting conditions, etc. via image transformations \citep{DeepTest2018}. Metamorphic relations are defined based on such image transformations and used to detect erroneous behaviours. \citet{Datamorphic2019} reported a case study on the testing of four real industry applications of face recognition. They used feature-editing operators like changing the subject's age, gender, skin tone, make-up etc., to generate synthetic test cases from existing pictures. In \citet{TestingClassifier2011} and \citet{Yang2019}'s work, manipulations of datasets were used to test clustering and classification algorithms. The transformations of images and manipulation of datasets are actually datamorphisms. 

In general, metamorphic testing differs from the work of this paper because it belongs to confirmatory testing, i.e. it checks that the metamorphic relations are satisfied, rather than discover the behaviour of the software under test. New test cases are usually generated based on existing test cases, where the new test cases are called \emph{follow-up test cases} in the literature. Thus, metamorphic testing replicates the feature of steering in the exploratory testing process. When metamorphic testing is combined with data mutation testing as in the examples discussed above, test designs can be represented in the form of datamorphisms. 

\subsection{Search-Based Testing}

Search-based testing regards testing as an optimization problem \citep{SearchBasedSESurveyHarman2012, SearchBaseTestingSurvey2015} to maximize the test effectiveness or test coverage by searching on the space of test cases. Search-based testing techniques can also be applied to ART by considering even spread test cases as the goal of optimisation. 

Genetic algorithms, and other algorithms within evolutionary computing, are often employed to realize such optimizations. In the evolution process, new test cases are generated from existing ones in the population through mutation, crossover and randomisation operators to improve the fitness of the population. Therefore, genetic algorithms provide a steer, just as exploratory test algorithms do. Test design in ET can be represented in the form of the mutation and crossover operators of evolutionary computing, but this fact is rarely studied and used explicitly. Depending on what the fitness metrics encode, a new test case may be executed if it requires information about the program's behaviour. Therefore, search based testing has most of the essential features of ET, but the key difference is in their goals: search-base testing aims to optimise, while ET aims to discover. 
		
\subsection{Domain Testing}		

The work reported in this paper is inspired by the domain testing method, in which the input space of the software under test is decomposed into a number of sub-domains according to either the specification or the program code of the software under test. Test cases are then selected on or near to the borders between sub-domains. Domain errors are very common programming errors; for example, sub-domains could be missing and/or the boundaries between sub-domains could be incorrectly implemented. The method of domain testing aims to detect such errors.  

Research into and practical uses of domain testing can be traced backed to the late 1970s and early 1980s. For example, \cite{WhiteAndCohen1980} studied how programming errors are related to domain modifications and proposed a strategy to select $N$ test cases on the borders and 1 test case near to the borders of the subdomains in order to detect boundary parallel shift errors for linear borders, where $N$ is the dimension of the input space. Similarly, \citet{Clarke_et_al1982} proposed a strategy to select $N$ test cases on the border and $N$ test cases nearby. They proved this strategy is capable of detecting both a parallel shift and a rotation of the linear boundary. \citet{Afifi_et_al1992} proposed a strategy that selects $N+2$ test cases on and nearby to each border of a subdomain. By applying Zeil's theory of perturbation testing \citep{Zeil1983,Zeil1989}, they proved that the strategy is capable of detecting linear errors of boundaries defined by non-linear functions \citep{ZeilAfifiWhite1992}, where linear errors are linear transformations of the boundary function. A survey of the research on domain testing in the 1980s and 1990s can be found in \citep{ZhuHallMay1997}. Since then little progress has been reported in the literature. 

Domain testing is a typical traditional scripted and confirmatory testing method that derives a complete test set before testing is actually executed and the test results are compared with predetermined expected outputs. There is no immediate execution of test cases after generation, no steering of the testing using the output of previous tests, and the design of test cases is not focused on the variations in the behaviour space. The purpose of domain testing is to confirm that the borders between sub-domains are correctly drawn. 

\begin{figure}[hbtp]
	\centering
	\includegraphics[width=10cm]{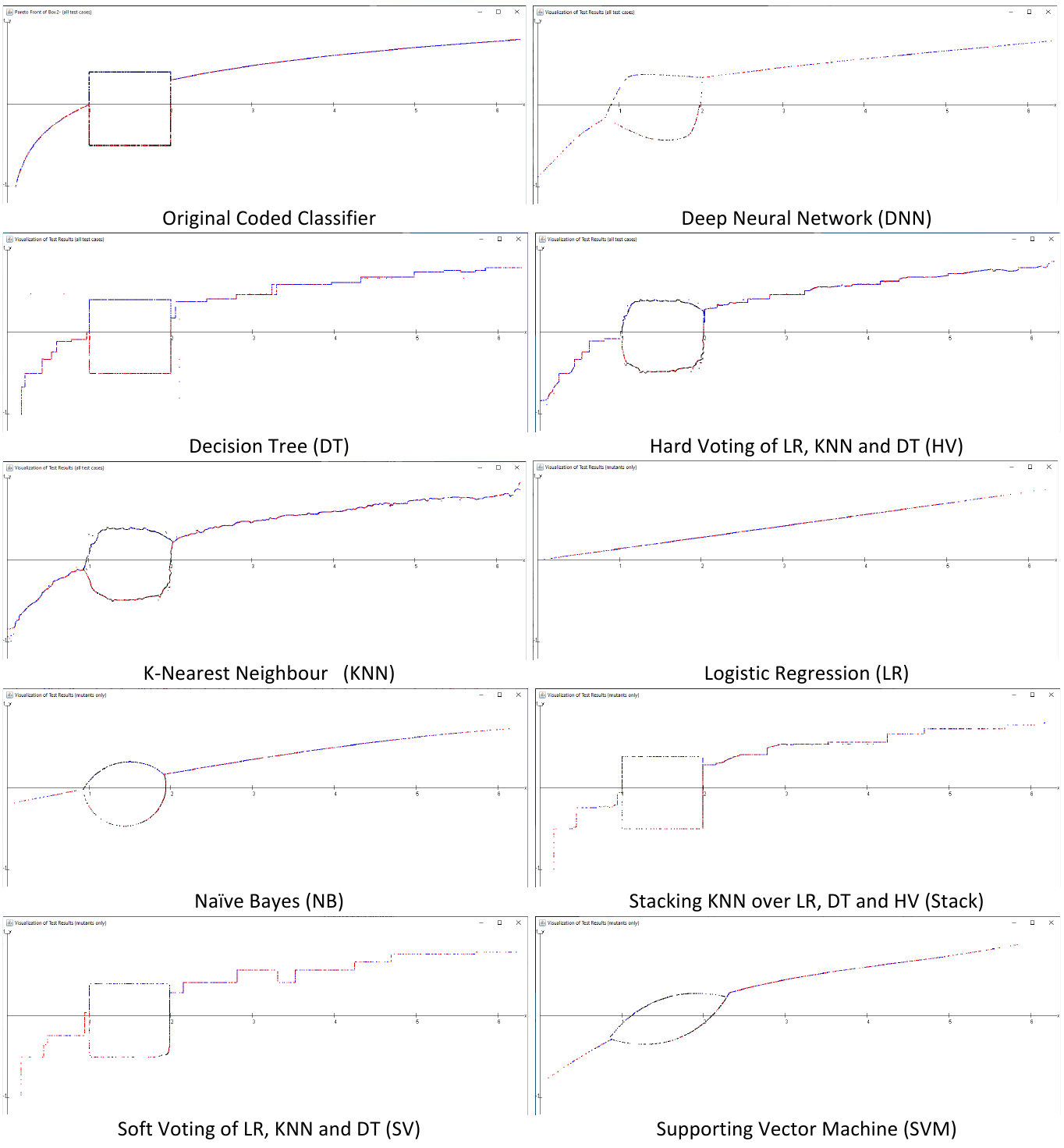}
	\caption{Subdomain Boundaries of Various Machine Learning Models of Box 2}
	\label{fig:MLModels}
\end{figure}

The research on domain testing demonstrated that programming errors often manifest themselves as changes in the boundaries between sub-domains. Test cases on or near those borders are effective at detecting these errors. Errors in a machine learning model must also occur around the borders between sub-domains. It is very useful to know where borders are actually drawn between sub-domains defined by the model. However, the existing domain testing techniques cannot easily be applied to classifiers built from machine learning techniques, because both the expected border as specified and the implemented border as coded are usually not available. Moreover, the domain errors of a machine learning model could be much more complicated than traditional programming errors. Figure \ref{fig:MLModels} visualises the Pareto fronts of the original coded classifier Box 2 and various machine learning models built from a dataset obtained by a random sampling of Box 2 on 5000 points. It shows that the boundary errors of these machine learning models are highly complicated and significantly different from the coding errors assumed in the research on domain testing. 

\subsection{Summary of The Comparison}

To summarise the differences between the proposed approach and the related testing methods discussed above, we contrast these methods on the four essential elements of ET; see Table \ref{tab:ComparisonTestMethod}. 

\begin{table}[h]
\caption{Comparison of Related Testing Methods}\label{tab:ComparisonTestMethod}
\begin{small}
\begin{center}
\begin{tabular}{|l|c|c|c|c|}
\hline
\textbf{Test Method} & \textbf{Test Design} & \textbf{Execution} & \textbf{Learning} & \textbf{Steering} \\ \hline\hline 
Proposed Method (ET)	&\cmark	&\cmark	&\cmark	&\cmark\\\hline
Fuzz Testing 		&\xmark 		&\cmark 	&\xmark 		&\xmark \\ \hline
Data Mutation Testing &\cmark &\cmark &\xmark 	&\xmark\\ \hline
Adaptive Random Test	 &\xmark		&\xmark		&\xmark		&\cmark\\\hline 
Metamorphic Testing	&\xmark/\cmark	 &\cmark		&\xmark	&\cmark\\\hline 
Search-Based Testing	&\xmark/\cmark 	&\xmark/\cmark	&\xmark	&\cmark\\\hline 
Domain Testing 		&\xmark 		&\xmark 		&\xmark 		&\xmark \\\hline
\end{tabular} 
\end{center}
\end{small}
\end{table} 

\section{Conclusion and Future Work} \label{sec:Conclusion}

The Pareto fronts generated by the algorithms studied in this paper contain a huge amount of information about behaviour of the ML models and we are exploring their potential benefits.  First, the Pareto front brings a number of possible new ways to analyse and improve ML models. We are currently working on how to use Pareto fronts in the measurement and comparison of ML model's performance. 

Another possible benefit is in explaining and/or interpreting the output of a ML model, which has been an active research topic recently; see, for example, \citep{Linardatos_et_al2021, Molnar2021}. Given a Pareto front that represents the borders between classes, a model's classification of a data point could be explained and interpreted, for example, by contrasting it against the nearest points on the surrounding borders and the distance of the point to these boundary points.  

The test cases contained in a Pareto front seem also useful to improve model's performance. For example, when the training data is imbalanced, those Pareto front points in minority classes could be used as additional synthetic training data similar to the SMOTE technique \citep{Fernandez_et_al2018}.  

How to present the information contained in Pareto fronts is another interesting topic for future research. For example, the visualisation of Pareto fronts of various ML models in Figure \ref{fig:MLModels} provides a clear view of their behaviours. An interesting research question for future work is how to visualise models on higher dimensional data spaces. There are a few existing techniques to visualise higher dimension spaces, such as contour charts for visualising 3D models on a 2D space. The effectiveness of such techniques needs to be tested with empirical studies. 

There are also many possible variations of the strategies proposed and studied in this paper. In particular, the strategy's algorithms do not need a measurement of the distance between two test cases. However, a distance measurement can be used to decide when to terminate the refinement loop, thereby improving the effectiveness. We are conducting further research on strategies that improve both cost and capability. 

This paper focused on multi-class feature-based classifiers whose data spaces are symbolic or numerical values of features and each instance of data is assigned with a single label. A valuable topic for further study is to extend the approach to classifiers on other types of data spaces, such as time series, images, audio and video data, and natural language texts, etc. 

Moreover, a machine learning classifier can be:
\begin{itemize}
\item \emph{single labelled}, where one label is assigned to each instance in the data space, thus the classes are non-overlapping sub-domains; 
\item \emph{multi-labelled}, where multiple labels can be assigned to an instance, thus, an data point may belong to multiple classes and the sub-domains may overlap with each other; and
\item \emph{hierarchical}, where labels are organised in a hierarchical structure thus the sub-domain of a class can be divided into a number of subclasses, etc. 
\end{itemize} 

In this paper, we have focused on single labelled classifiers. It will be interesting to investigate how to extend the theory and their algorithms to multi-labelled and hierarchical classifiers. 

More generally, classifiers are classification models that map from a data space to a set of categorical labels, while predictors are models of functions of continuous or ordered numerical values. Such predictors are often constructed through regression analysis and used for numeric predictions \citep{DataMiningTextBook}. It will also be interesting to adapt the approach studied in this paper to predictors. 

\section*{Acknowledgement}

The work reported in this paper is partially supported by the 2020 Research Excellence Award of Oxford Brookes University. The authors are grateful to the members of the AI Software Engineering research group at the School of Engineering, Computing and Mathematics, Oxford Brookes University, for their comments on the drafts of the paper, and discussions at the reading group's activities. 

\section*{References}

\bibliographystyle{model2-names}\biboptions{authoryear}
%\bibliography{DatamorphicTestingLiterature}

\newpage
\section*{Appendix A. Data of The Controlled Experiments}

Table \ref{tab:ResultsOfDirectedWalkExperiment} shows the summary data from controlled experiments with the directed walk strategy. For a particular number of test cases, indicated by column \emph{\#Seeds}, generated at random from the uniform distribution, we walk 20 steps in one direction using the upward datamorphism. The columns \emph{Avg \#Runs} and \emph{Avg \#Mutants} give the average number of test executions of the subject program under test and the average number of mutant test cases generated ie the number of test cases on the border of the clusters.

\begin{table}[htbp]
\caption{Experiments Data of The Directed Walk Strategy} 
\label{tab:ResultsOfDirectedWalkExperiment}
\begin{center}
\includegraphics[width=13cm]{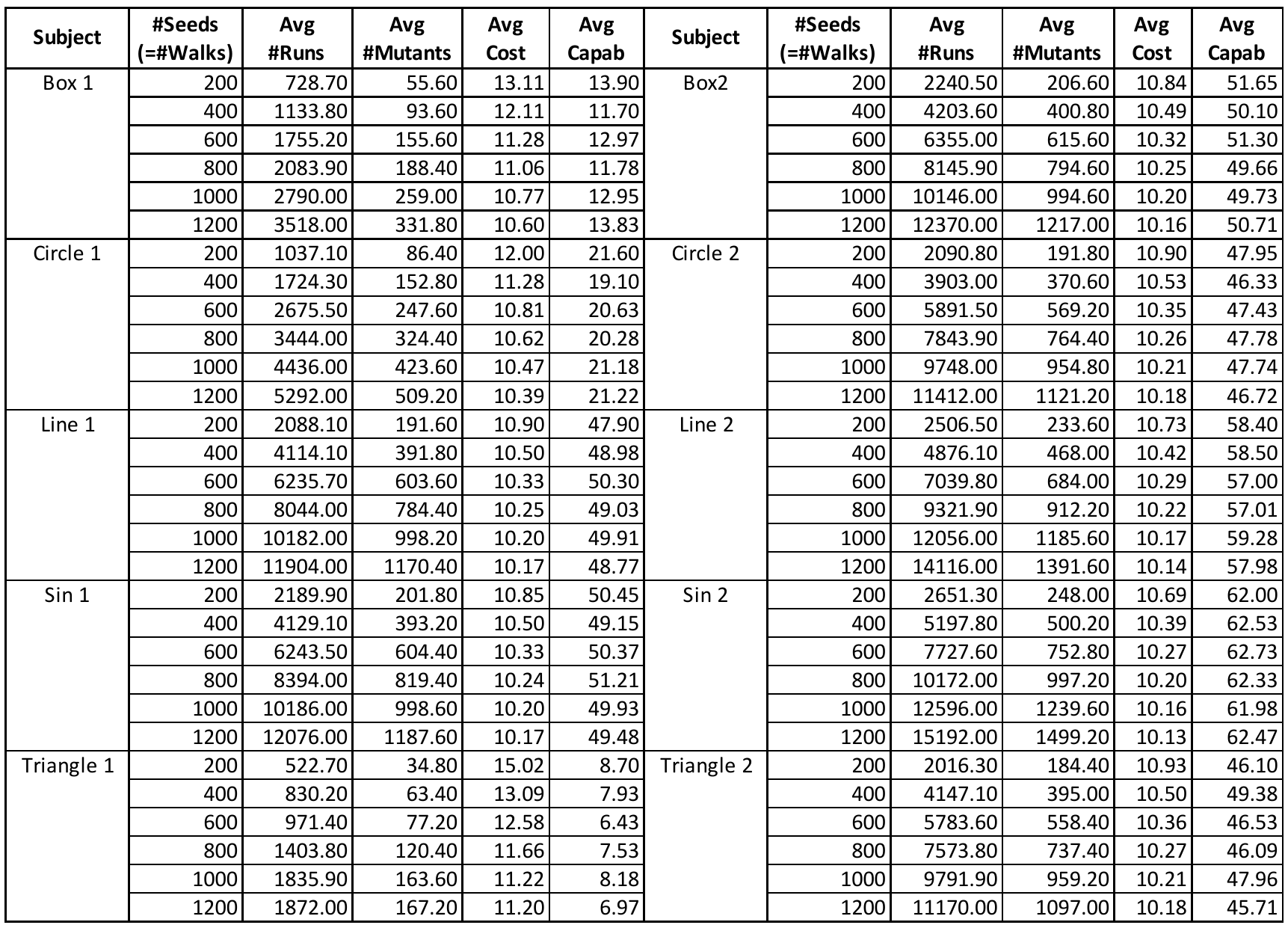}
\end{center}
\end{table}

Table \ref{tab:ResultsOfRandomWalkExperiment} gives the results of the first experiment with the random walk strategy in which the set of test cases is fixed but the number of walks varies. It shows the average number of test executions and the average number of mutant test cases for each number of random walks for each subject. 	

\begin{table}[htbp]
\caption{Experiments Data of The Random Walk Strategy with Variable Number of Walks} 
\label{tab:ResultsOfRandomWalkExperiment}
\begin{center}
\includegraphics[width=13cm, height=8cm]{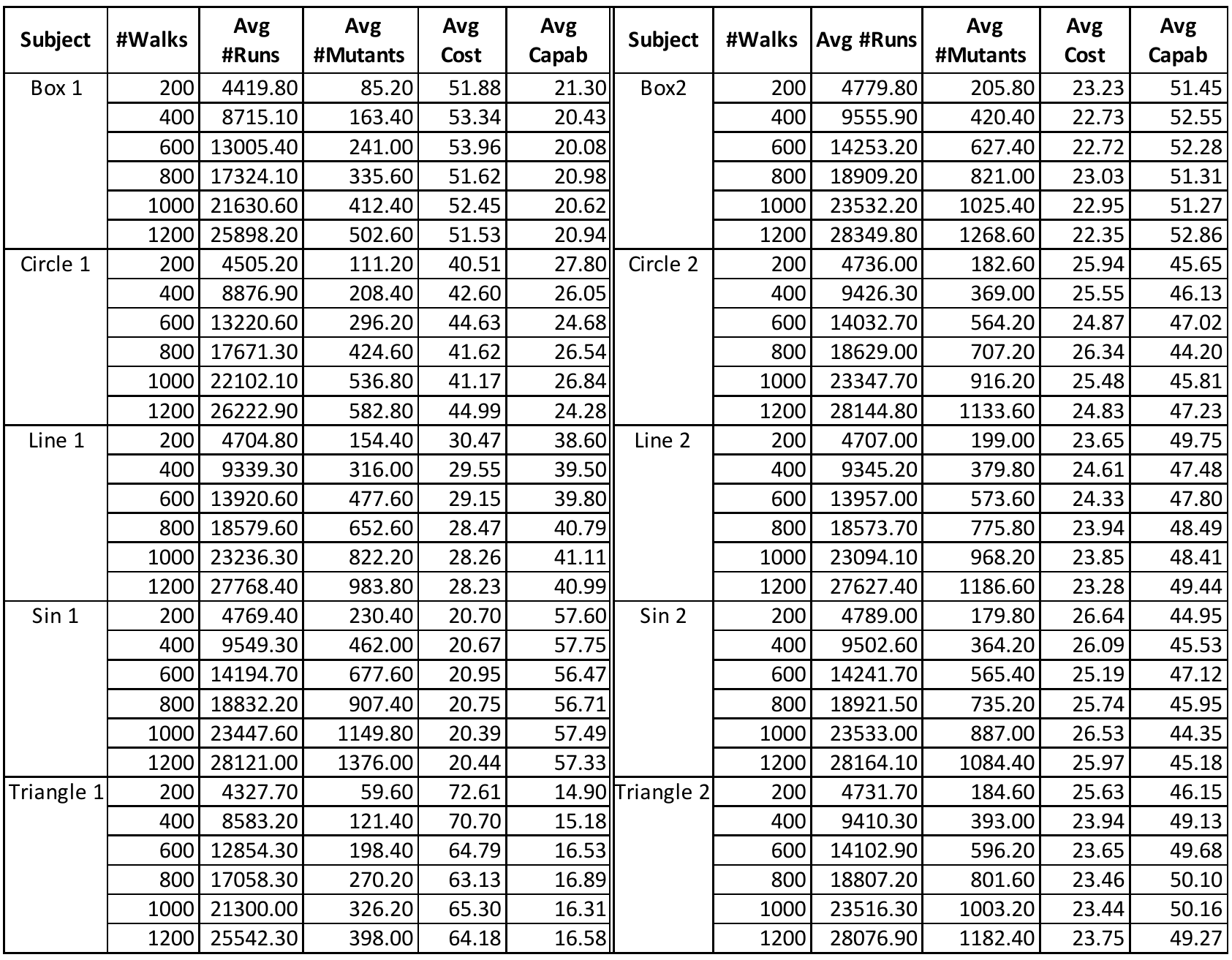}
\end{center}
\end{table}

In the second set of experiments on random walk strategy, the number of walks (800) was fixed, but the number of seed test cases varies. Table \ref{tab:ResultsOfRandomWalkTCExperiment} shows the results of experiments. 

\begin{table}[htbp]
\caption{Experiments Data of The Random Walk Strategy with Variable Number of Test Cases} 
\label{tab:ResultsOfRandomWalkTCExperiment}
\begin{center}
\includegraphics[width=13cm, height=8cm]{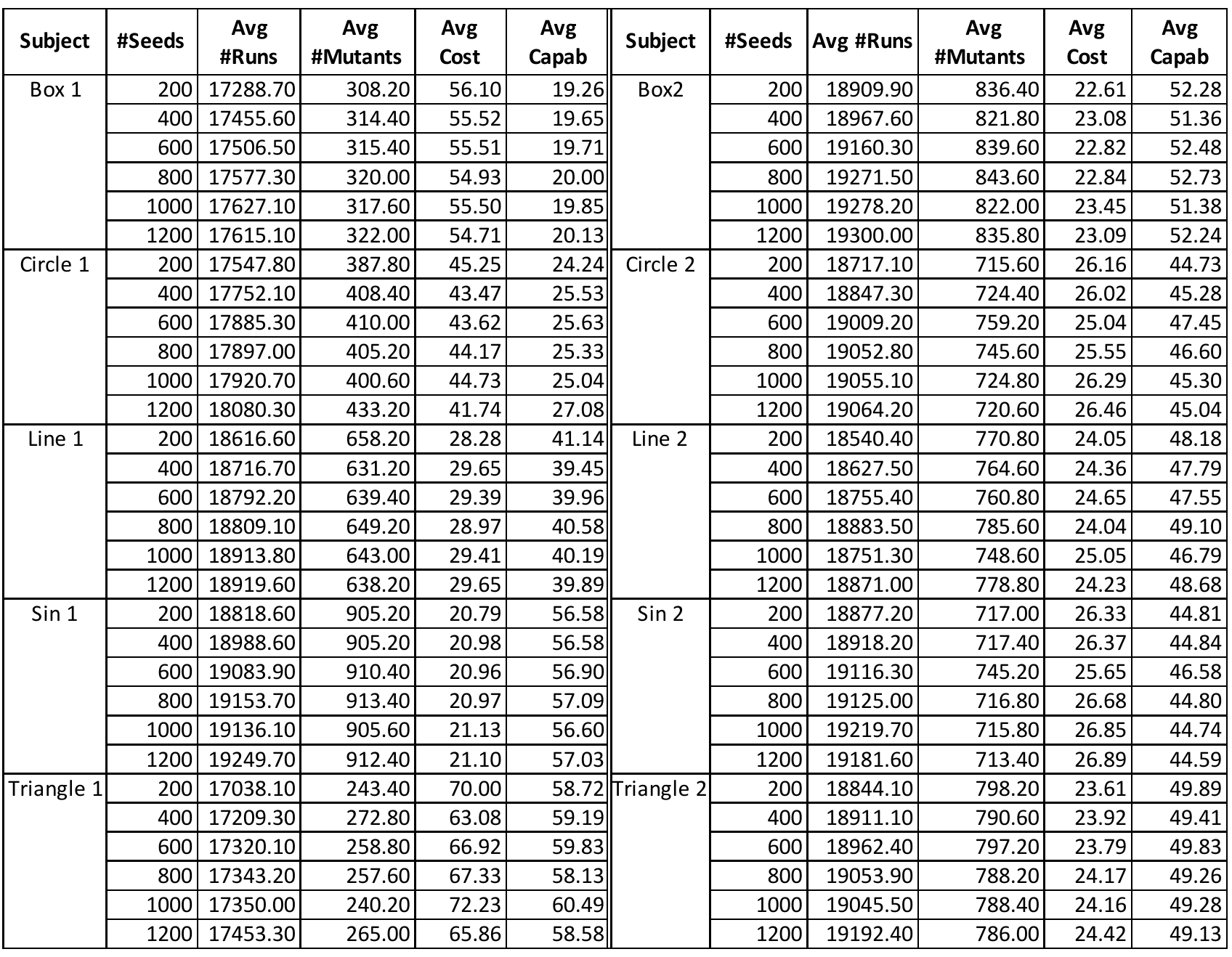}
\end{center}
\end{table}

The random target strategy only has one parameter: the number of pairs of test cases selected at random. The experiments are conducted with this parameter ranging from 200 to 1200 and the results are given in Table \ref{tab:ResultsOfAimedWalkExperiment}. 

\begin{table}[htbp]
\caption{Experiments Data of The Random Target Strategy} 
\label{tab:ResultsOfAimedWalkExperiment}
\begin{center}
\includegraphics[width=13cm, height=8cm]{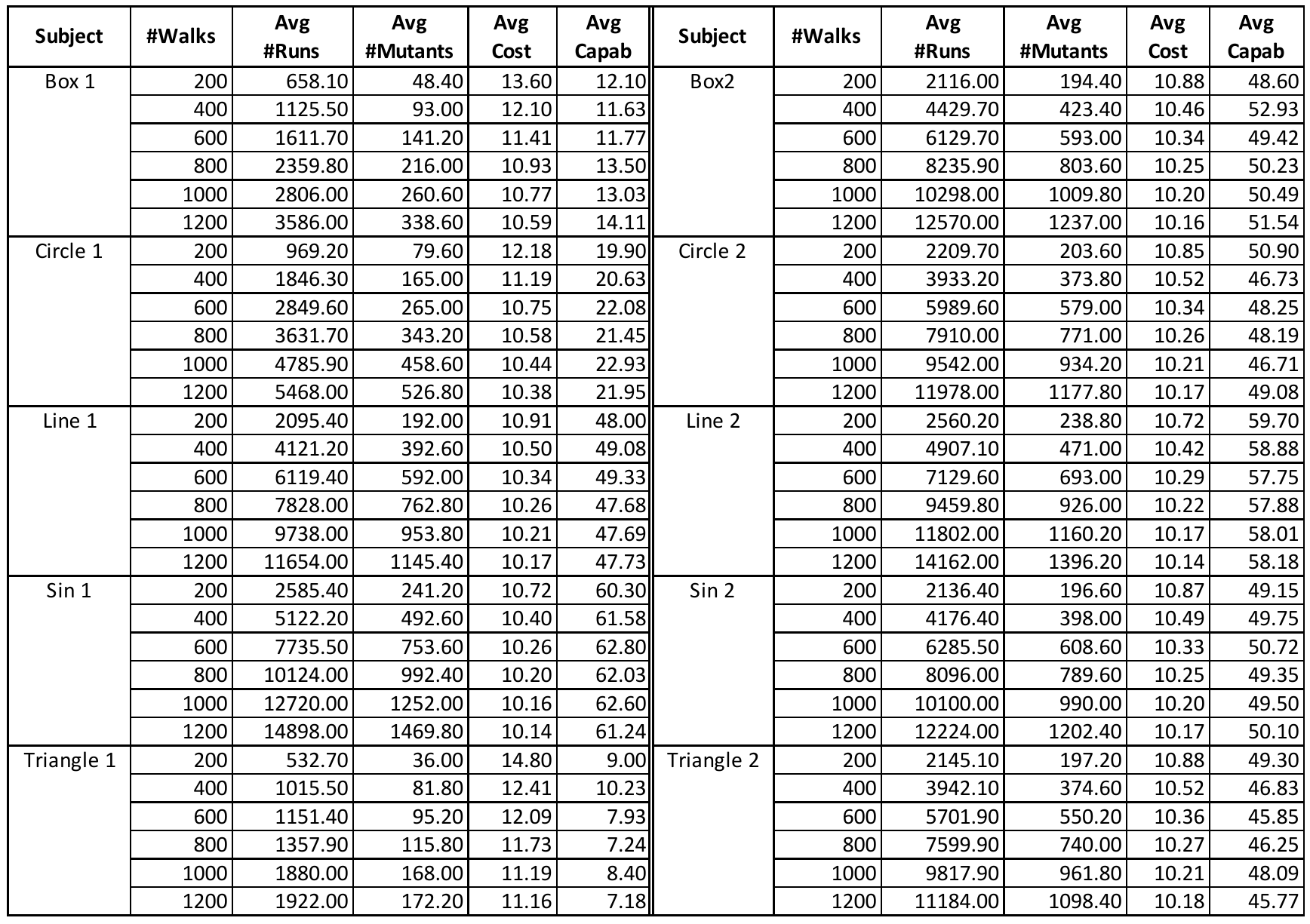}
\end{center}
\end{table}

\newpage
\section*{Appendix B. Data of The Case Studies}

\subsection*{B.1. Accuracy of the Machine Learning Models}

Table \ref{tab:ModelQualityInfo} shows the quality of each model. Column \emph{Accuracy} is the percentage of the items in the dataset that are correctly classified by the model. Column \emph{Cross-validation score} is the average score of 10 fold cross-validations of the model. Column \emph{Accuracy on test data} is as for the first column but restricted just to the test set. Column \emph{Time} is the total length of time in seconds spent on training the model, cross-validating the model and calculating the accuracies of the model on the whole data set and on the test set. It gives an indication of the time efficiency of the model.

\begin{table}[htbp]
	\caption{Model Quality}	\label{tab:ModelQualityInfo}
	\begin{center}
	\includegraphics[width=13cm]{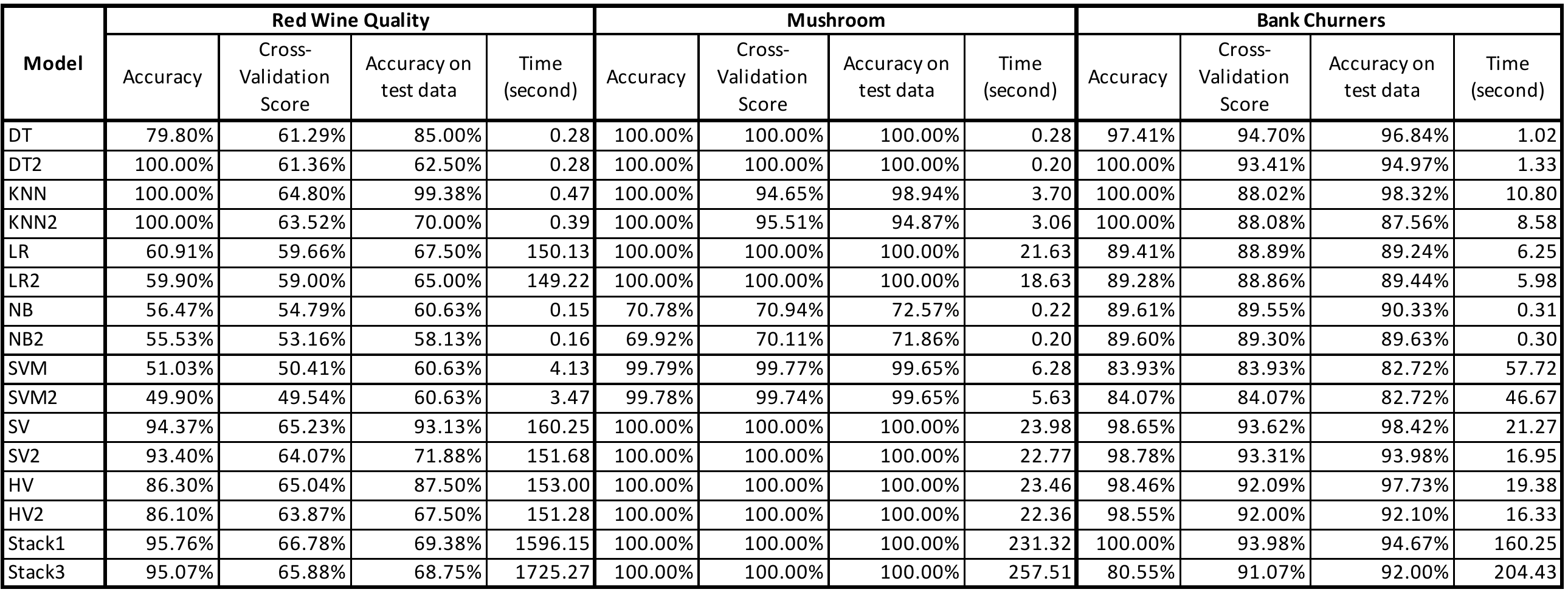}
	\end{center}
\end{table}

\newpage
\subsection*{B.2. Data of The Case Study on Red Wine Quality}

Table \ref{tab:RedWineRuns}, \ref{tab:RedWineMutants}, \ref{tab:RedWineEffectiveness} and \ref{tab:RedWineCapability} give the average numbers of runs, mutants, cost and capability of   testing the red wine quality classification models using three strategies.  

\begin{table}[htbp]
	\caption{Average Number of Runs}	\label{tab:RedWineRuns}
	\begin{center}
	\scriptsize{(a) Random Target Strategy}\\
	\includegraphics[width=13cm, height=2.3cm]{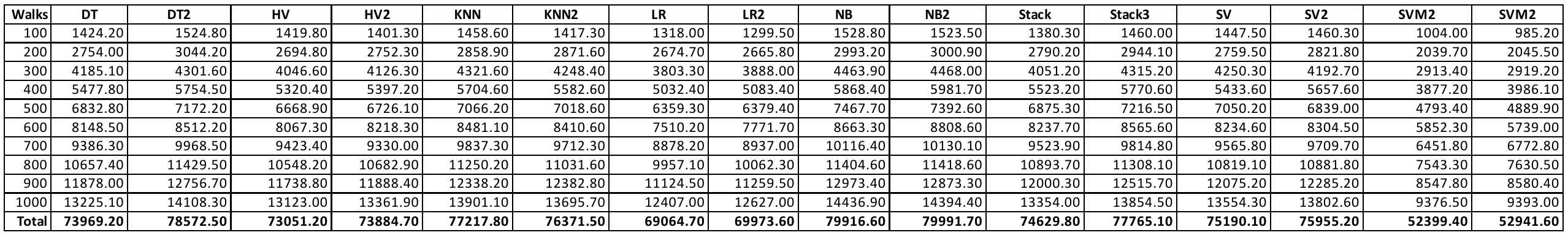}\\
	\scriptsize{(b) Random Walk Strategy}\\
	\includegraphics[width=13cm, height=2.3cm]{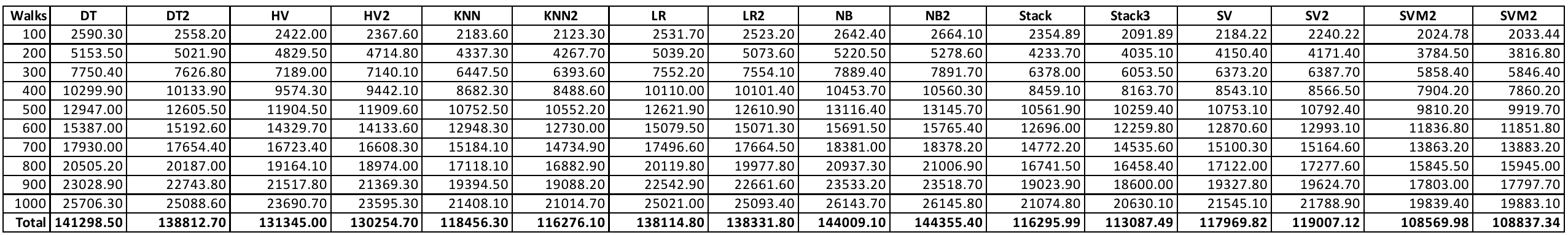}\\
		\scriptsize{(c) Directed Walk Strategy}\\
	\includegraphics[width=13cm, height=2.3cm]{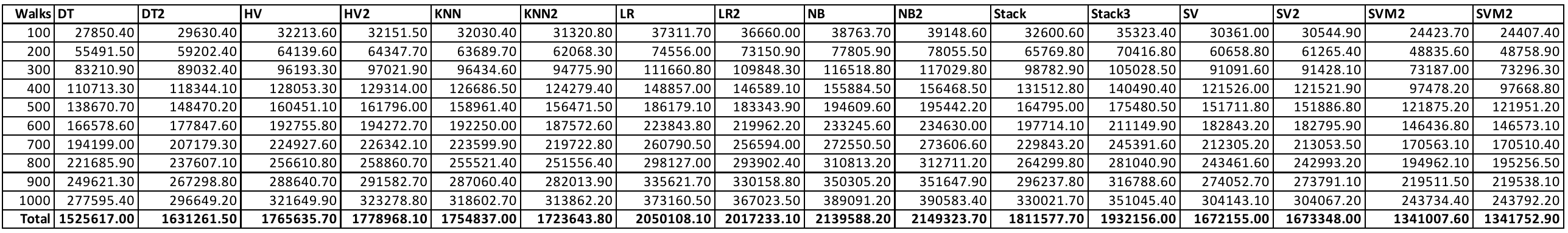}\\
	\end{center}
\end{table}

\begin{table}[htbp]
	\caption{Average Number of Mutants}	\label{tab:RedWineMutants}
	\begin{center}
	\scriptsize{(a) Random Target Strategy}\\
	\includegraphics[width=13cm, height=2.3cm]{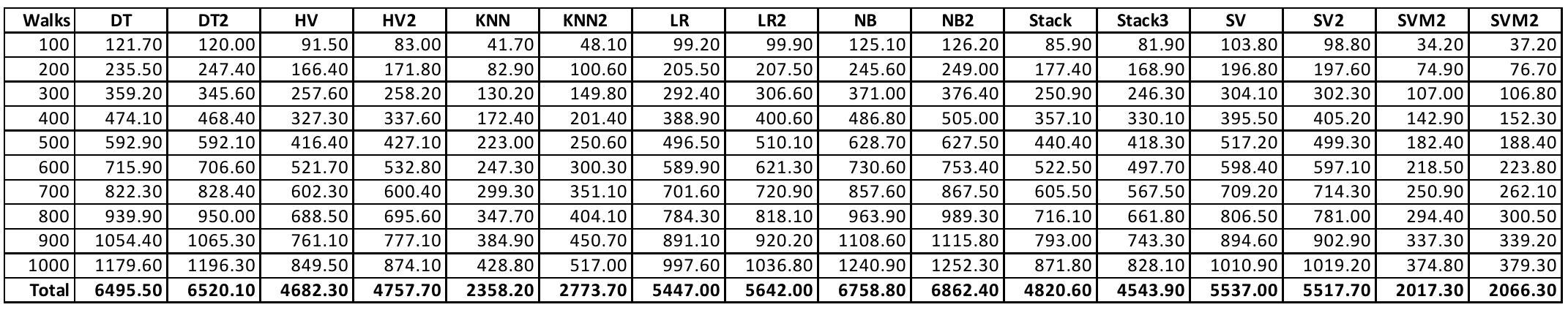}\\
	\scriptsize{(b) Random Walk Strategy}\\
	\includegraphics[width=13cm, height=2.3cm]{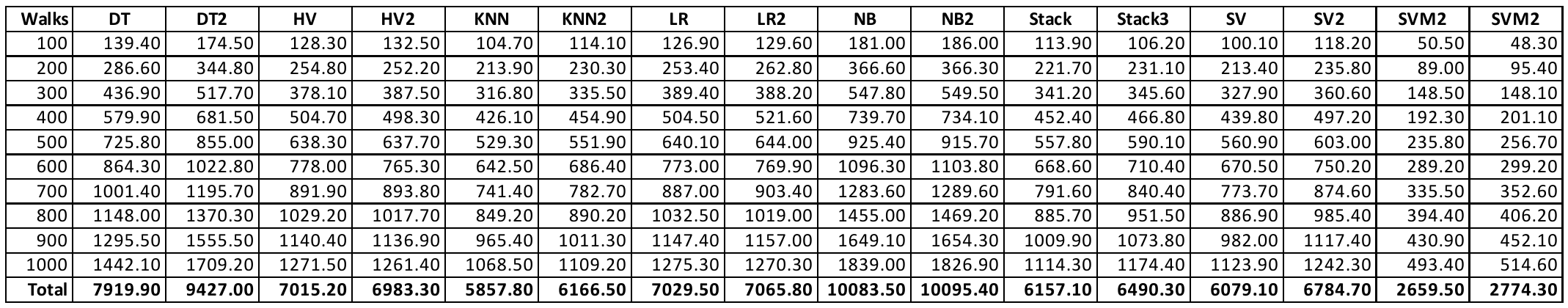}\\
	\scriptsize{(c) Directed Walk Strategy}\\
	\includegraphics[width=13cm, height=2.3cm]{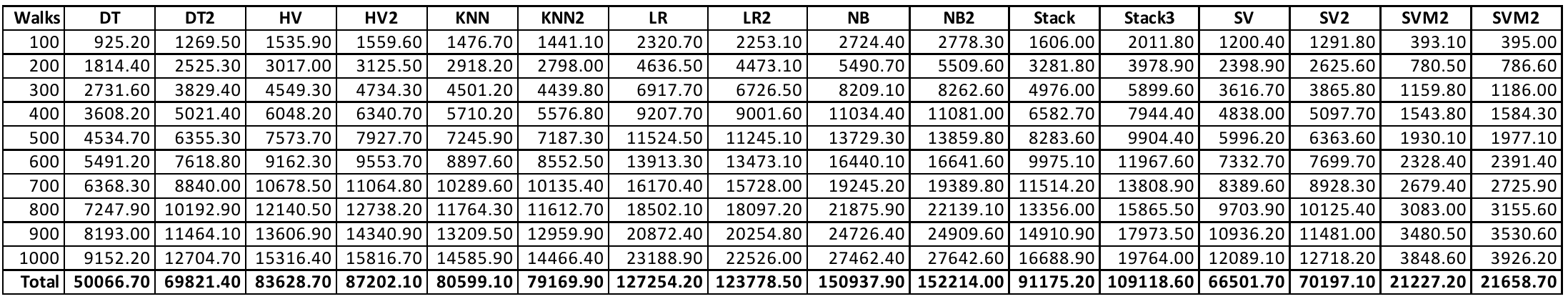}\\
	\end{center}
\end{table}

\begin{table}[htbp]
	\caption{Average Cost}	\label{tab:RedWineEffectiveness}
	\begin{center}
	\scriptsize{(a) Random Target Strategy}\\
	\includegraphics[width=13cm, height=2.3cm]{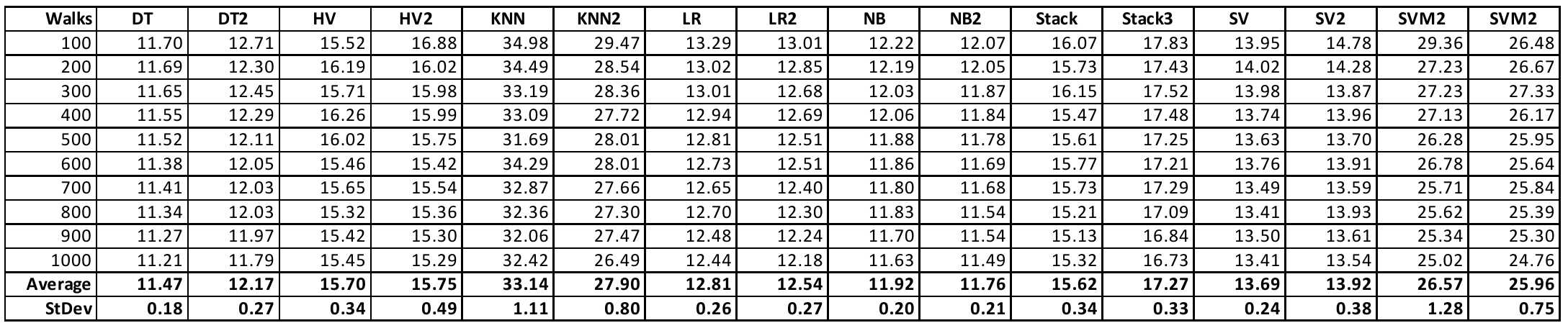}\\
	\scriptsize{(b) Random Walk Strategy}\\
	\includegraphics[width=13cm, height=2.3cm]{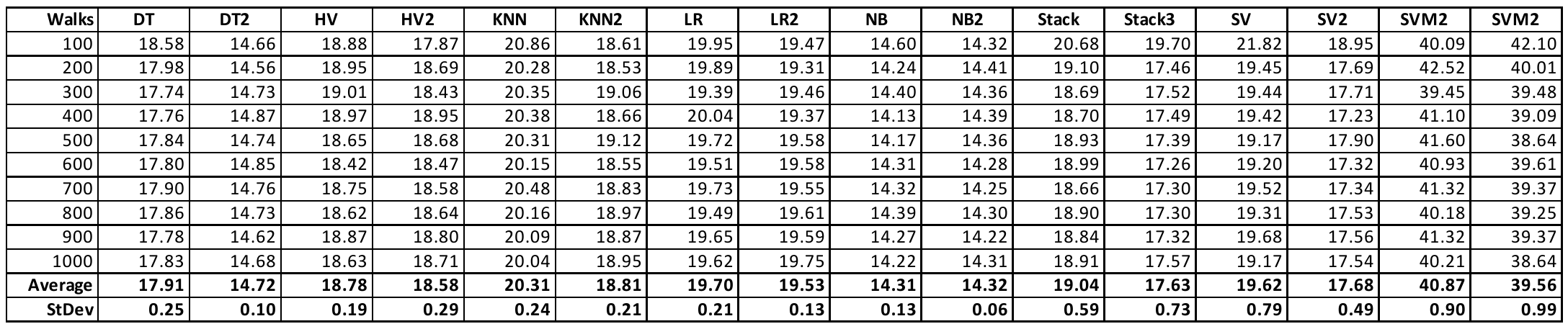}\\
		\scriptsize{(c) Directed Walk Strategy}\\
	\includegraphics[width=13cm, height=2.3cm]{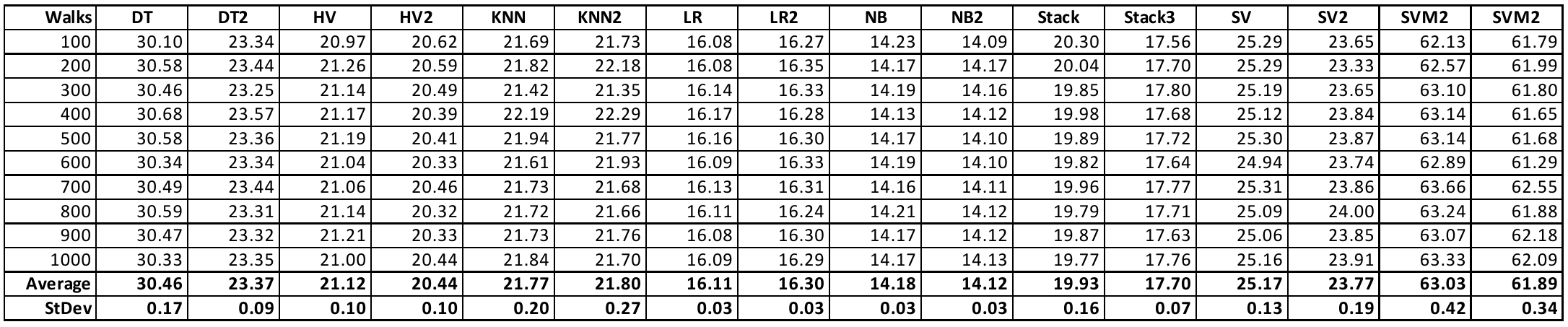}\\
	\end{center}
\end{table}

\begin{table}[htbp]
	\caption{Average Capability}	\label{tab:RedWineCapability}
	\begin{center}
	\scriptsize{(a) Random Target Strategy}\\
	\includegraphics[width=13cm, height=2.3cm]{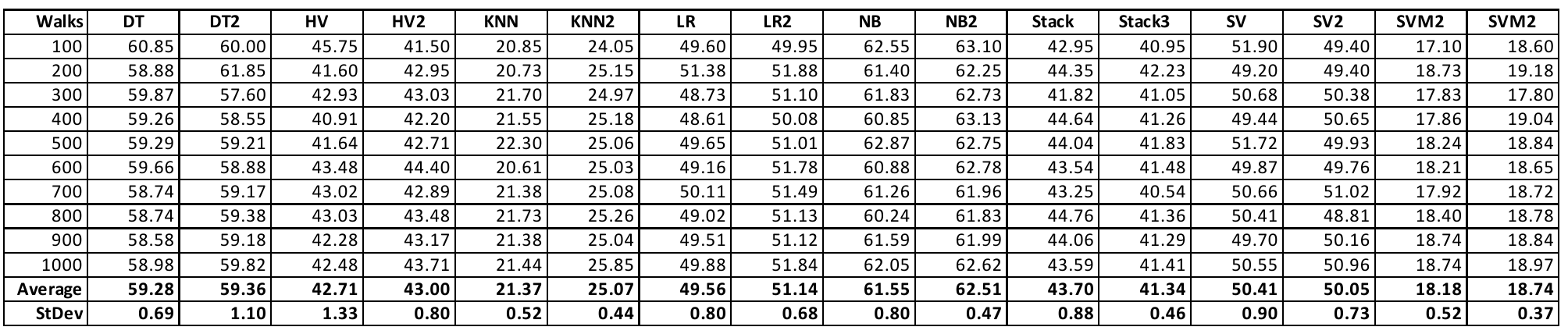}\\
	\scriptsize{(b) Random Walk Strategy}\\
	\includegraphics[width=13cm, height=2.3cm]{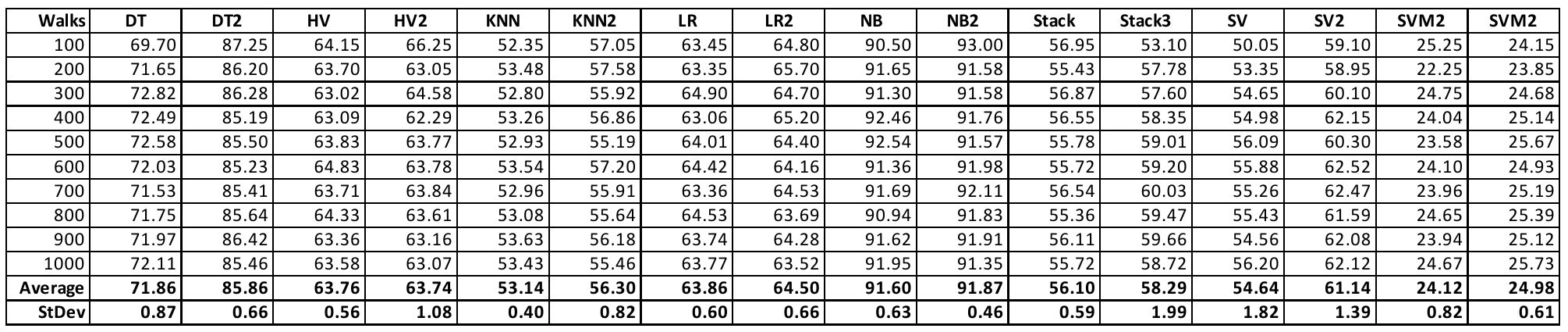}\\
		\scriptsize{(c) Directed Walk Strategy}\\
	\includegraphics[width=13cm, height=2.3cm]{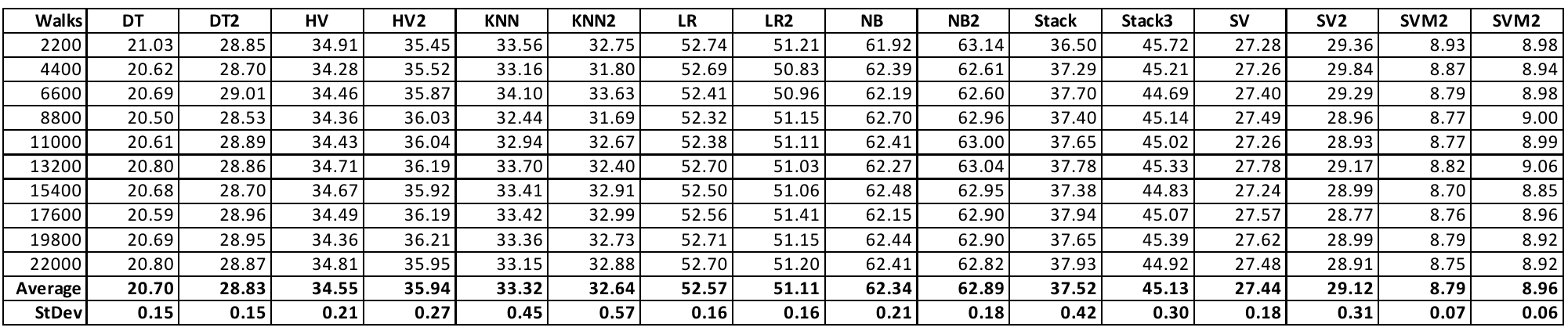}\\
	\end{center}
\end{table}

\newpage
\subsection*{B.3. Data of The Case Study on Mushroom Edibility}

Table \ref{tab:MushroomRuns}, \ref{tab:MushroomMutants}, \ref{tab:MushroomEffectiveness} and \ref{tab:MushroomCapability} give the average numbers of runs, mutants, cost and capability of testing the mushroom edibility classification models using three strategies.  

\begin{table}[htbp]
	\caption{Average Number of Runs}	\label{tab:MushroomRuns}
	\begin{center}
	\scriptsize{(a) Random Target Strategy}\\
	\includegraphics[width=13cm, height=2.3cm]{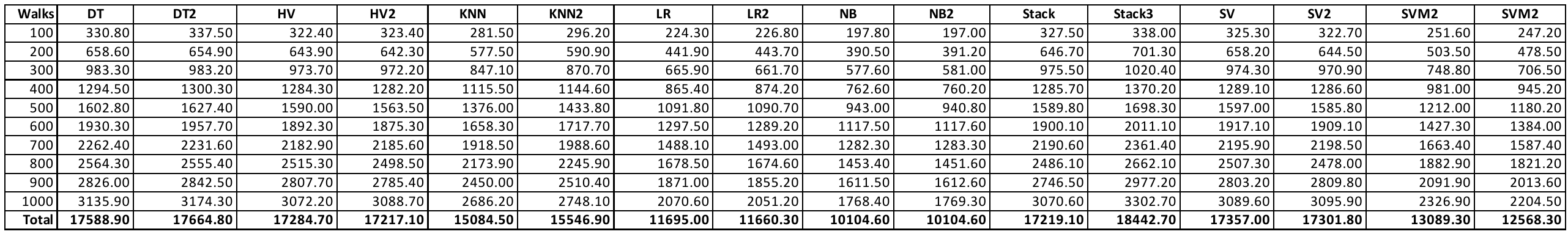}\\
	\scriptsize{(b) Random Walk Strategy}\\
	\includegraphics[width=13cm, height=2.3cm]{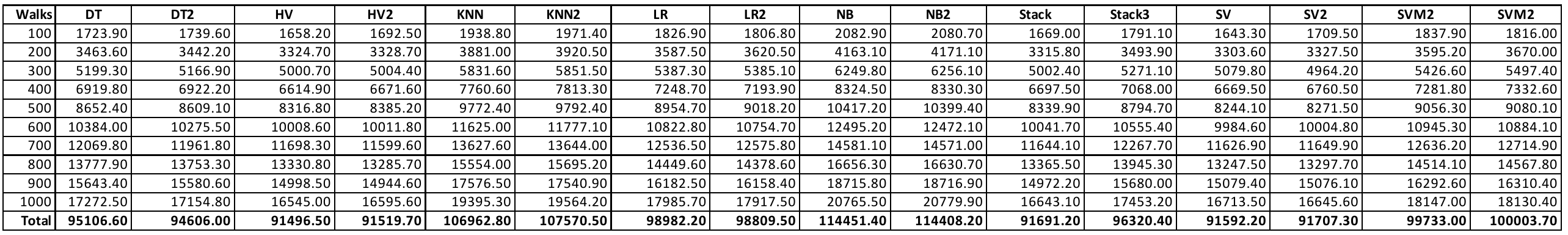}\\
		\scriptsize{(c) Directed Walk Strategy}\\
	\includegraphics[width=13cm, height=2.3cm]{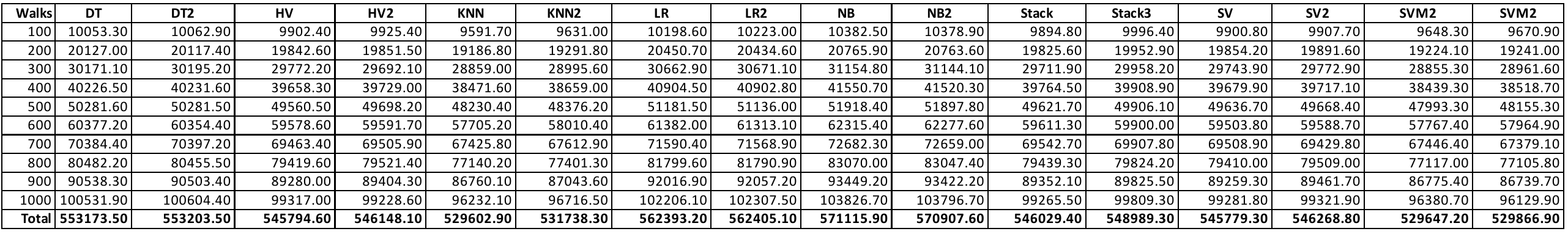}\\
	\end{center}
\end{table}

\begin{table}[htbp]
	\caption{Average Number of Mutants}	\label{tab:MushroomMutants}
	\begin{center}
	\scriptsize{(a) Random Target Strategy}\\
	\includegraphics[width=13cm, height=2.3cm]{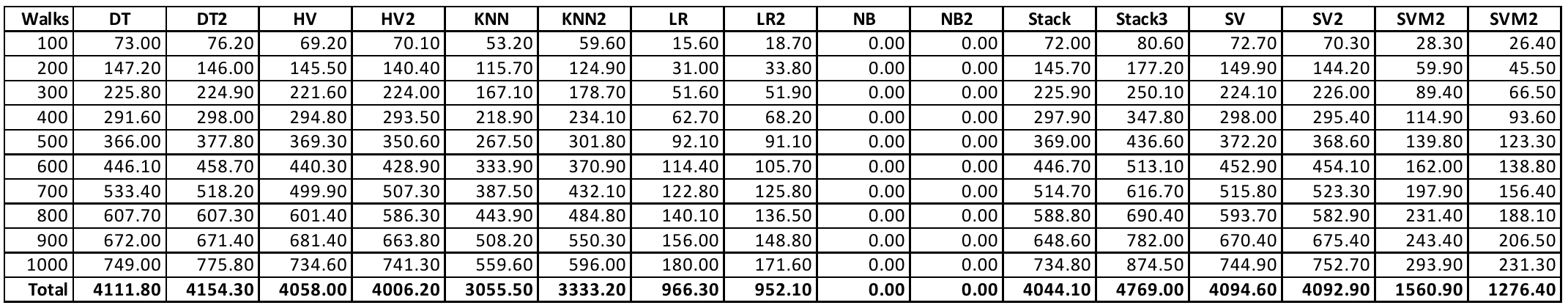}\\
	\scriptsize{(b) Random Walk Strategy}\\
	\includegraphics[width=13cm, height=2.3cm]{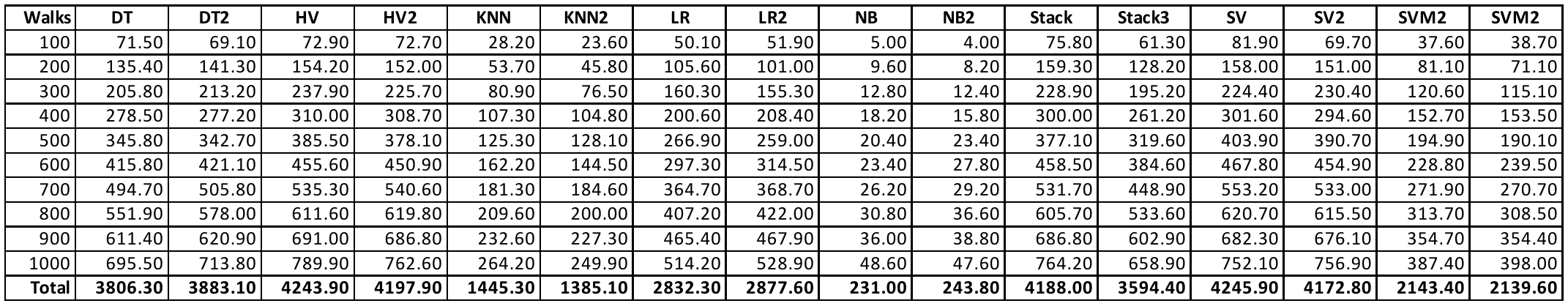}\\
		\scriptsize{(c) Directed Walk Strategy}\\
	\includegraphics[width=13cm, height=2.3cm]{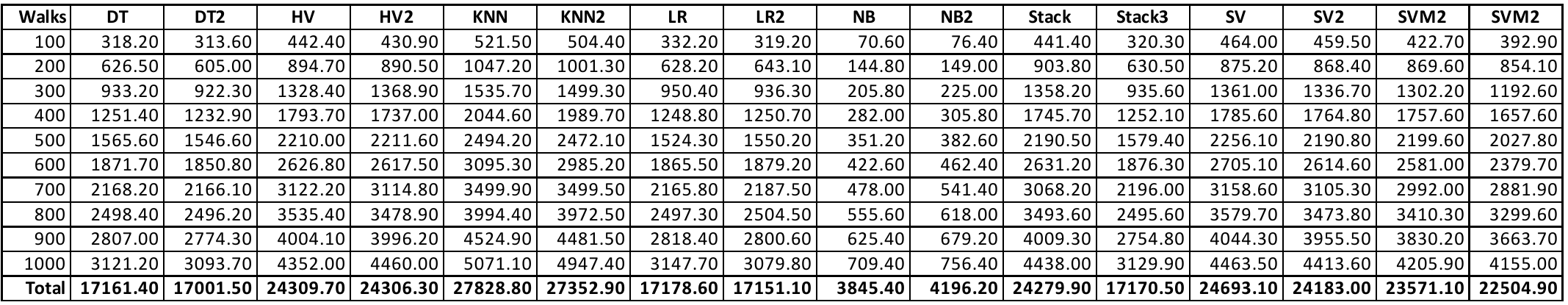}\\
	\end{center}
\end{table}

\begin{table}[htbp]
	\caption{Average Cost}	\label{tab:MushroomEffectiveness}
	\begin{center}
	\scriptsize{(a) Random Target Strategy}\\
	\includegraphics[width=13cm, height=2.3cm]{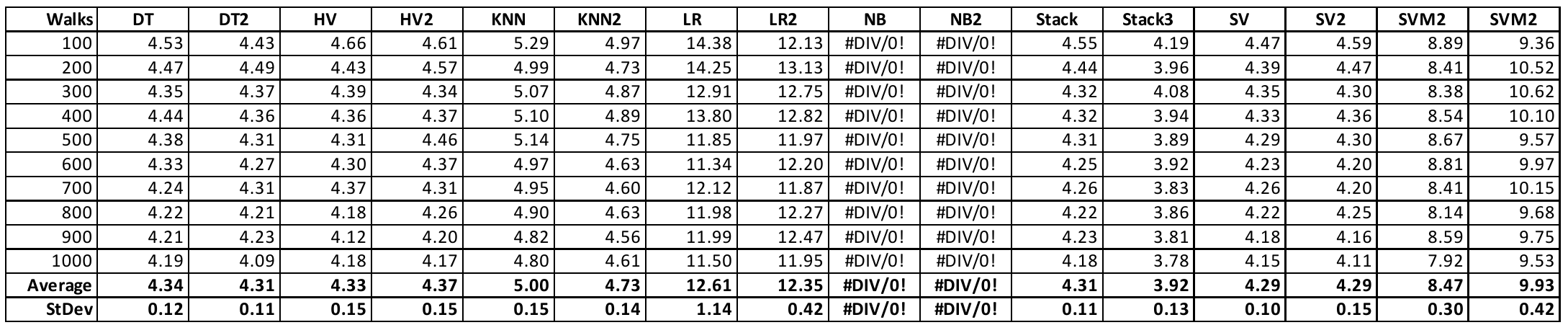}\\
	\scriptsize{(b) Random Walk Strategy}\\
	\includegraphics[width=13cm, height=2.3cm]{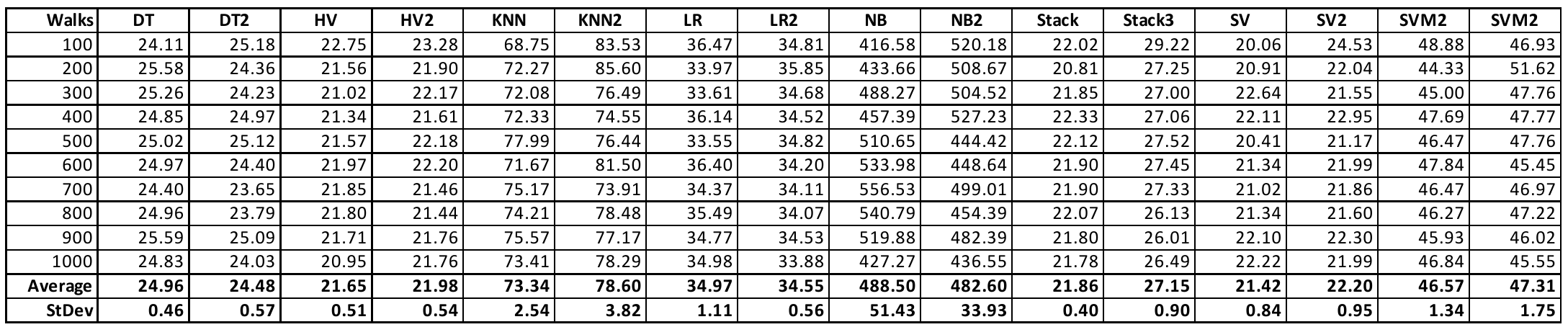}\\
		\scriptsize{(c) Directed Walk Strategy}\\
	\includegraphics[width=13cm, height=2.3cm]{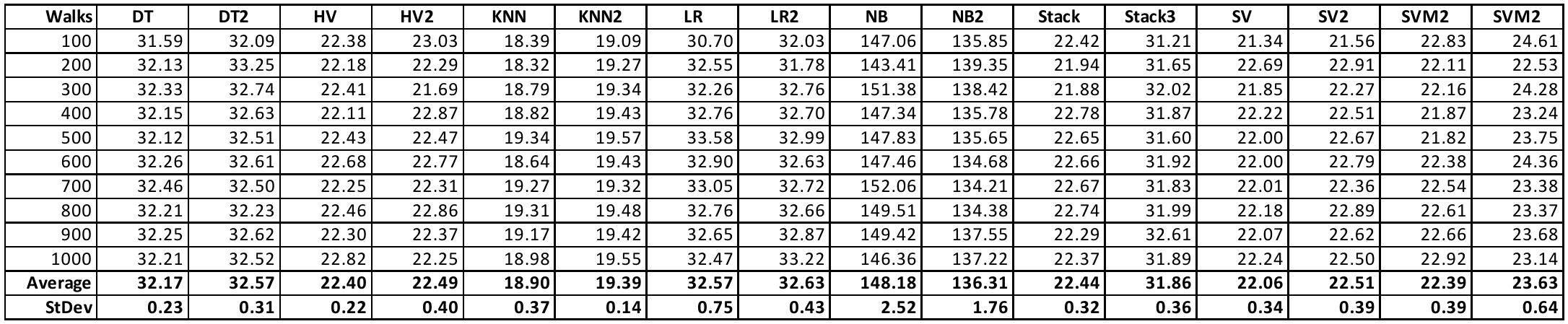}\\
	\end{center}
\end{table}

\begin{table}[htbp]
	\caption{Average Capability}	\label{tab:MushroomCapability}
	\begin{center}
	\scriptsize{(a) Random Target Strategy}\\
	\includegraphics[width=13cm, height=2.3cm]{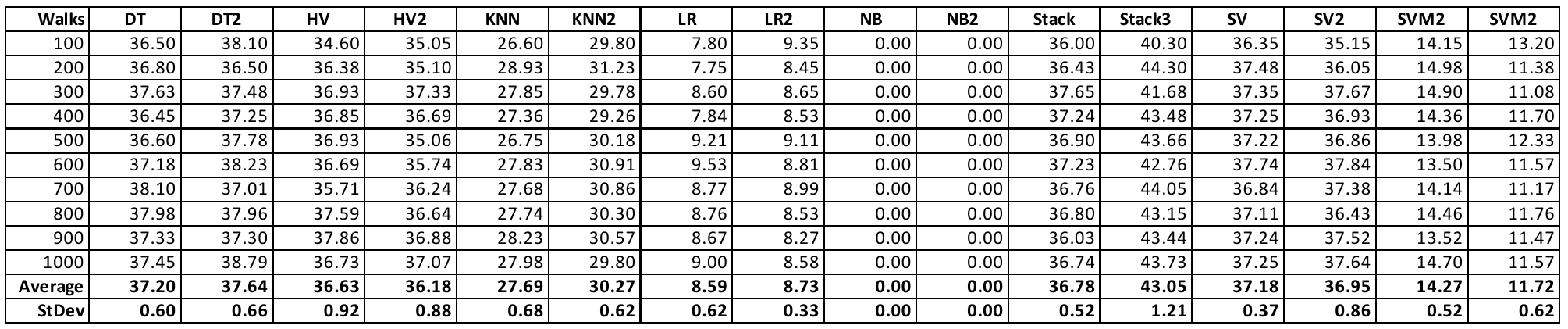}\\
	\scriptsize{(b) Random Walk Strategy}\\
	\includegraphics[width=13cm, height=2.3cm]{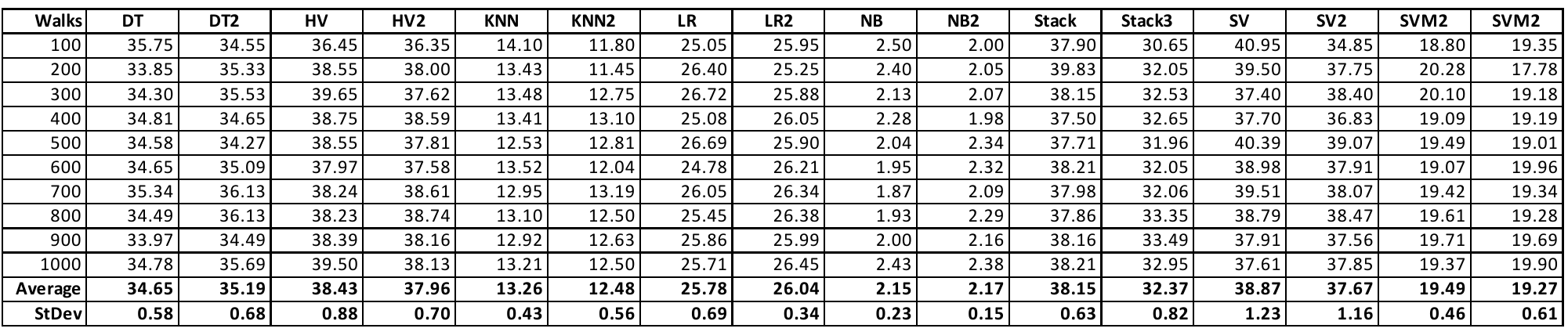}\\
		\scriptsize{(c) Directed Walk Strategy}\\
	\includegraphics[width=13cm, height=2.3cm]{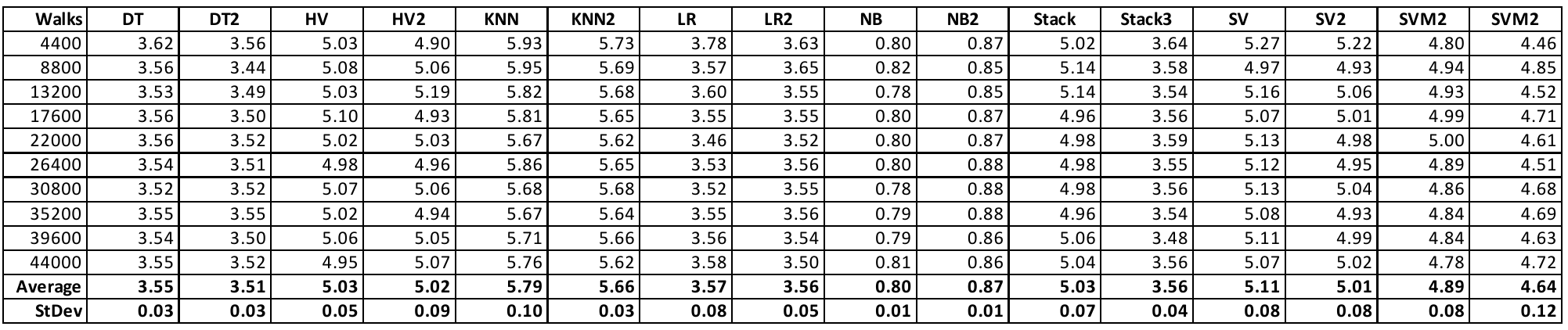}\\
	\end{center}
\end{table}

\newpage
\subsection*{B.4. Data of The Case Study on Bank Churner Prediction}

Table \ref{tab:BankRuns}, \ref{tab:BankMutants}, \ref{tab:BankEffectiveness} and \ref{tab:BankCapability} give the average numbers of runs, mutants, cost and capability of testing the bank churners prediction models using three strategies.  

\begin{table}[htbp]
	\caption{Average Number of Runs}	\label{tab:BankRuns}
	\begin{center}
	\scriptsize{(a) Random Target Strategy}\\
	\includegraphics[width=13cm, height=2.3cm]{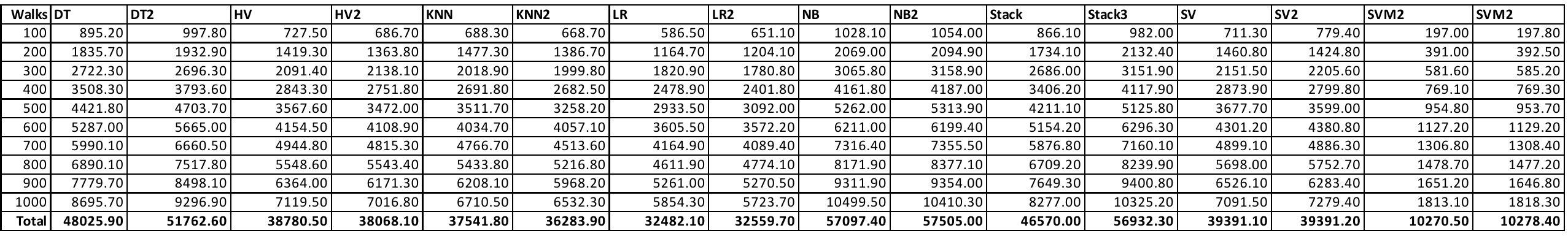}\\
	\scriptsize{(b) Random Walk Strategy}\\
	\includegraphics[width=13cm, height=2.3cm]{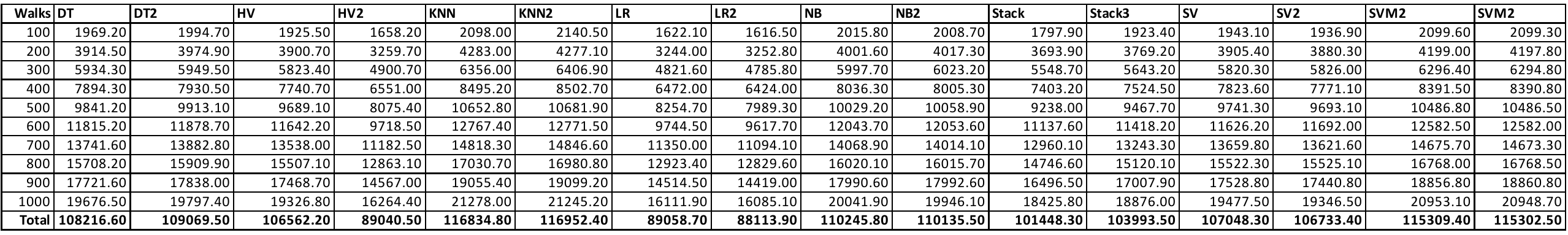}\\
		\scriptsize{(c) Directed Walk Strategy}\\
	\includegraphics[width=13cm, height=2.3cm]{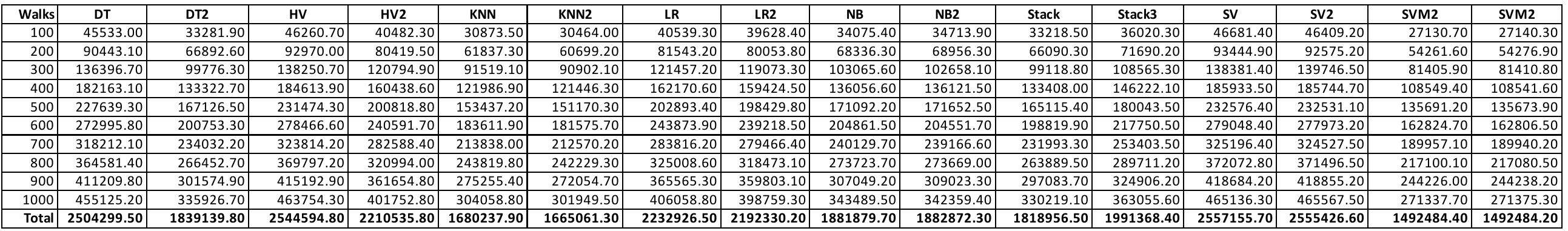}\\
	\end{center}
\end{table}

\begin{table}[htbp]
	\caption{Average Number of Mutants}	\label{tab:BankMutants}
	\begin{center}
	\scriptsize{(a) Random Target Strategy}\\
	\includegraphics[width=13cm, height=2.3cm]{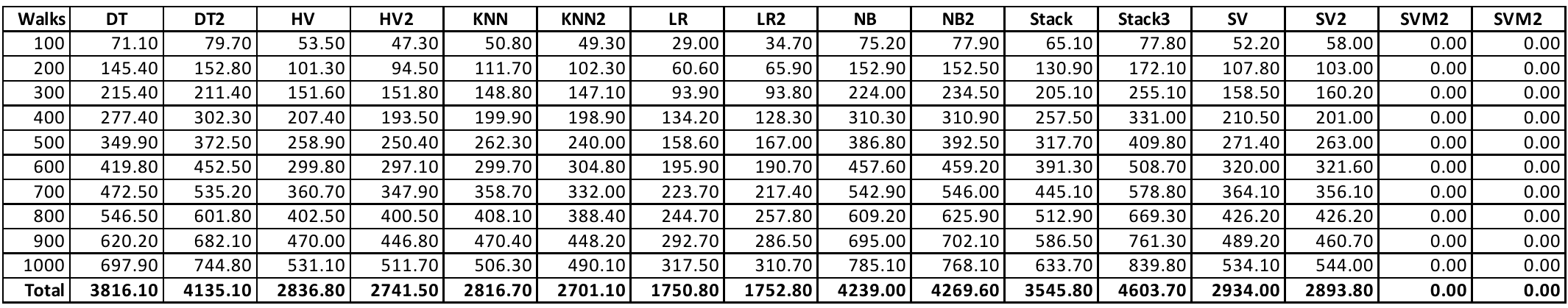}\\
	\scriptsize{(b) Random Walk Strategy}\\
	\includegraphics[width=13cm, height=2.3cm]{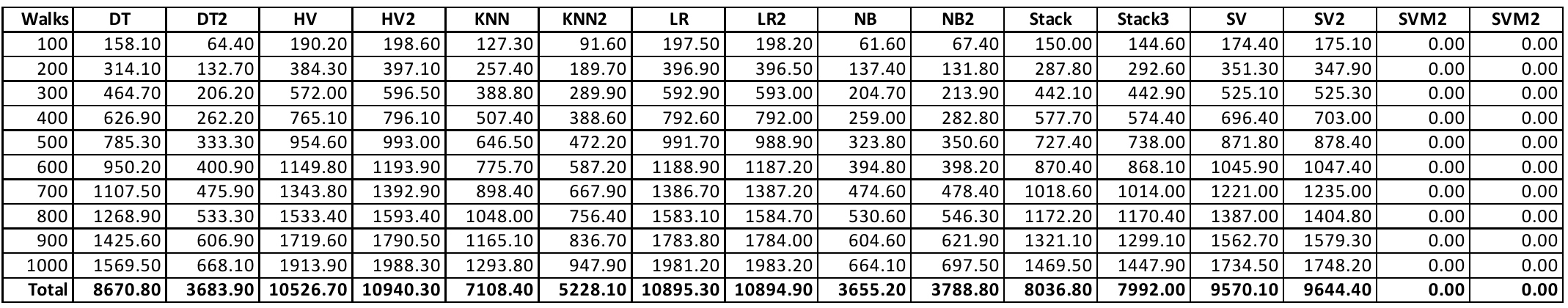}\\
		\scriptsize{(c) Directed Walk Strategy}\\
	\includegraphics[width=13cm, height=2.3cm]{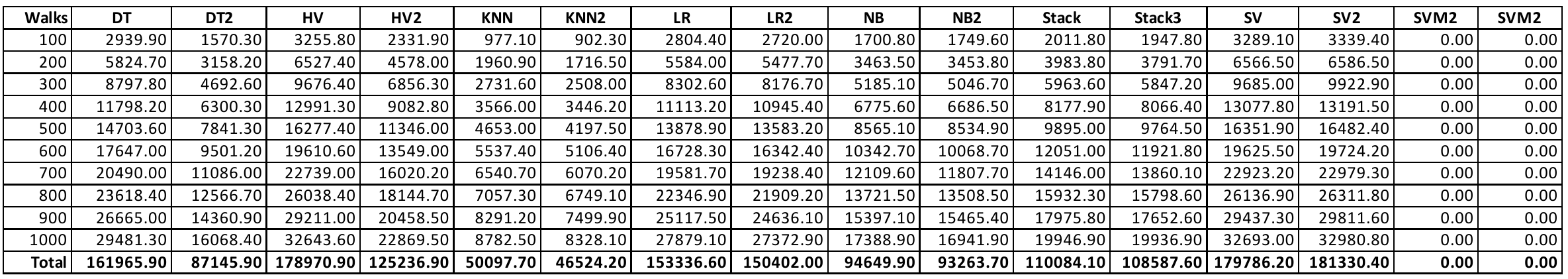}\\
	\end{center}
\end{table}

\begin{table}[htbp]
	\caption{Average Cost}	\label{tab:BankEffectiveness}
	\begin{center}
	\scriptsize{(a) Random Target Strategy}\\
	\includegraphics[width=13cm, height=2.3cm]{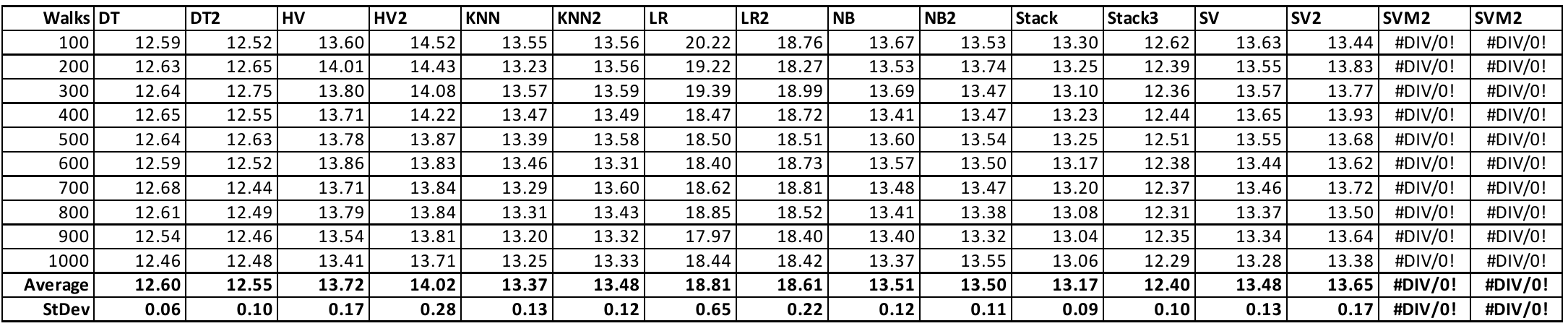}\\
	\scriptsize{(b) Random Walk Strategy}\\
	\includegraphics[width=13cm, height=2.3cm]{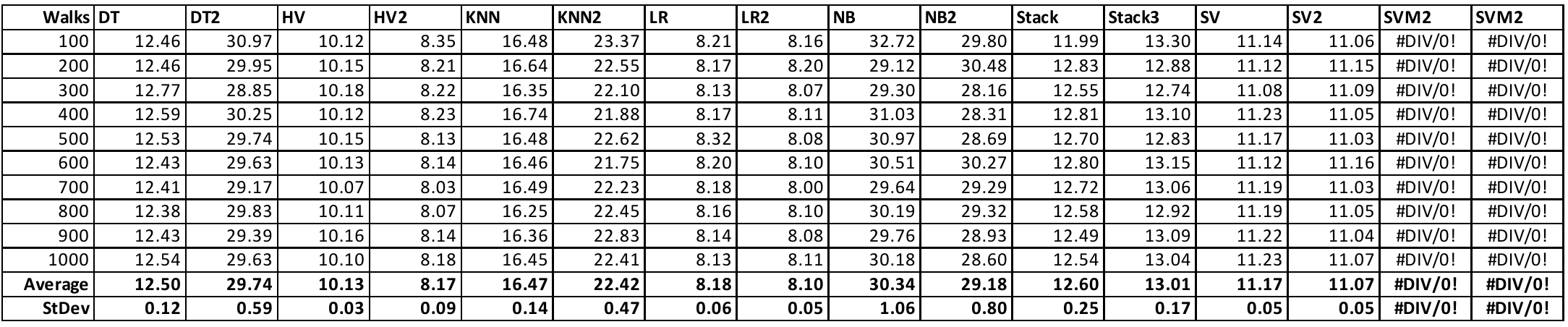}\\
		\scriptsize{(c) Directed Walk Strategy}\\
	\includegraphics[width=13cm, height=2.3cm]{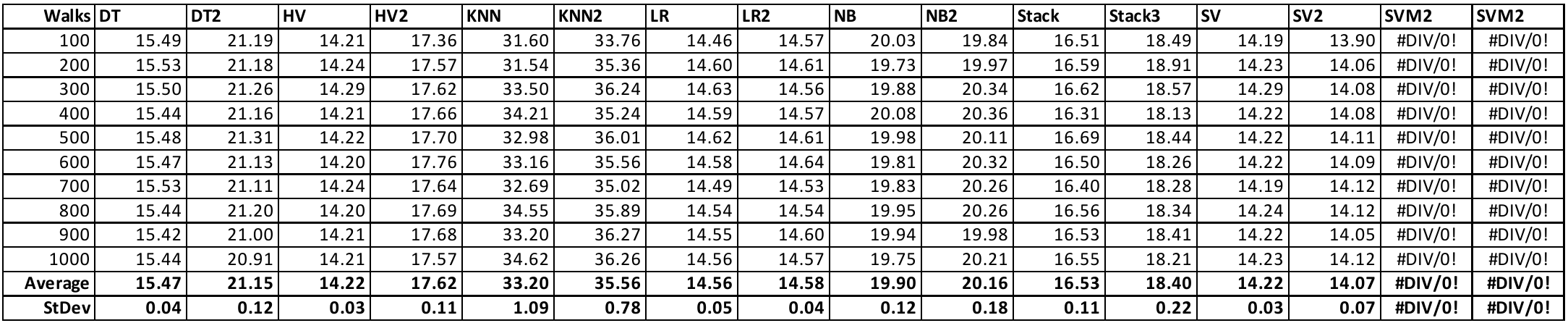}\\
	\end{center}
\end{table}

\begin{table}[htbp]
	\caption{Average Capability}	\label{tab:BankCapability}
	\begin{center}
	\scriptsize{(a) Random Target Strategy}\\
	\includegraphics[width=13cm, height=2.3cm]{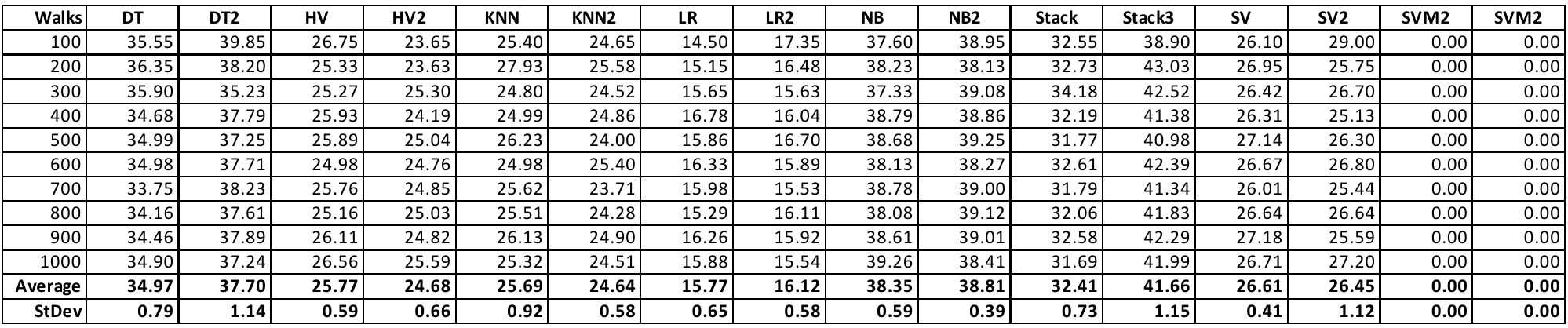}\\
	\scriptsize{(b) Random Walk Strategy}\\
	\includegraphics[width=13cm, height=2.3cm]{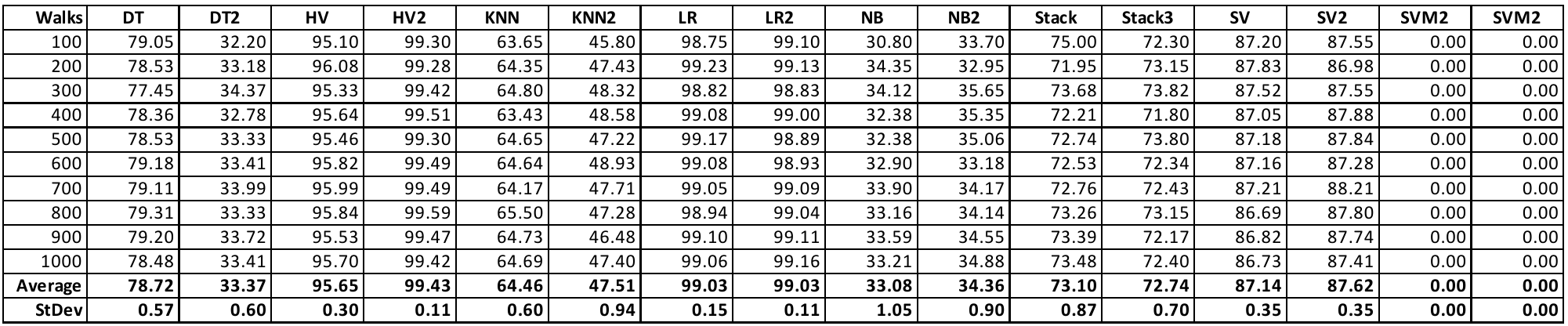}\\
		\scriptsize{(c) Directed Walk Strategy}\\
	\includegraphics[width=13cm, height=2.3cm]{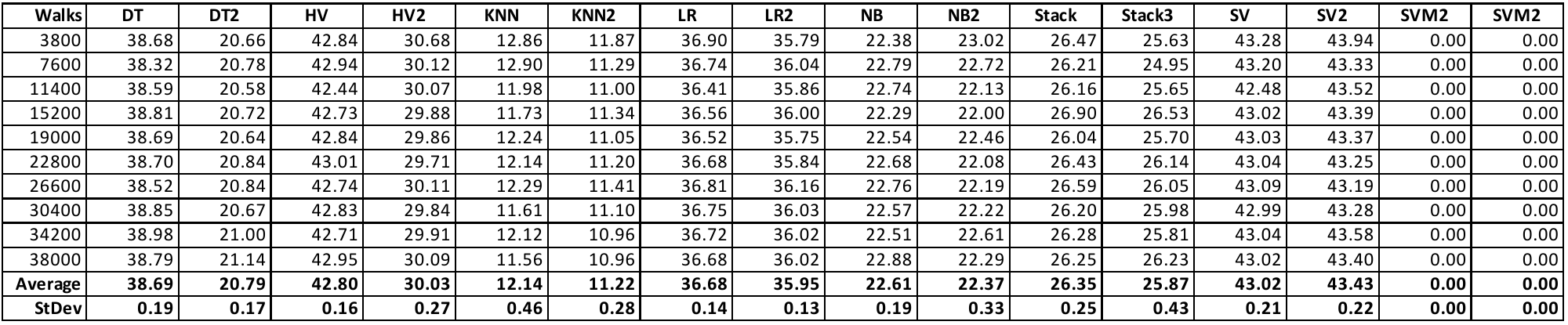}\\
	\end{center}
\end{table}

\end{document}